\documentclass[final,3p,times]{elsarticle}

\usepackage{graphicx}
\usepackage{amssymb}
\usepackage{amsthm}
\usepackage{amsmath}
\usepackage{amsfonts}
\usepackage{amssymb}
\usepackage{bbm} 

\usepackage{caption}
\usepackage{subcaption}
\usepackage{multirow}
\usepackage{array}

\usepackage{lineno}

\usepackage[T1]{fontenc}
\usepackage[utf8]{inputenc}
\usepackage{latexsym}
\usepackage{amsmath}

\journal{Journal Name}

\begin{document}

\begin{frontmatter}


\title{A New Artificial Neuron Proposal with Trainable Simultaneous Local and Global Activation Function}



\author[UFRPE,SEAS]{Tiago A. E. Ferreira\corref{mycorrespondingauthor}}
\cortext[mycorrespondingauthor]{Corresponding author}
\ead{tiago.espinola@ufrpe.br}

\author[SEAS] {Marios Mattheakis}
\author[SEAS] {Pavlos Protopapas}

\address[UFRPE]{Universidade Federal Rural de Pernambuco, 
                Departamento de Estat\'{i}stica e Inform\'{a}tica,\
                Dois Irm\~{a}os, 52171-900 Recife - PE, Brasil }
                

\address[SEAS]{John A. Paulson School of Engineering and Applied Sciences, Harvard University \\ Cambridge, Massachusetts 02138, United States}

\begin{abstract}
The activation function plays a fundamental role in the artificial neural network learning process. However, there is no obvious choice or procedure to determine the best activation function, which depends on the problem. This study proposes a new artificial neuron, named global-local neuron, with a trainable activation function composed of two components, a global and a local. The global component term used here is relative to a mathematical function to describe a general feature present in all problem domain. The local component is a function that can represent a localized behavior, like a transient or a perturbation. This new neuron can define the importance of each activation function component in the learning phase. Depending on the problem, it results in a purely global, or purely local, or a mixed global and local activation function after the training phase.  Here, the trigonometric sine function was employed for the global component and the hyperbolic tangent for the local component. The proposed neuron was tested for problems where the target was a purely global function, or purely local function, or a composition of two global and local functions. Two classes of test problems were investigated, regression problems and differential equations solving. The experimental tests demonstrated the Global-Local Neuron network's superior performance, compared with simple neural networks with sine or hyperbolic tangent activation function, and with a hybrid network that combines these two simple neural networks.  
\end{abstract}


\end{frontmatter}


\section{Introduction}\label{sec:Introduction}

An Artificial Neural Network (ANN) is a universal approximator for mathematical functions~\cite{Cybenko1989}. In this way, the ANNs are employed in a vast set of problems with many numerical applications, like in regressions problems \cite{Ertugrul2018,Niu2019} and differential equations solving \cite{Lagaris1998,Jagtap2020,Meng2020,mattheakis2020hamiltonian,mattheakis2020Covid,mattheakis2020Quantum}. The choice of the activation function plays a vital role in the ANN convergence process and precision performance. The activation function depends on the study's problem, and there is no obvious choice or procedure to determine the best activation function before the training process.  Several works in the literature board the ideal activation function determination problem proposing a trainable or adaptive activation function. \citet{Yu2002} proposed an adaptive activation function and an effective learning method based on the backpropagation algorithm to adjust the activation function parameters and the ANN weights. \citet{Dushkoff2016} proposed a multiple activation functions for a convolutional ANN, where each neuron has a specific activation function.  \citet{Li2013} presented a tunable activation function employed for Extreme Learning Machine, while \citet{Shen2004} applied a similar idea, but extended for a tunable activation function with multiple outputs. \citet{Jagtap2020} employed an adaptive activation function for problems of linear and nonlinear partial differential equations in Physics-Informed Neural Networks. 

In this pursuit by the ideal activation function, a very interesting work was developed by \citet{Qian2018}. This study proposed strategies for combining basic activation functions, like ReLU and its variants, in a convolutional neural network. One of the their proposals was to create a mixed activation function of the form $f_{\text{mix}}(x) = p \cdot f_1(x) + (1-p) \cdot f_2(x)$, where $p \in [0,1]$ is a combination coefficient, and $f_1$ and $f_2$ are two basic activation functions to be combined. More specifically, in the reference \cite{Qian2018}, ReLU, leaky ReLU (LReLU), parametric ReLU (PReLU), exponential linear unit (ELU), and parametric ELU (PELU) were employed as the two activation functions.

There are many situations where a mathematical function governs a given problem with two features, global and local behavior. These double-feature problems are relatively common in modeling mathematical, physical, and engineering problems, such as regression problems and differential equation solving.  The global component is some characteristic present in its entire problem domain, like an undulatory behavior of the wave signal. The local component describes behavior located in a particular part of the problem domain, like a transient or a localized perturbation. For this problem class, ANN modeling can improve precision and convergence time if the activation function has both global and local characteristics. 

Inspired in the work of \citet{Qian2018}, we propose a new artificial neuron named global-local neuron (GLN), with an activation function composed of two mathematical functions, one with a global behavior and the other with a local behavior. The GLN has the capability to adjust the relative importance of each of these components in its training process. A typical gradient descendant algorithm, like the Adam algorithm~\cite{kingma2014adam}, can be applied. Therefore, the GLN has the capability to adjust the importance of each global and local components. The GLN can choose if the activation function is purely global, purely local, or a combination of these two components. 

To demonstrate the GLN versatility, problems with local,  global, and a combination of these two components were employed in three regression tasks and seven differential equation solving.  Three other ANN models were also used in the same problems to establish relative baseline performance. A multilayer perceptron (MLP) with the sine activation function (MLP-Sin), an MLP with hyperbolic tangent activation function (MLP-Tanh), and hybrid MLP, with an MLP-Sin and an MLP-Tanh branches combination, called two branches network (TBN).

The article is drawn as follows. Section \ref{sec:ProposedNeuron} presents the proposed GLN and its global and local activation functions. Section \ref{sec:TestProb} introduces the basic ideas of the problem classes board here, the regression problem and the differential equation solving by ANNs. Section \ref{sec:Data} exhibits the data and the definitions of the differential equations. Section \ref{sec:ExperimentsResults} exposes all experimental results, where a total of 2400 computational simulations were done. Finally, Section \ref{sec:Conclusions} presents the proposed ANN architectures' final conclusions.      

\section{Proposed Neuron}\label{sec:ProposedNeuron}

This work proposes a new artificial neuron named global-local neuron (GLN), with an activation function composed of two mathematical functions, a local and a global function. The main idea is to give the neuron the ability to combine a function with a global feature, like a trigonometric sine, and a function with a local transient behavior, like a hyperbolic tangent or a sigmoid logistic function. 

\begin{figure}
\centering
\includegraphics[scale=1]{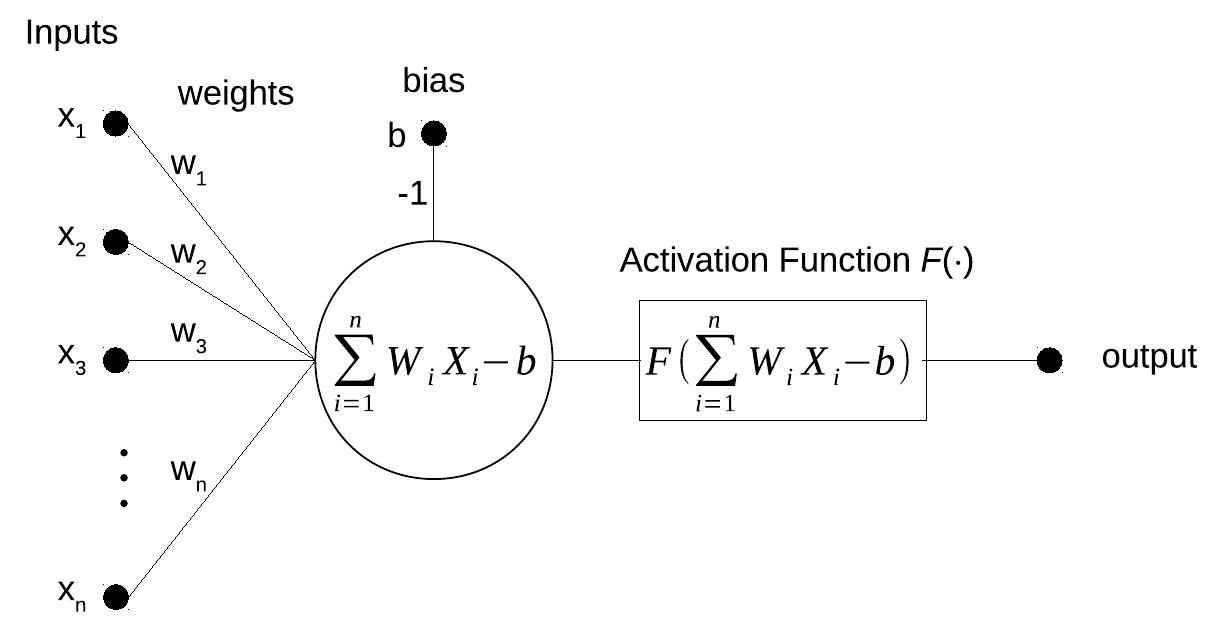}
\caption{Scheme of a simple artificial neuron.}\label{fig:simpleNeuron}
\end{figure}

In general, an artificial neuron can be described as a mathematical function with a weighted sum of arguments, as shown in Figure~\ref{fig:simpleNeuron}.  The neuron receives the inputs $(x_1, x_2, x_3, \dots, x_n)$, multiples by the respective weights $(w_1, w_2, w_3, \dots, w_n)$, sums them, and adds a bias $b$. This sum is the argument of the activation function $F(\cdot)$ and the result of the activation function processing is the output of the neuron.

\begin{figure}
\centering
\includegraphics[scale=0.8]{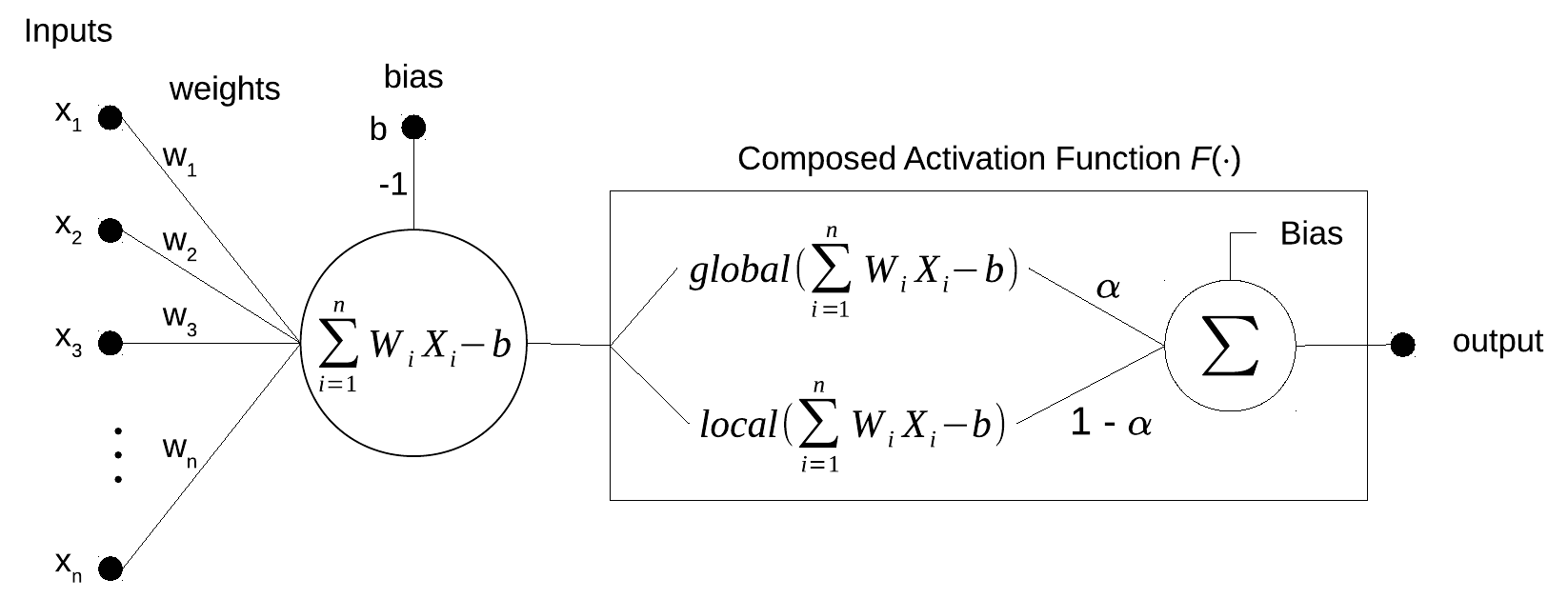}
\caption{Scheme of the Global-Local Neuron proposed.}\label{fig:GLN}
\end{figure}

The proposed GLN has a composed activation function $F(\cdot)$ with local and global components combined with a weight $0 \leq \alpha \leq 1$ given by,

\begin{equation}\label{eqn:actFunc}
    F(x) = (\alpha \cdot \text{global}(x) + (1-\alpha) \cdot \text{local}(x)) - \text{Bias}, 
\end{equation}
where global($\cdot$) is the global function component, and local($\cdot$) is the local function component. The weight $\alpha$ and the Bias are adjusted in the ANN training process. The scheme of GLN is shown in Figure~\ref{fig:GLN}. Worth note that $\alpha$ and Bias are part of the activation function. In this way, if the same activation function is used in a layer, the adjustment of these parameters is unique for all neuron of the layer. In the development process of GLN, the $\alpha$ combination weight was implemented as

\begin{equation}
    \alpha = \text{sig}(z), \, z \in \mathbbm{R} 
\end{equation}
where sig($\cdot$) is the sigmoid logistic function given by,
\begin{equation}
    \text{sig}(z) = \frac{1}{1+\exp(-z)}
\end{equation}

The implementation employed the sine as the global component and hyperbolic tangent as the local component. All the network's hidden layers have the same activation function. Therefore, with the proposed GLN, an ANN of the MLP type was built, where all hidden neurons are GLN. This multilayer perceptron of GLN (MLP-GLN) has the same activation in each hidden layer (implying an $\alpha$ weight per hidden layer), while in the last layer (the output layer) all the GLN outputs are weighted summed in a neuron with a linear activation function.     

\section{Test Problems}\label{sec:TestProb}

In this work, two types of problems were employed to test the proposed GLN, the regression problem, and differential equations solving. For both classes of problems, the idea is to test the MLP-GLN capability to solve them by combining the global (sine function) and local (hyperbolic tangent) components in its activation function.

\subsection{The Regression Problem}\label{sec:Def_Regr_Prob}

Typically, when a neural network is applied to solve a regression problem, it is sought to find some functional description of given data. The regression of a dependent variable $Y$, given an independent variable $X$, consists of finding the model to determine the most probable value of $Y$ for each value of $X$, based on a finite data set $(X,Y)$.

In this way, the ANN maps the independent variable $X$ to the dependent variable $Y$. In other words, the ANN learns functional structure $\mathcal{G}(\cdot)$, where for a given value of $X$, it is possible to compute $Y=\mathcal{G}(X)$.

\subsection{Differential Equations Solving with Neural Networks}

Differential equations can model many scientific and engineering problems. However, for many physical systems of practical interest, these differential equations are analytically intractable. Consequently, there is a great interest in developing computational techniques to solve differential equations numerically. 

ANN had been applied to solve differential equations \cite{Lagaris1998,Jagtap2020,Randle2020,Giovanni2020b}. The basic principle for solving differential equations with ANNs is to look at the problem as an optimization problem. Defining a general differential equation in the form, 
\begin{equation}\label{eqn:diffeq}
    \mathcal{D}(u)-\mathcal{F} = 0
\end{equation}
where $\mathcal{D}(\cdot)$ is a differential operator, $u$ is a possible solution of $\mathcal{D}(\cdot)$, and $\mathcal{F}$ is a known forcing function. 

Let $\widehat{u}$ the output of the ANN. If $\widehat{u}$ is a solution for the differential equation $\mathcal{D}$, then the residual is given by

\begin{equation}
    \mathcal{R}(\widehat{u}) = \mathcal{D}(\widehat{u}) - \mathcal{F}
\end{equation}

The main idea is to use the $\mathcal{R}^2(\widehat{u})$ as the loss function in the ANN training process, where the solving differential equation problem is reduced to a minimization problem.

To guarantee that the initial conditions are satisfied, $\widehat{u}$ can be substituted by a modified  solution $\widetilde{u}$. For example, if a certain differential equation in space $x$ and time $t$ has a initial condition in $t=t_0$  given by function $u_{t_0}(x)$, the  solution can be written as \cite{Giovanni2020},

\begin{equation}\label{eqn:trail1D}
    \widetilde{u}(x,t) = u_{t_0}(x) + \left( 1-e^{-(t-t_0)} \right) \widehat{u}(x,t)
\end{equation}
implying that when $t=t_0$, the trial solution is exactly $u_{t_0}(x)$.

This is implemented in the Python library \texttt{NeuroDiffEq} \cite{Giovanni2020} with many other initial and boundary conditions in the form, 
\begin{equation}
    \widetilde{u}(x,t) = A(x,t; x_{boundary},t_0)\widehat{u}(x,t),
\end{equation}
where $A(x,t; x_{boundary},t_0)$ is selected so that $\widetilde{u}(x,t)$ has the correct initial and boundary conditions.

\section{Data Sets for Regression and Differential Equations}\label{sec:Data}

In this study, the proposed GLN were tested in two types of problems. The first problem is the regression problem, and the second problem is the differential equation solving.



\subsection{Regression Problem Data Sets}

The data sets employed to test the MLP-GLN in the regression problem are presented in the next subsections. Two of them are simulated data set, and one is a real-world data set. The artificial data sets  encompass features with global and local components, with an addictive and multiplicative composition of the global and local characteristics. The real-world data, the sunspot yearly observation, also has global (a quasi-periodic behavior) and local (fluctuations) components.

\subsubsection{The EES Artificial Data Set}

\begin{figure}
\centering
\includegraphics[scale=0.6]{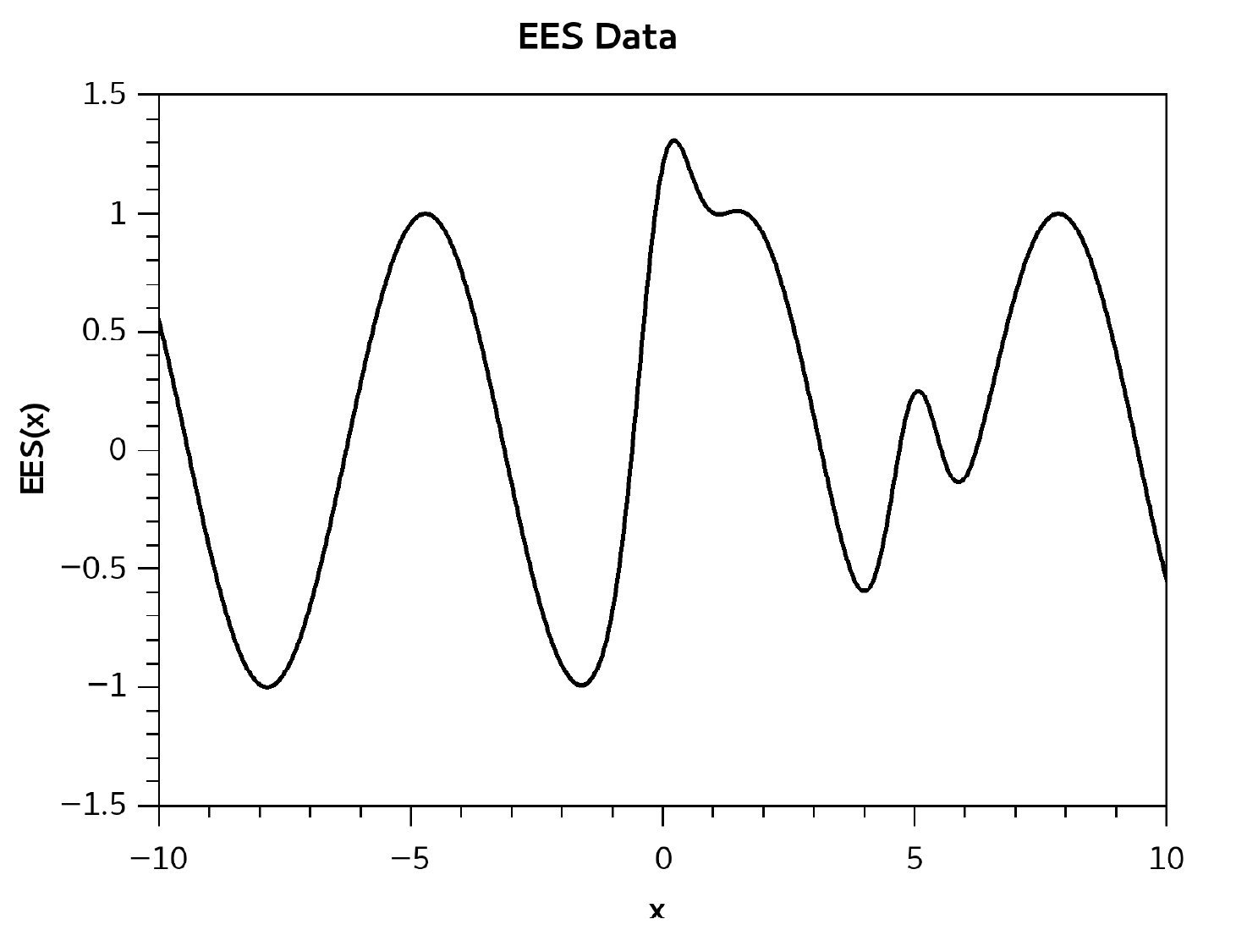}
\caption{EES data set from Equation~\ref{eqn:EES} with the parameters values from Table~\ref{tab:EES}.}\label{fig:EES}
\end{figure}

The first artificial data set developed to test the proposed GLN network was generated by adding two Gaussian functions (the local components) and a trigonometric sine function (the global component). Given these functions addition, this data set was called EES (\emph{Exponential, Exponential, Sine}) and generated by the equation,

\begin{equation}\label{eqn:EES}
    EES(x) = E_1\exp \left( \frac{-(x-a)^2}{2 \sigma ^2} \right) + E_2\exp \left(\frac{-x^2}{2\sigma ^2} \right) + \sin(\omega x)
\end{equation}
where $E_1$ and $E_2$ are constants representing each Gaussian function's amplitude, $a$ is the mean, $\sigma$ is the constant standard deviation for both exponential, and $\omega$ is the angular frequency of the sine function. Here, to build the EES data set, the values of these parameters are presented in Table~\ref{tab:EES}. Figure~\ref{fig:EES} shows the plot of Equation ~\ref{eqn:EES}, where we observe global behavior with a local fluctuation. The domain used here was $x \in [-10,10]$.

\begin{table}
\centering
\caption{The parameters values employed to build the EES data set.} \label{tab:EES}
\begin{tabular}{cc}
\hline
Parameters & Numeric Values \\
\hline
$E_1$ & 1.2\\
$E_2$ & 0.9\\
$a$ & 5.0 \\
$\sigma$ & 0.5\\
$\omega$ & 1.0\\
\hline
\end{tabular}
\end{table}

\subsubsection{The SE Artificial Data Set}

\begin{figure}
\centering
\includegraphics[scale=0.6]{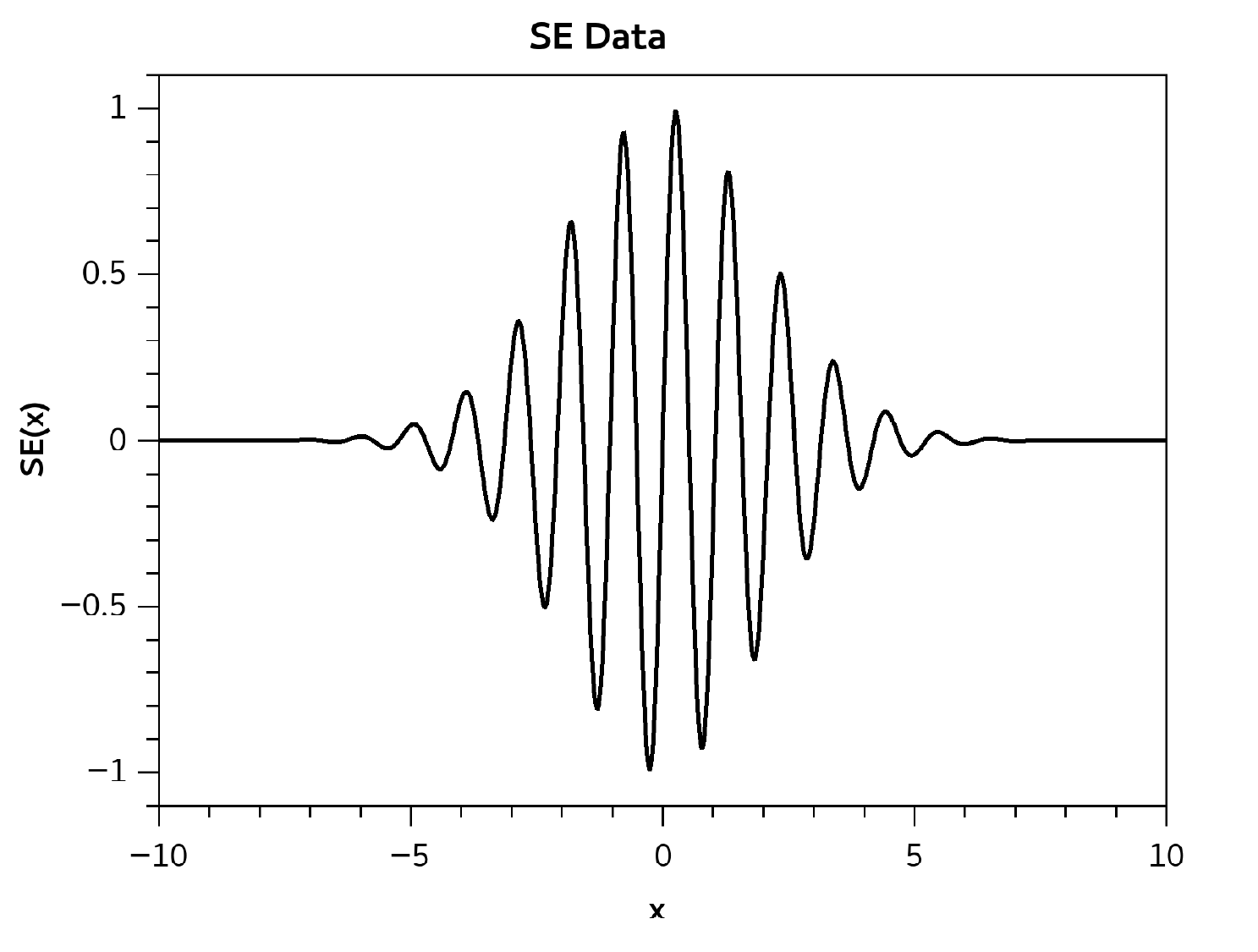}
\caption{SE Data set from Equation~\ref{eqn:SE} with the parameters' values from Table~\ref{tab:SE}.}\label{fig:SE}
\end{figure}

The second data set developed here was generated by multiplying a sine function with an exponential function. This data set was named SE and is given by the equation,

\begin{equation}\label{eqn:SE}
    SE(x) = E_1\sin(\omega x) \exp \left( \frac{-x^2}{2 \sigma ^2} \right)
\end{equation}
where one more time $E_1$ is constant, $\sigma$ is a constant standard deviation, and $\omega$ is sine function's angular frequency. In Table~\ref{tab:SE} are shown the values of these parameters used to build the SE data set. In Figure~\ref{fig:SE} is shown the plot of the Equation ~\ref{eqn:SE}, where now it is possible to note a local behavior dominance. The domain used was $x \in [-10,10]$.

\begin{table}
\centering
\caption{The parameters values employed to build the SE data set.} \label{tab:SE}
\begin{tabular}{cc}
\hline
Parameters & Numeric Values \\
\hline
$E_1$ & 1.0\\
$\sigma$ & 2.0\\
$\omega$ & 6.0\\
\hline
\end{tabular}
\end{table}

\subsubsection{The Sunspot Data Set}

Sunspots are spots darker than the surrounding areas on the Sun's photosphere. These spots are temporary phenomena that may last anywhere from a few days to a few months. Besides the solar activity study, the sunspot occurrence can influence the space weather, impacting the earth's ionosphere state, generating interference on satellite communications, and affecting the conditions of short-wave radio propagation~\cite{Banerjee2016,Hou2018,Prantika2018}.

\begin{figure}
\centering
\includegraphics[scale=0.6]{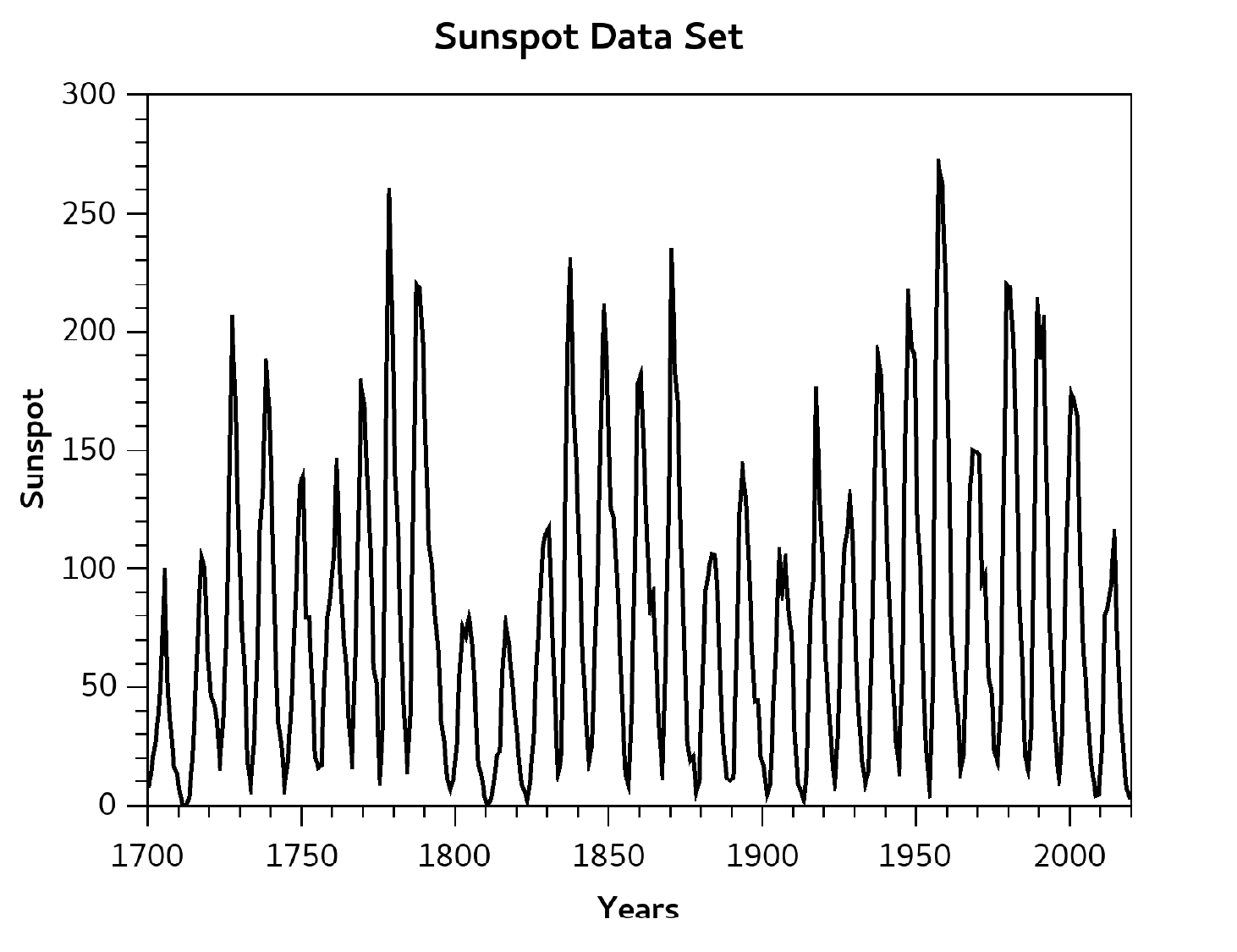}
\caption{Plot of the Sunspot data set.}\label{fig:Sunspot}
\end{figure}

The third data set was the yearly mean total sunspot number from 1700 until 2019. This data set consists of 320 points. The sunspot data used here were downloaded from the website SILSO (Sunspot Index and Long-term Solar Observations)\footnote{\emph{http://www.sidc.be/silso/datafiles}.}. The Figure~\ref{fig:Sunspot} shows the plot of the Sunspot Data set. This natural phenomenon presents a combination of global and local behavior.

\subsection{Differential Equations}\label{sec:DiffEq}

The differential equations used to test the proposed GLN are presented in the next sections. The equations selected here have possible solutions with only global features, only local features, and both global and local features. 

\subsubsection{Exponential Decay Equation}\label{sec:decayEq}

The first differential equation investigated is,
\begin{equation}
    \frac{du}{dt} = -u.
\end{equation}
For the initial condition $u(t=0) = 1.0$, the solution is given by $u(t) = e^{-t}$. 


\subsubsection{Catenary Equation}\label{sec:Catenary}

A catenary function \cite{Lockwood1961} is a mathematical curve that an idealized hanging rope assumes under its weight when supported only at its ends, like an electrical cable between two posts in an overhead electrical power. The differential equation whose solution is a catenary function can be written as,
\begin{equation}\label{eqn:catenary}
    \frac{d^2u}{dx^2} - \sqrt{1+\left( \frac{du}{dx} \right)^2} = 0.
\end{equation}

The solution is $u(x) = \cosh(x)$ (the catenary). The boundary conditions used here were $u(x=-2) = u(x=2) = 3.76$.


\subsubsection{Simple Harmonic Oscillator Equation}\label{sec:SimpleOscillator}

A simple harmonic oscillator~\cite{Marion2004} in one dimension can be described by the differential equation,

\begin{equation}\label{eqn:harmOsc}
    \frac{d^2u}{dt^2}+u = 0
\end{equation}

The solution of Equation \ref{eqn:harmOsc} is $u(t) = \sin(t)$, for the initial  conditions of $u(t=0) = 0.0$ and $\frac{du}{dt}|_{t=0} = 1.0$.

\subsubsection{Damped Harmonic Oscillator}\label{sec:DampedOsc}

A damped harmonic oscillator~\cite{Marion2004} in one dimension can be modeled by the differential equation,
\begin{equation}
    \frac{d^2u}{dt^2}+\frac{du}{dt}+ u = 0
\end{equation}

The solution of this equation is $u(t) = \exp\left( {\frac{-t}{2}} \right) \sin(t)$, for the initial conditions $u(t=0) = 0.0$ and $\frac{du(t)}{dt}|_{t=0} = 1.0$.

\subsubsection{Laplace Equation}\label{sec:Laplace}

The Laplace equation is a second order partial differential equation, where in two dimensions can be written as,
\begin{equation}
    \frac{\partial^2 u}{\partial x^2} + \frac{\partial^2 u}{\partial y^2} = 0
\end{equation}
The solution $u(x,y)$ is a function of two variables. For the boundary conditions $u(x=0,y) = \sin(\pi y)$, $u(x=1,y) = 0.0$, $u(x, y=0) = 0.0$, and $u(x,y=1)= 0$, an analytic solution for the Laplace equation in two dimensions is,
\begin{equation}\label{eqn:AnaSolLaplace}
    u(x,y) = \frac{\sin(\pi y) \sinh(\pi (1-x))}{\sinh(\pi)}.
\end{equation}

\subsubsection{Heat Equation}\label{sec:Heat}

The heat equation describes how the heat diffuses through a given region. In one spatial dimension, the heat equation is:
\begin{equation}
    \frac{\partial u}{\partial t} - k\frac{\partial^2 u}{\partial x^2} = 0,
\end{equation}
where $k$ is the diffusivity of the medium. This equation models the flow of heat in a homogeneous and isotropic medium, where $u(x,t)$ represents the temperature at the point $x$ in time $t$.

For the initial conditions $u(x,t=0) = \sin(\pi x)$, $\frac{\partial u(x,t)}{\partial x}|_{x=0} = \pi  e^{-k \pi^2 t}$, and $\frac{\partial u(x,t)}{\partial x}|_{x=1} = - \pi  e^{-k \pi^2 t}$, the analytical solution is given by,
\begin{equation}\label{eqn:solHeat}
    u(x,t) = \sin \left( \pi \frac{x}{L} \right) \exp\left( \frac{-k \pi^2 t}{L^2}\right),
\end{equation}
where $L$ is the size of the heat propagation medium.

\subsubsection{Kuramoto–Sivashinsky Equation}\label{sec:KuSiEq}

The Kuramoto-Sivashinsky Equation is a nonlinear fourth-order partial differential equation applied to the study of many continuous medium physical systems with instabilities and chaotic behavior \cite{Kassam2005,Lakestani2012}. The Kuramoto-Sivashinsky Equation is given by,
\begin{equation}
\frac{\partial u}{\partial t} + u \frac{\partial u}{\partial x} + \beta \frac{\partial^2 u}{\partial x^2} + \gamma \frac{\partial^4 u} {\partial x^4} = 0
\end{equation}
where $\beta$ and $\gamma$ are constants. Here, it was used $\beta = \gamma =1.0$. To solve this equation, it was employed the initial condition $u(x,t=0) = e^{-x^2}$, and the boundary conditions $u(x=-40,t) = u(x=40, t) = 0.0$.

\section{Experiments and Results}\label{sec:ExperimentsResults}

All problems presented in Section \ref{sec:Data} were employed to assess the performance of the MLP-GLN.

\begin{figure}
\centering
\includegraphics[scale=1]{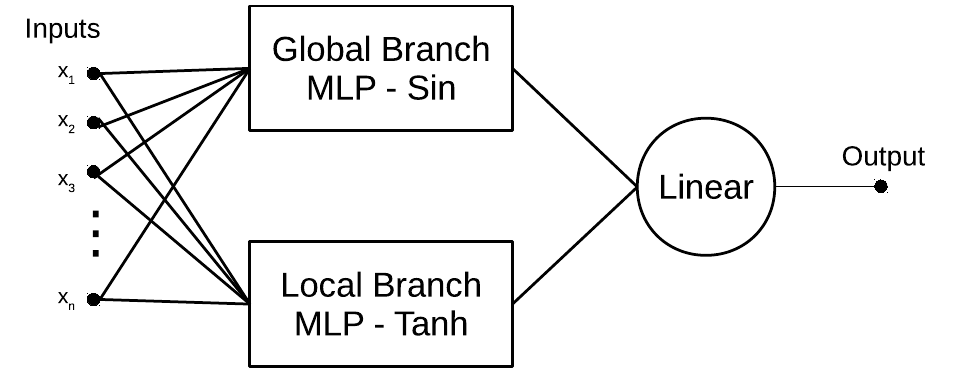}
\caption{The scheme of the Two Branches Network.}\label{fig:TBN}
\end{figure}

Standard MLP networks were also applied to the same benchmark tests for the same conditions used in experiments with the MLP-GLN. Three different ANN models were used to create a comparative analysis. The first architecture was an MLP with the same architecture of the MLP-GLN, but replacing the GLN's double activation function by the sine function (MLP-Sin). The second architecture replaced the GLN's activation function by the hyperbolic tangent function (MLP-Tanh). The third architecture was a hybrid architecture, combining an MLP-Sin branch and an MLP-Tanh branch. The ANN resultant from this combination was called Two Branch Network (TBN), and its scheme can be viewed in Figure~\ref{fig:TBN}. To maintain the same number of hidden neurons for all networks in the TBN network, the hidden layers of each branch have half the number MLP-GLN's hidden neurons. For example, if the MLP-GLN architecture is $1-20-20-1$ (one input, two hidden layers with 20 neurons, and one output), both TBN sine and hyperbolic tangent branches will have two hidden layers with 10 neurons each.  

For each benchmark test, 30 independent experiments were performed with ANN weights randomly initialized. The statistical behaviors of each analyzed ANN were compared.

\subsection{Regression Problem}

To test the regression performance of the proposed GLN, the three data sets presented in Section~\ref{sec:Data} were used. The proposed benchmark is to fit the data set, learning the respective functions in a regression task.

Two different architectures were used for the full conected  MLP-GLN, MLP-Sin, and MLP-Tanh. The first architecture was  $1-20-1$, i.e., one input, one hidden layer with 20 neurons, and one output. The second architecture was $1-20-20-1$, namely, one input, two hidden layers with 20 neurons each, and one output. For the TBN network, each branch had the respective architectures of $1-10-1$ and $1-10-10-1$, corresponding to the same number of hidden neurons. These ANN architectures were defined after the performance of the preliminary experiments to guarantee a minimum sufficient capability to solve the test problems.

Initially, a data repository with 2000 points was created for the EES and SE artificial data sets. For the sunspot data, the data repository is the observations themselves, 320 points. For each data repository, three disjoints sets, training with $50\%$ of the data, validation with $25\%$, and test sets with $25\%$ were created. For the training phase, the mini-batch scheme was employed with a batch size of 64.

Each pattern in the data sets were formed by a pair $\left(x,\mathcal{G}(x) \right)$ (as definided in Section \ref{sec:Def_Regr_Prob}), where $x$ is the ANN input and $\mathcal{G}(\cdot )$ is the target (or ANN wished output) for the input $x$, the respective data set function or data observation. 
 
The computational implementation was done with the pyTorch library (Python3 Programming Language). The training algorithm employed was the Adam algorithm ~\cite{kingma2014adam}, with a learning rate of $10^{-3}$. The stopping criteria were the maximum number of epochs of $200,000$, and validation early stopping, where the training process stops if the validation loss increases more than its minimum value for 30 consecutive epochs. The smallest validation loss ANN was returned as the trained ANN at the end of the training process.  The loss function used to measure the performance was the MSE error between the ANN output $\widehat{\mathcal{G}}(x)$ and the real value of the corresponding data set, $\mathcal{G}(x)$, given by the equation,
\begin{equation}\label{eqn:MSE}
MSE = \frac{1}{N_{batch}}\sum_{i=1}^{N_{batch}} \left( \widehat{\mathcal{G}}(x_i) - \mathcal{G}(x_i) \right)^2
\end{equation}
where $N_{batch}$ is the size of the mini-batch.

\subsubsection{EES Data Set Results}

\begin{figure}[!ht]
     \centering
     \begin{subfigure}[b]{0.47\textwidth}
         \centering
         \includegraphics[width=\textwidth]{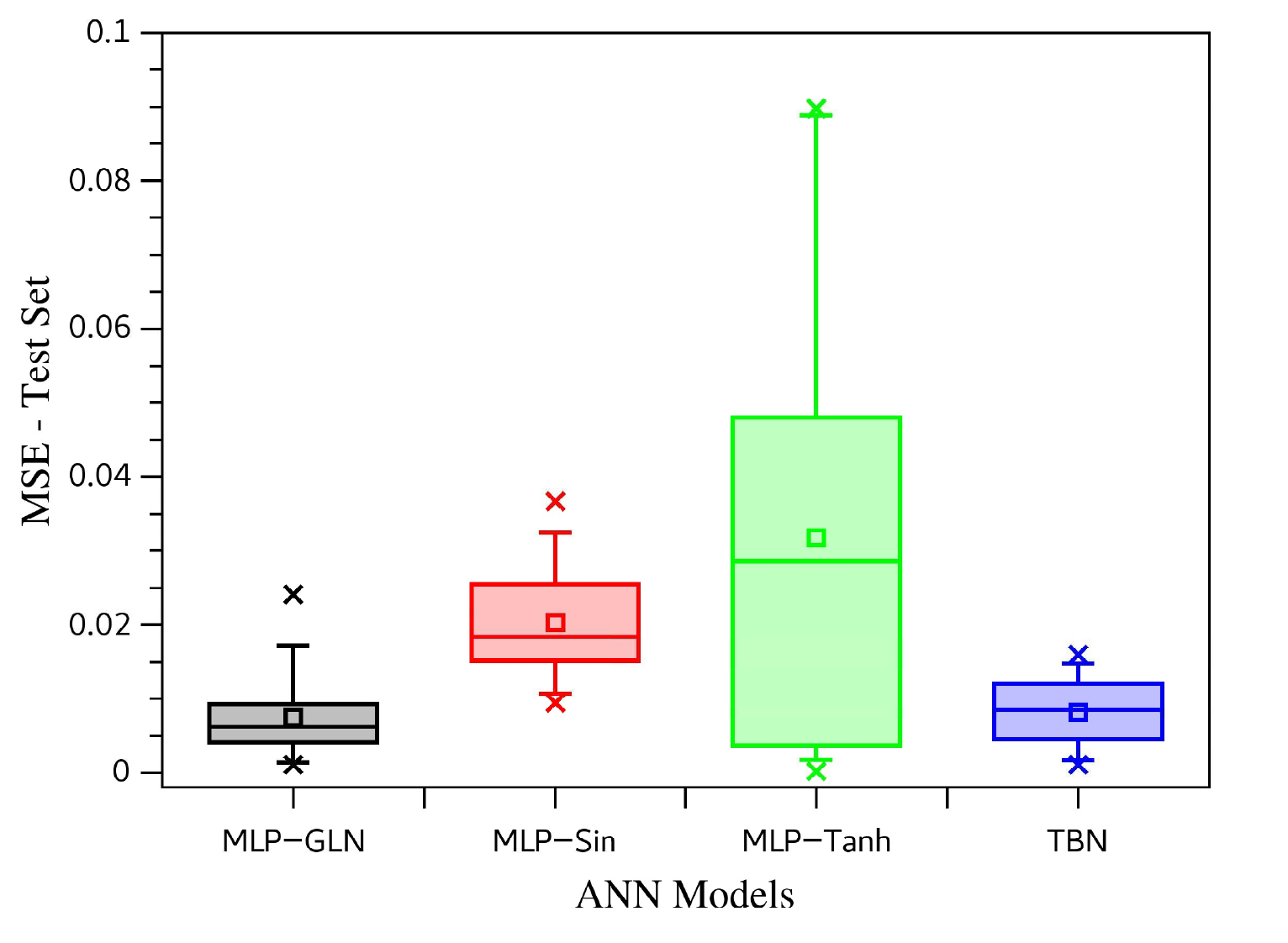}
         \caption{$1-20-1$ Architecture.}
         \label{fig:EESTestMSE_a}
     \end{subfigure}
     \hfill
     \begin{subfigure}[b]{0.47\textwidth}
         \centering
         \includegraphics[width=\textwidth]{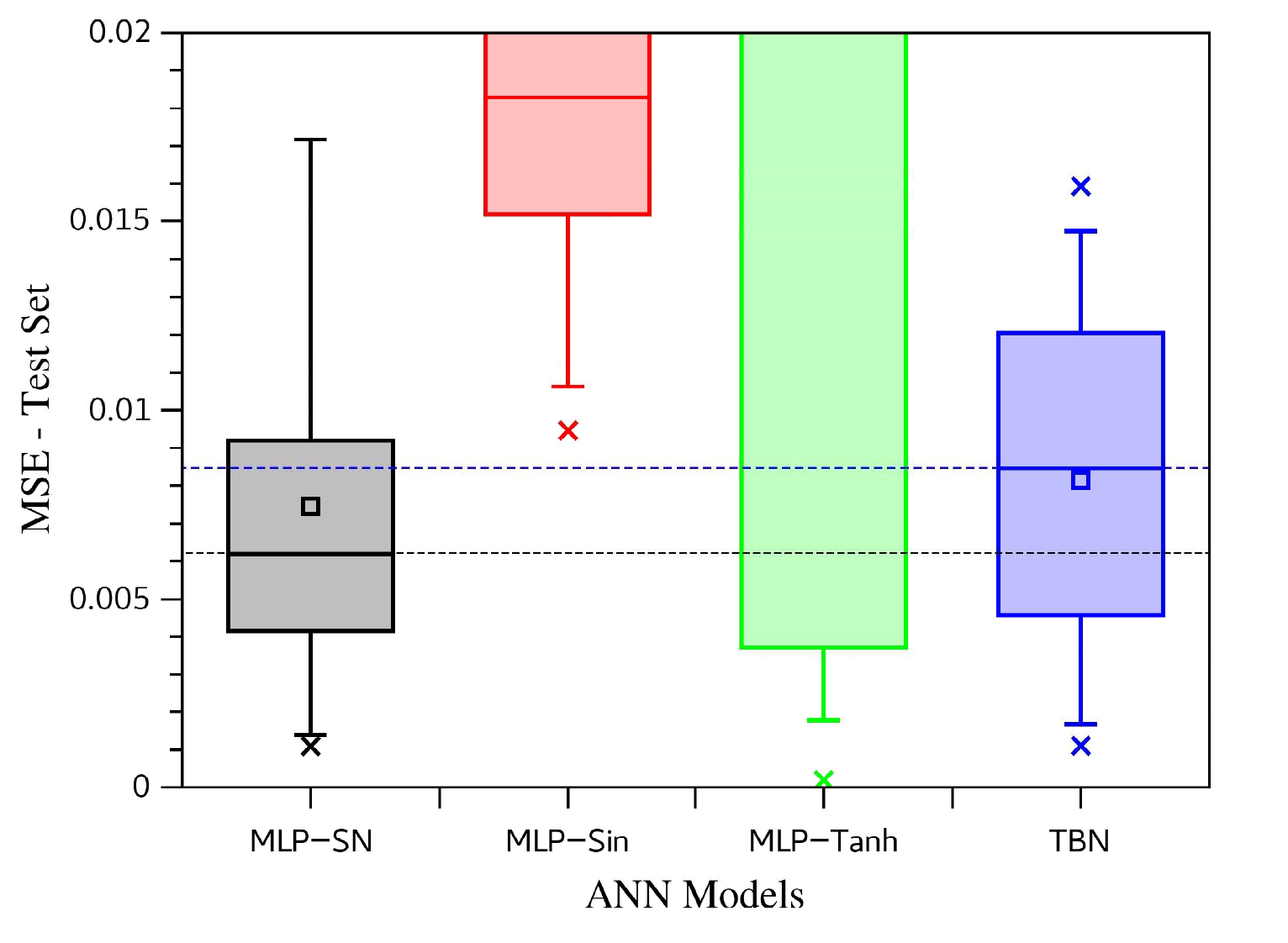}
         \caption{$1-20-1$ Architecture - Zoom scale.}
         \label{fig:EESTestMSE_b}
     \end{subfigure}     
    \hfill \vspace{20pt}    
     \begin{subfigure}[b]{0.47\textwidth}
         \centering
         \includegraphics[width=\textwidth]{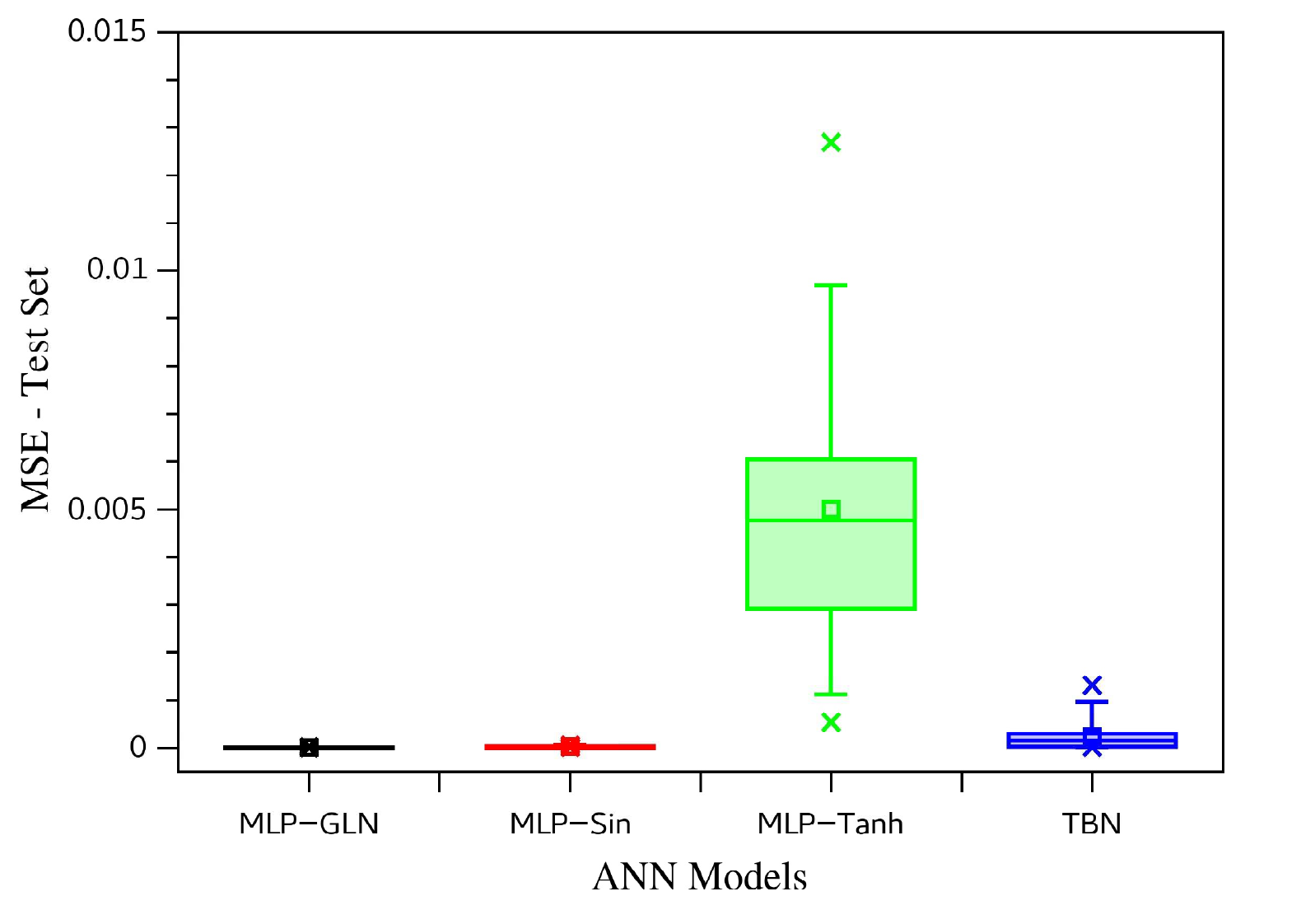}
         \caption{$1-20-20-1$ Architecture.}
         \label{fig:EESTestMSE_c}
     \end{subfigure}
     \hfill
     \begin{subfigure}[b]{0.47\textwidth}
         \centering
         \includegraphics[width=\textwidth]{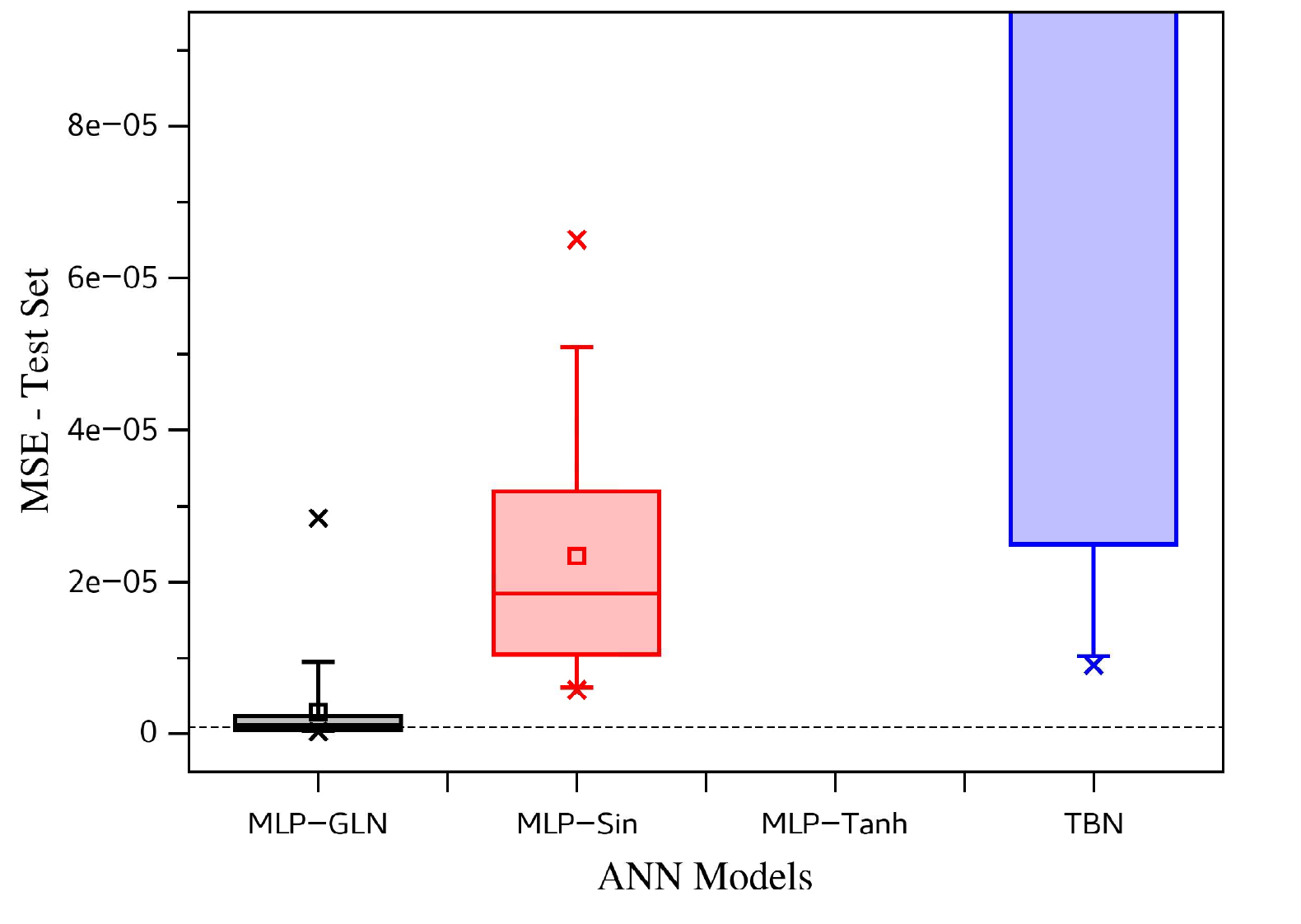}
         \caption{$1-20-20-1$ Architecture - Zoom scale.}
         \label{fig:EESTestMSE_d}
     \end{subfigure}
    \caption{The test set MSE box plot for all ANN models studied and both architectures for EES data set. In (a) and (c) are presented the MSE distributions for the 30 repetitions for each ANN model for architectures $1-20-1$ and $1-20-20-1$. In (b) and (d) are shown a zoom scale to visualize the better performance models for both architectures, where the dashed lines are the median values references.} \label{fig:EESTestMSE}
\end{figure}

\begin{figure}[!ht]
    \centering
     \begin{subfigure}[b]{0.47\textwidth}
         \centering
         \includegraphics[width=\textwidth]{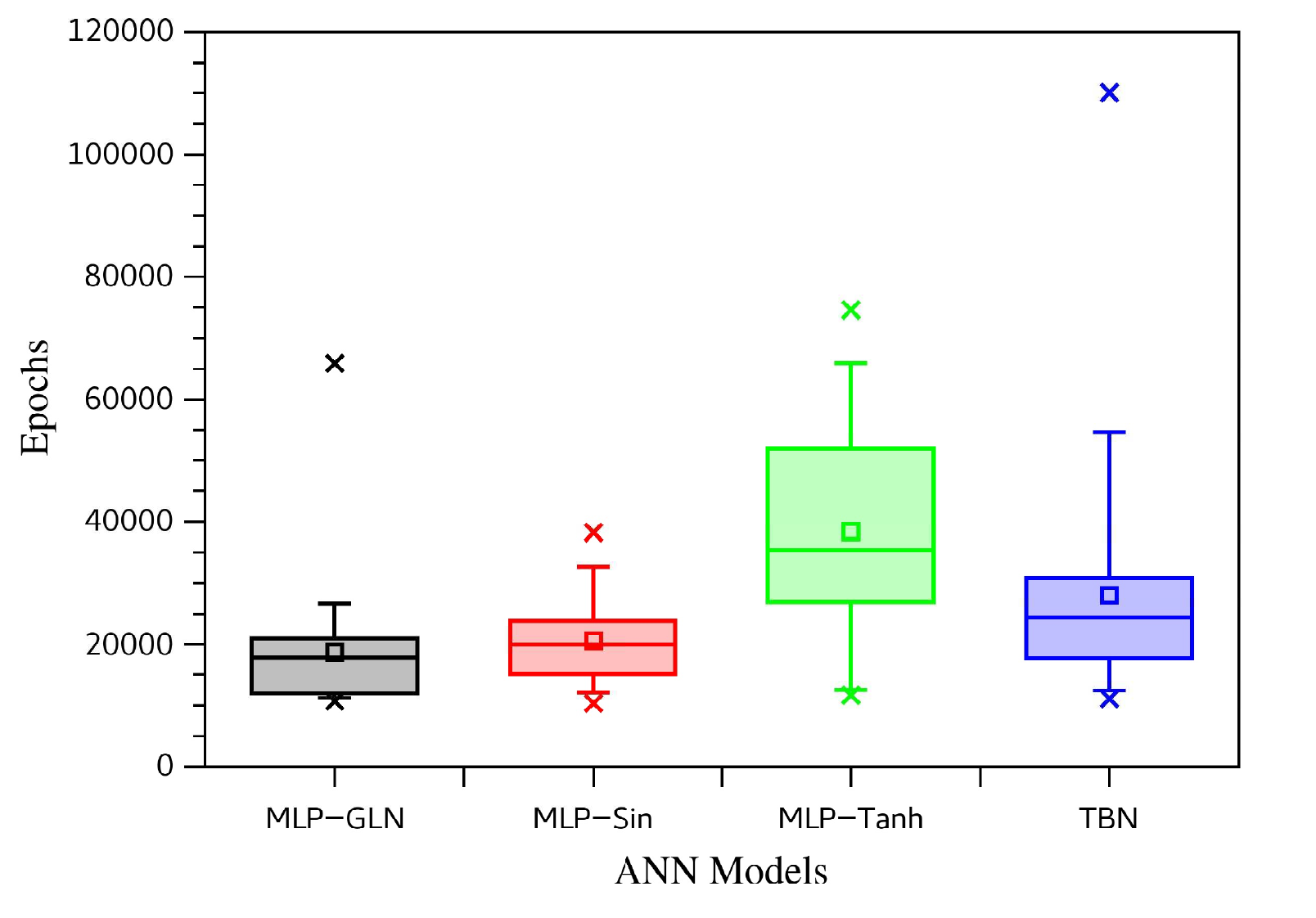}
         \caption{$1-20-1$ Architecture.}
         \label{fig:EES_Epochs_a}
     \end{subfigure}
     \hfill
     \begin{subfigure}[b]{0.47\textwidth}
         \centering
         \includegraphics[width=\textwidth]{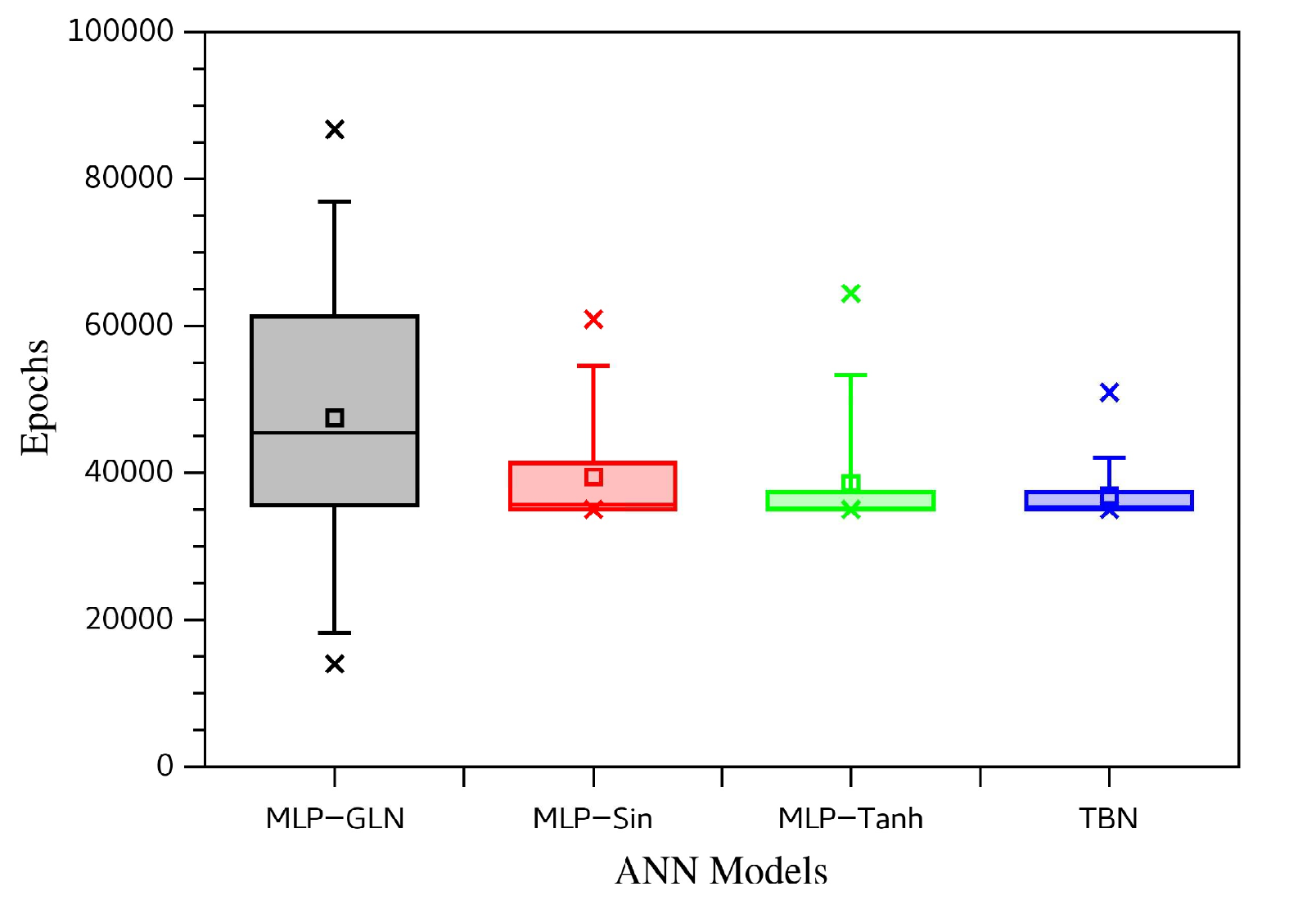}
         \caption{$1-20-20-1$ Architecture.}
         \label{fig:EES_Epochs_b}
     \end{subfigure}  
     \caption{The epoch distributions in the training for all ANN models with the EES Data set. (a) is related to architecture $1-20-1$, and (b) is related to architecture $1-20-20-1$.} \label{fig:EES_Epochs}
\end{figure}

Figure \ref{fig:EESTestMSE} presents the test set MSE distribution for all ANN models studied for the EES data set. Figures \ref{fig:EESTestMSE_a} and \ref{fig:EESTestMSE_b} refer to the architecture $1-20-1$, while Figures \ref{fig:EESTestMSE_c} and \ref{fig:EESTestMSE_d} correspond to the architectures $1-20-20-1$. Figure \ref{fig:EES_Epochs} presents the epoch distributions for the training process, where Figure \ref{fig:EES_Epochs_a} refers to the architecture $1-20-1$ and Figure \ref{fig:EES_Epochs_b} is relative to $1-20-20-1$. Table \ref{tab:EESTestMSE} shows the descriptive statistics for the test set MSE distributions and for the epochs distribution in the training, minimum value observed (Min.), maximum value observed (Max.), mean, median, standard deviation (Std.), and the coefficient of variation (CV). The CV is defined as the ratio between the standard deviation and the mean (CV$=\frac{Std}{Mean}$). Here, the CV can be seen as stability criterion, where smaller the CV, greater the stability.   

Observing Figures \ref{fig:EESTestMSE_a} and \ref{fig:EESTestMSE_b}, and Table \ref{tab:EESTestMSE}, for the experiments with architecture $1-20-1$ the best models were the MLP-GLN and TBN, when looking MSE results for the test set. The nonparametric two-sample Kolmogorov-Smirnov (KS) test~\cite{Chakraborti2003} was applied to confirm (or to refute) that these two ANN models are statistically similar. The Kolmogorov-Smirnov two-sample test is used to decide if two samples come from the same population with a specific distribution. It is a general nonparametric test for comparing two samples. It is sensitive to differences in both location and shape of the two samples' empirical cumulative distribution functions, independent of the distributions. The null hypothesis is that \emph{the MSE distributions of the two models are from the same continuous distribution}. The alternative hypothesis is that \emph{the MSE distributions of the two models are not from the same continuous distribution}. At the $5\%$ significance level, the KS test does not reject the null hypothesis, reaching a $p$-value of $0.5372$. In this way, the models MLP-GLN and TBN with the architecture $1-20-1$ have a statistically similar MSE distribution for the EES data set. Observing the number of trained epochs with architecture $1-20-1$, Figure \ref{fig:EES_Epochs_a} demonstrates an apparent similarity between the models MLP-GLN, MLP-Sin, and TBN. Applying the KS test at the 5\% significance level, the models MLP-GLN and MLP-Sin have the same statistical behavior for the number of training epochs, where the KS test again does not reject the null hypothesis, reaching a $p$-value of $0.3420$. All KS test results for the comparison between the MLP-GLN and all other ANN models for the test set MSE distribution and epochs distribution are presented in Table~\ref{tab:KS_EES}.

\begin{table}[!ht]
    \centering
     \caption{The descriptive statistics for all ANN models. All MSE measures are relative to the first artificial data EES test set. The epochs measures are referent to the training process. The best results are highlighted in bold-face.}
    \label{tab:EESTestMSE}
    \begin{tabular}{ccccccc}
    \hline
    \multicolumn{3}{c}{\multirow{2}{*}{\textbf{Statistics}}} & \multicolumn{4}{c}{\textbf{ANN Models}}\\ 
    \cline{4-7}
     & & & MLP-GLN & MLP-Sin & MLP-Tanh & TBN \\
    \hline
    \multirow{12}{*}{\rotatebox[origin=c]{90}{$1-20-1$}} &\multirow{6}{*}{\rotatebox[origin=c]{90}{MSE}}
    & Min. & $1.095\cdot 10^{-3}$ & $9.463\cdot 10^{-3}$ & $\mathbf{1.968 \cdot 10^{-4}}$& $1.112\cdot 10^{-3}$\\
    & & Max. & $2.414\cdot 10^{-2}$ & $3.665\cdot 10^{-2}$ & $8.975\cdot 10^{-2}$& $\mathbf{1.594\cdot 10^{-2}}$\\
    & & Mean & $\mathbf{7.455\cdot 10^{-3}}$   & $2.031\cdot 10^{-2}$ & $3.175\cdot 10^{-2}$ & $8.147 \cdot 10^{-3}$\\
    & & Median & $\mathbf{6.181\cdot 10^{-3}}$ & $1.827 \cdot 10^{-2}$ & $2.851\cdot 10^{-2}$ & $8.446\cdot 10^{-3}$\\
    & & Std. & $\mathbf{5.291\cdot 10^{-3}}$   & $7.204\cdot 10^{-3}$ & $2.884 \cdot 10^{-2}$ & $4.482\cdot 10^{-2}$\\
    & & CV & $0.710$ & $\mathbf{0.355}$ & $0.908$ & $0.550$\\
    \cline{2-7}
    &\multirow{6}{*}{\rotatebox[origin=c]{90}{Epochs}}
    & Min. & $10697$ & $\mathbf{10395}$ & $11676$ & $11065$ \\
    & & Max. & $65850$ & $\mathbf{38253}$ & $74557$ & $110055$\\
    & & Mean & $\mathbf{18732.000}$ & $20506.667$& $38452.000$ & $27992.933$ \\
    & & Median & $\mathbf{1777.050}$ & $19836.000$ & $35336.500$ & $24343.500$ \\
    & & Std. & $10126.650$ & $6943.948$ & $\mathbf{16580.554}$ & $19106.861$  \\
    & & CV & $0.541$ & $\mathbf{0.339}$ & $0.431$ & $0.683$ \\
    \hline
    \multirow{12}{*}{\rotatebox[origin=c]{90}{$1-20-20-1$}} &\multirow{6}{*}{\rotatebox[origin=c]{90}{MSE}} 
    & Min. & $\mathbf{2.753\cdot10^{-7}}$& $5.792\cdot10^{-6}$ & $5.373\cdot 10^{-4}$& $9.025\cdot10^{-6}$\\
    & & Max. & $\mathbf{2.845\cdot 10^{-5}}$& $6.507\cdot10^{-5}$& $1.270\cdot 10^{-2}$& $1.317\cdot 10^{-3}$\\
    & & Mean & $\mathbf{2.862\cdot 10^{-6}}$ & $2.339\cdot 10^{-5}$& $4.997\cdot 10^{-3}$ & $2.372\cdot 10^{-4}$\\
    & & Median & $\mathbf{1.039\cdot10^{-6}}$& $1.849\cdot10^{-5}$& $4.764\cdot 10^{-3}$& $1.442\cdot10^{-4}$\\
    & & Std. & $\mathbf{5.410\cdot 10^{-6}}$& $1.587\cdot10^{-5}$& $2.838\cdot 10^{-3}$ & $3.241\cdot 10^{-4}$\\
    & & CV & $1.890$ & $0.679$ & $\mathbf{0.568}$ & $1.367$\\
    \cline{2-7}
    &\multirow{6}{*}{\rotatebox[origin=c]{90}{Epochs}}
    & Min. & $\mathbf{13997}$ & $35002$ & $35005$ & $35001$ \\
    & & Max. & $86719$ & $60869$ & $64465$ & $\mathbf{50956}$ \\
    & & Mean & $47492.633$ & $39489.867$ & $38569.300$ & $\mathbf{36867.633}$  \\
    & & Median & $45434.000$ & $35611.500$ & $\mathbf{35156.500}$ & $35340.000$  \\
    & & Std. & $18955.527$ & $6960.878$ & $7149.699$ & $\mathbf{3338.719}$ \\
    & & CV & $0.399$ & $0.176$ & $0.185$ & $\mathbf{0.091}$ \\
    \hline
    \end{tabular}
\end{table}

\begin{table}[!ht]
    \centering
    \caption{Two-sample Kolmogorov-Smirnov Test at the $5\%$ significance level for the MSE and number of epochs distributions in the training between the MLP-GLN and all other models for both architectures studied -- EES data set.}
    \label{tab:KS_EES}
    \begin{tabular}{ccccc}
    \hline
    \multicolumn{3}{c}{\textbf{Tested Model}} & \multicolumn{2}{c}{\textbf{KS Test Results}}\\ 
    \cline{4-5}
    \multicolumn{3}{c}{\textbf{whit MLP-GLN}} & Statistically Similar & $p$-values\\
    \hline
    \multirow{6}{*}{\rotatebox[origin=c]{90}{$1-20-1$}} &\multirow{3}{*}{\rotatebox[origin=c]{90}{MSE}}
    & MLP-Sin  & No & $1.1088\cdot10^{-8}$ \\
    & & MLP-Tanh & No & $6.1740\cdot10^{-5}$ \\
    & & TBN      & Yes & $0.5372$ \\
    \cline{2-5}
    &\multirow{3}{*}{\rotatebox[origin=c]{90}{Epochs}}
    & MLP-Sin  & Yes & $0.3420$ \\
    & & MLP-Tanh & No & $1.1088\cdot10^{-8}$ \\
    & & TBN      & No & $0.0046$ \\
    \hline
    \multirow{6}{*}{\rotatebox[origin=c]{90}{$1-20-20-1$}} &\multirow{3}{*}{\rotatebox[origin=c]{90}{MSE}}
    & MLP-Sin  & No & $8.3842\cdot10^{-12}$ \\
    & & MLP-Tanh & No & $1.7965\cdot10^{-14}$ \\
    & & TBN      & No & $1.1620\cdot10^{-12}$ \\
    \cline{2-5}
    &\multirow{3}{*}{\rotatebox[origin=c]{90}{Epochs}}
    & MLP-Sin  & No & $0.0257$ \\
    & & MLP-Tanh & No & $0.0017$ \\
    & & TBN      & No & $6.1740\cdot 10^{-5}$ \\
    \hline
    \end{tabular}
    
\end{table}

For the experiments with the architecture $1-20-1$, the MLP-GLN together with the TBN model reached the best regression performance. The MLP-GLN was more efficient than the TBN model with respect to the expected number of epochs to train the model. In this way, the MLP-GLN model was the better option among those tested. 

\begin{figure}[!ht]
    \centering
     \begin{subfigure}[b]{0.47\textwidth}
         \centering
         \includegraphics[width=\textwidth]{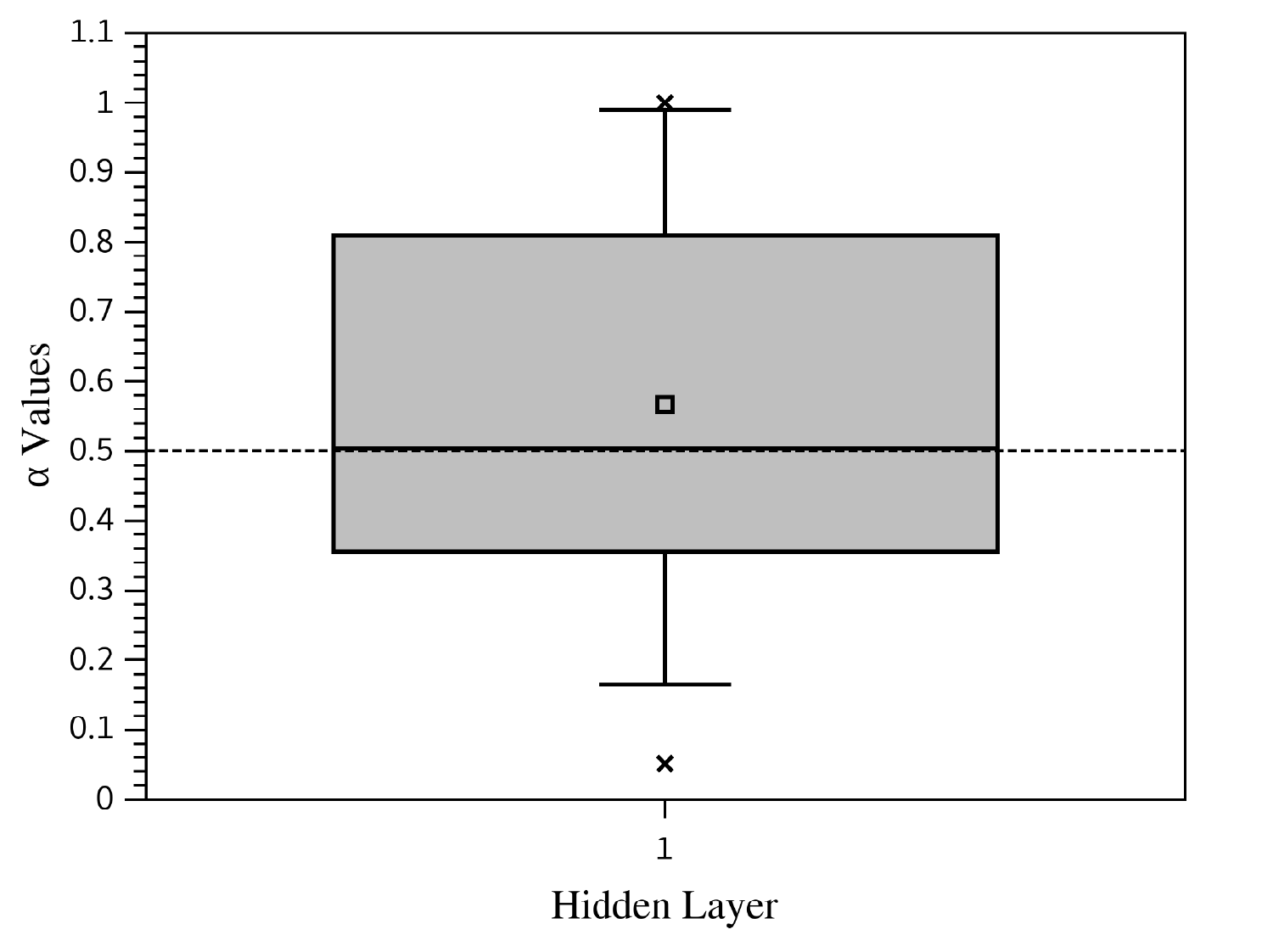}
         \caption{$1-20-1$ Architecture.}
         \label{fig:EES_alphas_a}
     \end{subfigure}
     \hfill
     \begin{subfigure}[b]{0.47\textwidth}
         \centering
         \includegraphics[width=\textwidth]{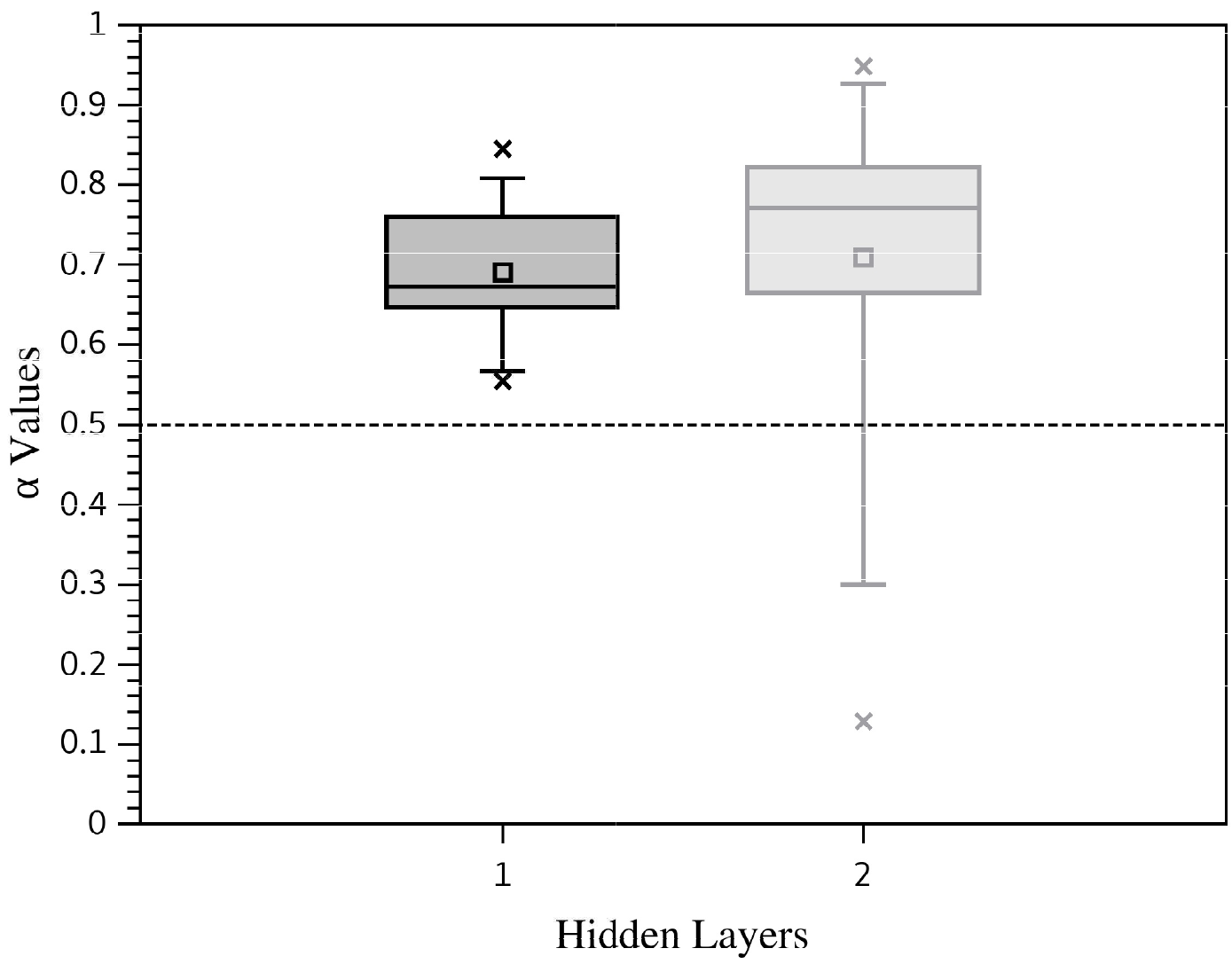}
         \caption{$1-20-20-1$ Architecture.}
         \label{fig:EES_alphas_b}
     \end{subfigure}
     \caption{The $\alpha$ distribution for the MLP-GLN models after the training process -- EES data ser.}
     \label{fig:EES_alphas}
\end{figure}

Analyzing the MLP-GLN $\alpha$ values, defined in Equation \ref{eqn:actFunc}, it is possible to verify the importance given to each activation function component, where $\alpha = 1$ implies a purely global activation function, and $\alpha = 0$ a purely local. The distribution of the $\alpha$ weight values can be viewed in Figure \ref{fig:EES_alphas_a}. With a mean value of $ \alpha $ of $0.566$ and a median of $0.504$, the MLP-GLN on average gives a tiny bit more importance for the global (sine function) than the local (hyperbolic tangent) components for this data set. Here, the EES data set has a global component (sine function) perturbed by a local effect caused by the Gaussian function, where apparently the global behavior is dominant (see Figure \ref{fig:EES}). However, the perturbation created by the Gaussian function in the sine is sufficient to demand a strong importance of the local component in the activation function composition.

For the experiments with two hidden layers (architecture $1-20-20-1$) there will two effect. One is the combination of the global and local components in the hidden layer, and the other is the composition of the global and local activation function of the two hidden layers. For this architecture, the test set MSE distributions are shown in Figures \ref{fig:EESTestMSE_c} and \ref{fig:EESTestMSE_d}, with the descriptive statistics in Table \ref{tab:EESTestMSE}. In general, the MLP-GLN reached a better relative test set MSE performance when compared with other analyzed models, but with the biggest CV value. Higher the CV value, greater the relative dispersion, indicating that the MLP-GLN has the smallest relative stability. However, the worse MSE value reached by the MLP-GLN, the maximum MSE value, is less than all other models' maximum MSE values. 

The MLP-GLN had the smallest MSE value but it required the largest number of training epochs, as it is shown by Figure \ref{fig:EES_Epochs_b} and Table \ref{tab:EESTestMSE}. Comparing the models MLP-GLN and MLP-Sin (the second-best model for the test set MSE performance), the mean number of training epochs of the MLP-GLN is $20\%$ higher than the MLP-Sin epochs number. However, observing the MSE performance, the MLP-GLN mean MSE distribution is approximately ten times less than the MLP-Sin mean MSE distribution, which may justify the choice of the MLP-GLN model to the detriment of the MLP-Sin model. The $\alpha$ values distribution for the two hidden layers are shown in Figure \ref{fig:EES_alphas_b}. In general, for both hidden layers, the MLP-GLN gives more importance to the global component than the local component of its activation functions. For the first hidden layer, the mean $\alpha$ value was $0.691$ and the median of $0.673$. And for the second hidden layer, a mean $\alpha$ value of $0.709$ and the median of $0.771$. These $\alpha$ values demonstrate the global behavior of the EES data set.

The KS test was also applied to the $1-20-20-1$ experiments, demonstrating that the MSE distribution and epochs number distribution of the MLP-GLN are not statistically similar to other ANN models at a $5\%$ significance level. The results of the KS test can be viewed in Table~\ref{tab:KS_EES}.

\subsubsection{SE Data Set Results}

\begin{figure}[!ht]
     \centering
     \begin{subfigure}[b]{0.47\textwidth}
         \centering
         \includegraphics[width=\textwidth]{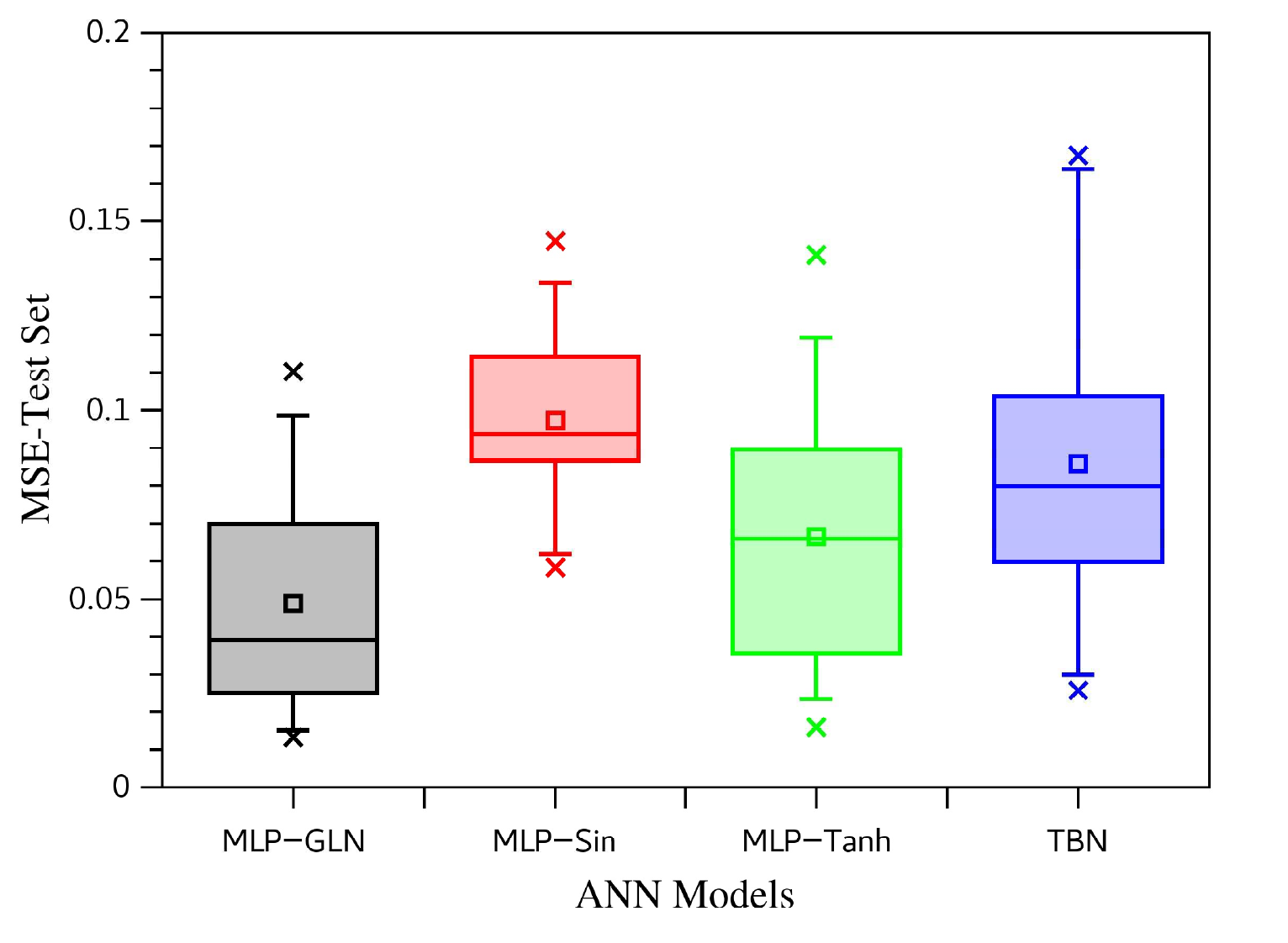}
         \caption{$1-20-1$ Architecture.}
         \label{fig:SETestMSE_a}
     \end{subfigure}
     \\
     \vspace{20pt}    
     \begin{subfigure}[b]{0.47\textwidth}
         \centering
         \includegraphics[width=\textwidth]{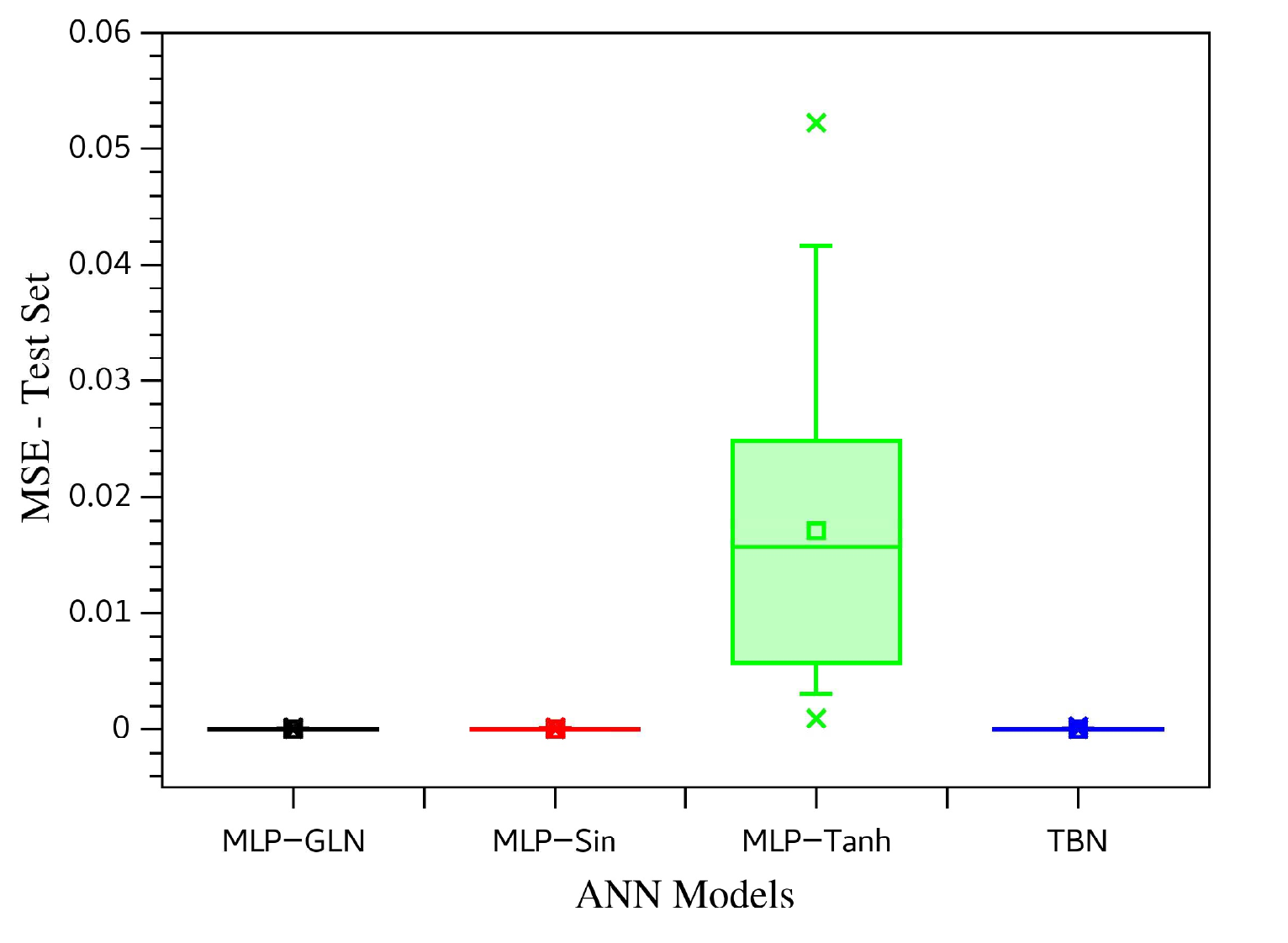}
         \caption{$1-20-20-1$ Architecture.}
         \label{fig:SETestMSE_b}
     \end{subfigure}
     \hfill
     \begin{subfigure}[b]{0.47\textwidth}
         \centering
         \includegraphics[width=\textwidth]{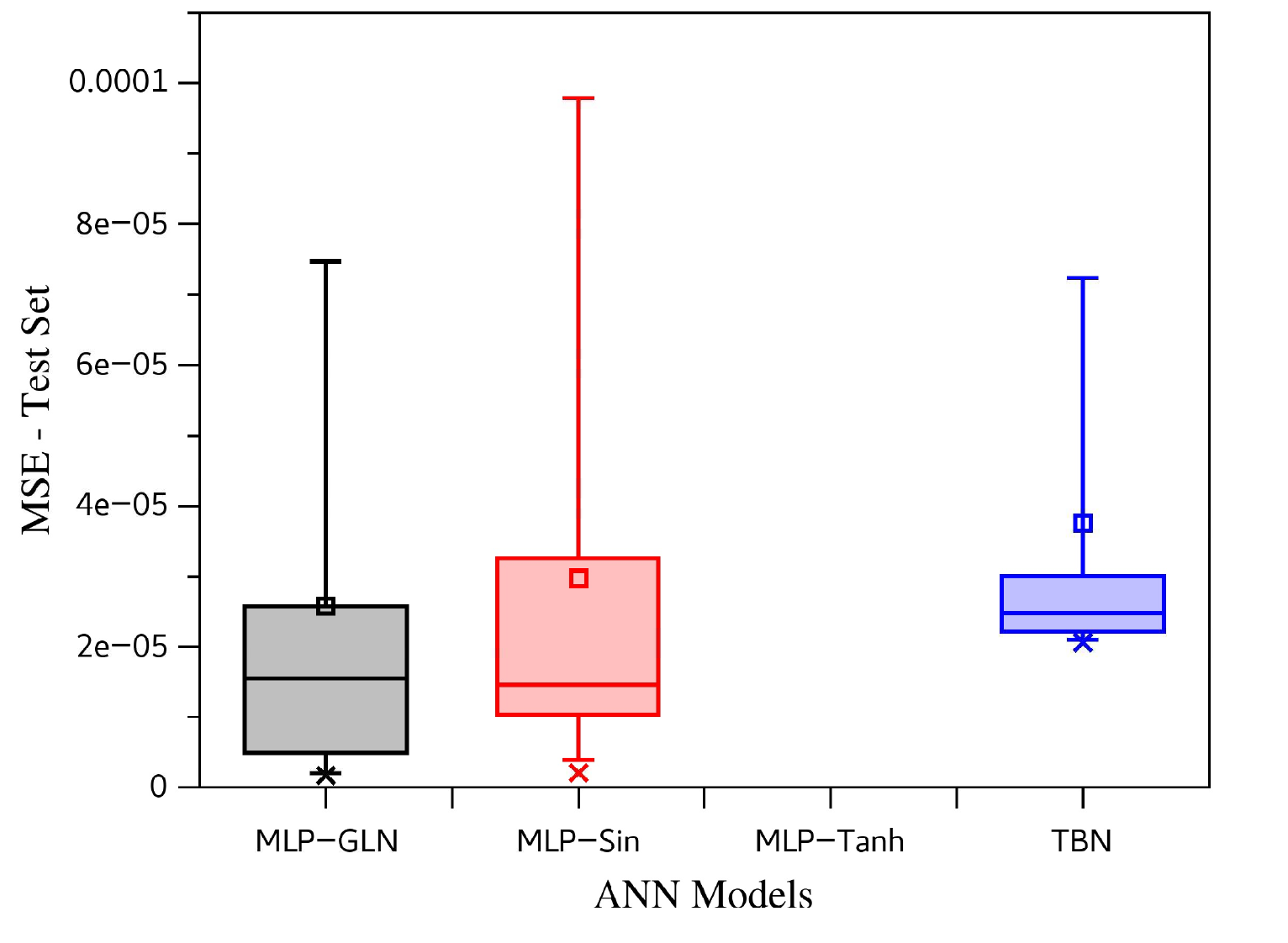}
         \caption{$1-20-20-1$ Architecture - Zoom scale.}
         \label{fig:SETestMSE_c}
     \end{subfigure}
    \caption{The test set MSE box plot for all ANN models studied and both architectures for the SE data set. In (a) are presented the MSE distributions (30 repetitions) for each ANN model for architecture $1-20-1$. In (b) and (c) are shown the results for architecture $1-20-20-1$ in normal and zoom scale, respectively.} \label{fig:SETestMSE}
\end{figure}

The same experimental procedure applied to the EES data set was also employed for the SE data set. The test set  MSE distributions for all ANN models are outlined in Figure \ref{fig:SETestMSE}. Figure \ref{fig:SETestMSE_a} presents the MSE distributions for the architecture with one hidden layer ($1-20-1$), where we observe that the MLP-GLN model performs slightly better than the other models. Table \ref{tab:SETestMSE} shows the descriptive statistics of these MSE distributions, where for the architecture $1-20-1$ the MLP-GLN has an MSE mean value of $25\%$ smaller than the second best model, MLP-Tanh. Moreover, the MSE median of the MLP-GLN is more than $40\%$ more efficient than MLP-Tanh. However, applying the two-sample KS test at the 5\% significance level, the MSE distributions of the MLP-GLN and MLP-Tanh are statistically similar, as presented in Table \ref{tab:KS_SE}. The other two ANN models are not statistically similar to the MLP-GLN. 

\begin{table}[!ht]
    \centering
     \caption{The descriptive statistics for all ANN models analyzed. All MSE measures are relative to the test set of the SE data set. The epochs measures are referent to the training process. The best results are highlighted in bold-face.}\label{tab:SETestMSE}
    \begin{tabular}{ccccccc}
    \hline
    \multicolumn{3}{c}{\multirow{2}{*}{\textbf{Statistics}}} & \multicolumn{4}{c}{\textbf{ANN Models}}\\ 
    \cline{4-7}
     & & & MLP-GLN & MLP-Sin & MLP-Tanh & TBN \\
    \hline
    \multirow{12}{*}{\rotatebox[origin=c]{90}{$1-20-1$}} &\multirow{6}{*}{\rotatebox[origin=c]{90}{MSE}}
    & Min.     & $\mathbf{0.013}$ & $0.058$ & $0.016$ & $0.026$ \\
    & & Max.   & $\mathbf{0.110}$ & $0.145$ & $0.141$ & $0.167$ \\
    & & Mean   & $\mathbf{0.049}$ & $0.097$ & $0.066$ & $0.086$ \\
    & & Median & $\mathbf{0.039}$ & $0.094$ & $0.066$ & $0.080$ \\
    & & Std.   & $0.029$ & $\mathbf{0.022}$ & $0.033$ & $0.041$ \\
    & & CV     & $0.600$ & $\mathbf{0.230}$ & $0.492$ & $0.482$ \\
    \cline{2-7}
    &\multirow{6}{*}{\rotatebox[origin=c]{90}{Epochs}}
    & Min.     & $10061$ & $10319$ & $\mathbf{10021}$ & $10056$ \\
    & & Max.   & $\mathbf{23695}$ & $36469$ & $37761$ & $33832$ \\
    & & Mean   & $\mathbf{15150.833}$ & $17871.833$ & $20754.133$ & $17071.167$ \\
    & & Median & $\mathbf{14419.500}$ & $16384.000$ & $20803.000$ & $15950.5$ \\
    & & Std.   & $\mathbf{3599.143}$ & $6004.269$ & $7791.903$ & $5560.436$ \\
    & & CV     & $\mathbf{0.238}$ & $0.336$ & $0.375$ & $0.326$ \\
    \hline
    \multirow{12}{*}{\rotatebox[origin=c]{90}{$1-20-20-1$}} &\multirow{6}{*}{\rotatebox[origin=c]{90}{MSE}} 
    & Min.     & $\mathbf{1.702\cdot10^{-6}}$ & $2.074\cdot10^{-6}$ & $9.192\cdot10^{-4}$ & $2.057\cdot10^{-5}$ \\
    & & Max.   & $2.189\cdot10^{-4}$ & $\mathbf{1.814\cdot10^{-4}}$ & $0.052$ & $2.820\cdot10^{-4}$ \\
    & & Mean   & $\mathbf{2.582\cdot10^{-5}}$ & $2.971\cdot10^{-5}$ & $0.017$ & $3.757\cdot10^{-5}$ \\
    & & Median & $1.551\cdot10^{-5}$ & $\mathbf{1.465\cdot10^{-5}}$ & $0.016$ & $2.476\cdot10^{-5}$ \\
    & & Std.   & $4.122\cdot10^{-5}$ & $\mathbf{3.806\cdot10^{-5}}$ & $0.013$ & $4.802\cdot10^{-5}$ \\
    & & CV     & $1.597$ & $1.281$ & $\mathbf{0.756}$ & $1.278$ \\
    \cline{2-7}
    &\multirow{6}{*}{\rotatebox[origin=c]{90}{Epochs}}
    & Min.     & $23100$ & $31700$ & $\mathbf{10453}$ & $19354$ \\
    & & Max.   & $111138$ & $112977$ & $\mathbf{65627}$ & $118763$ \\
    & & Mean   & $56181.233$ & $62887.567$ & $\mathbf{31380.367}$ & $71448.700$ \\
    & & Median & $56435.000$ & $59996.000$ & $\mathbf{28362.500}$ & $73318.000$ \\
    & & Std.   & $17737.494$ & $21061.202$ & $\mathbf{16553.550}$ & $28260.347$ \\
    & & CV     & $\mathbf{0.316}$ & $0.335$ & $0.528$ & $0.396$ \\
    \hline
    \end{tabular}
\end{table}

\begin{figure}[!th]
    \centering
     \begin{subfigure}[b]{0.47\textwidth}
         \centering
         \includegraphics[width=\textwidth]{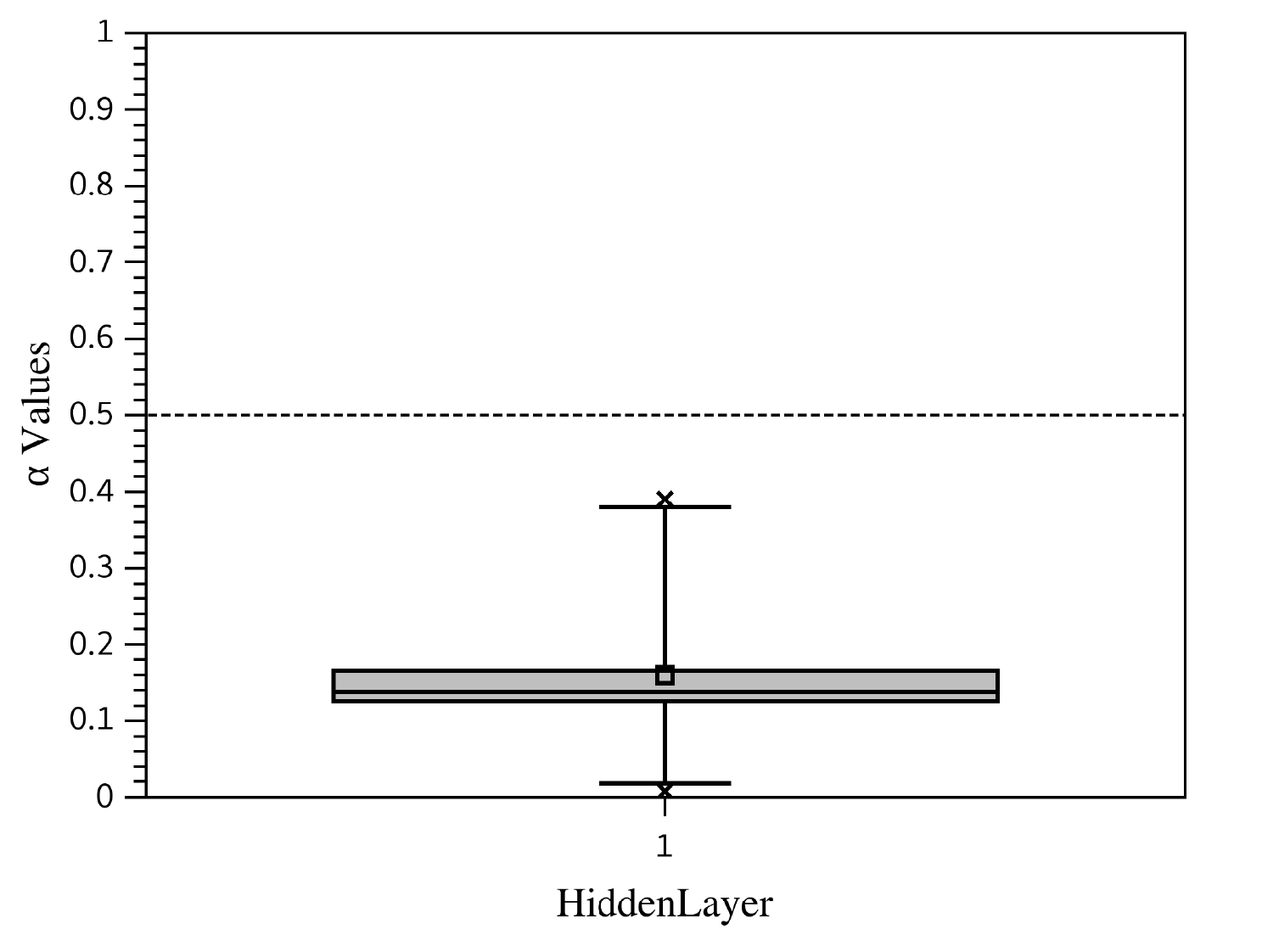}
         \caption{$1-20-1$ Architecture.}
         \label{fig:SE_alphas_a}
     \end{subfigure}
     \hfill
     \begin{subfigure}[b]{0.47\textwidth}
         \centering
         \includegraphics[width=\textwidth]{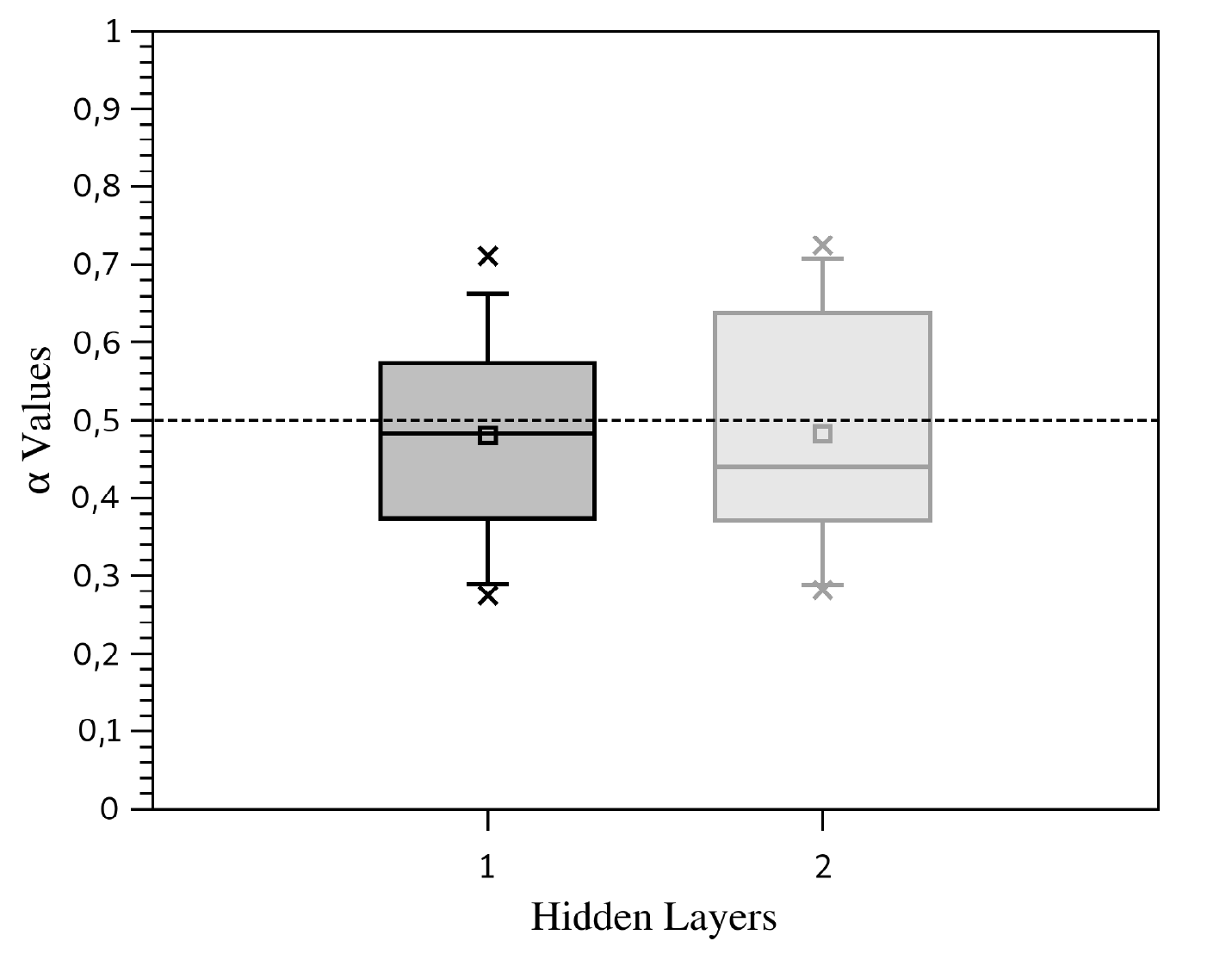}
         \caption{$1-20-20-1$ Architecture.}
         \label{fig:SE_alphas_b}
     \end{subfigure}
     \caption{The $\alpha$ distributions for the MLP-GLN models after the training process -- SE data set.}
     \label{fig:SE_alphas}
\end{figure}

Observing Figure \ref{fig:SETestMSE_a}, it is possible to note that the MLP-Tanh had a better MSE performance than the MLP-Sin, and the TBN had an intermediary MSE performance between the MLP-Sin and MLP-Tanh. In this way, the TBN can be viewed as a mean between the MLP-Sin and MLP-Tanh, where its dispersion will be something equivalent to the sum of the two models' dispersion. The MLP-GLN gave more importance to the local component of its activation function, as shown in Figure \ref{fig:SE_alphas_a}. In this way, the MLP-GLN managed to do a more efficient combination of the sine and hyperbolic tangent functions than the TBN.

\begin{figure}[!htb]
    \centering
     \begin{subfigure}[b]{0.47\textwidth}
         \centering
         \includegraphics[width=\textwidth]{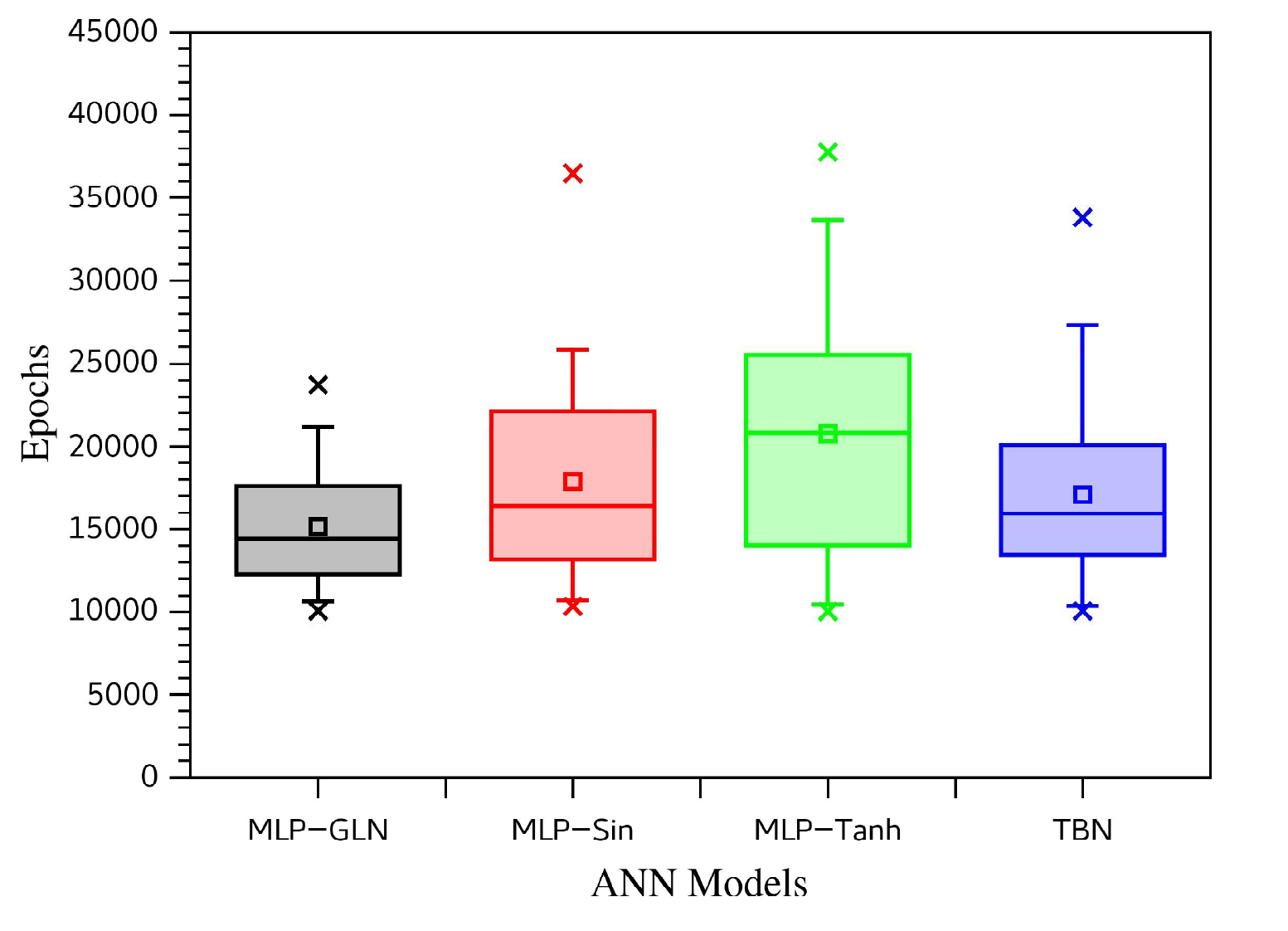}
         \caption{$1-20-1$ Architecture.}
         \label{fig:SE_Epochs_a}
     \end{subfigure}
     \hfill
     \begin{subfigure}[b]{0.47\textwidth}
         \centering
         \includegraphics[width=\textwidth]{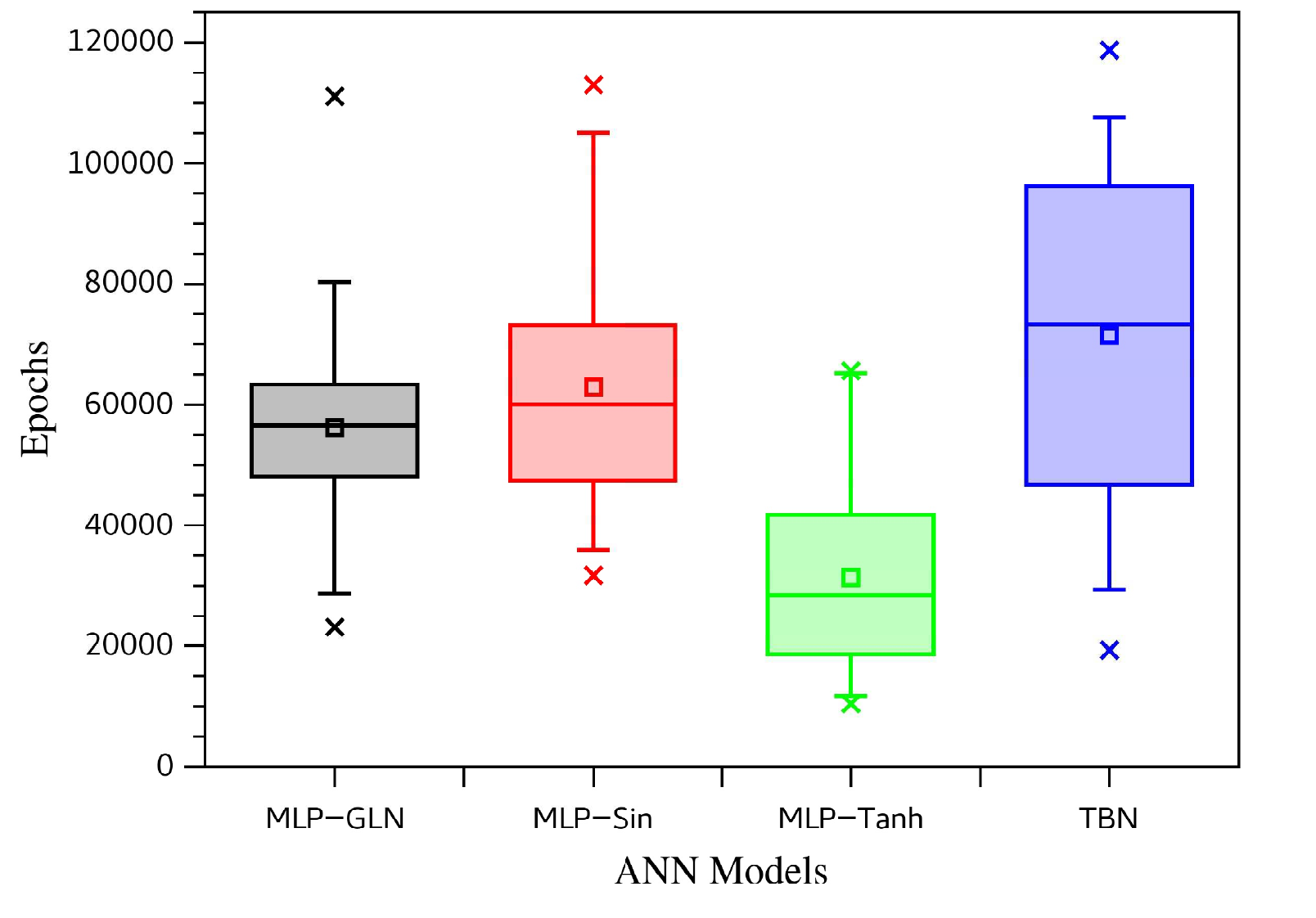}
         \caption{$1-20-20-1$ Architecture.}
         \label{fig:SE_Epochs_b}
     \end{subfigure}  
     \caption{The epoch distributions in the training for all ANN models for the SE data set. (a) is related to architecture $1-20-1$, and (b) is related to architecture $1-20-20-1$.} \label{fig:SE_Epochs}
\end{figure}

The distribution of the number of epochs for each ANN model (architecture $1-20-1$) is shown in Figure \ref{fig:SE_Epochs_a}. The MLP-GLN is more efficient than other models when looking at the measure of central tendency, mean, and median. Besides these measures of the central location, the MLP-GLN also has the lowest CV, indicating that the MLP-GLN has the lowest dispersion for the number of trained epochs.

For the architecture with two hidden layers, $1-20-20-1$, the MLP-GLN and the MLP-Sin are statistically similar as for the MSE performance as for the number of epochs trained, according to Figure \ref{fig:SETestMSE_c} and the KS test presented in Table \ref{tab:KS_EES}. In particular, for the number of trained epochs, the most efficient ANN is the MLP-Tanh. However, the MSE performance reached by the MLP-Tanh is the worst among the analyzed models. In this way, the MLP-GLN and MLP-Sin need approximately twice the MLP-Tanh's number of epochs to be trained, but the MSE reached by these two models is likely $10^3$ more accurate than the MLP-Tanh.

\begin{table}[!ht]
    \centering
    \caption{Two-sample Kolmogorov-Smirnov Test at the $5\%$ significance level for the MSE and number of epochs distributions between the MLP-GLN and all other models for both architectures studied -- SE data set.}
    \label{tab:KS_SE}
    \begin{tabular}{ccccc}
    \hline
    \multicolumn{3}{c}{\textbf{Tested Model}} & \multicolumn{2}{c}{\textbf{KS Test Results}}\\ 
    \cline{4-5}
    \multicolumn{3}{c}{\textbf{whit MLP-GLN}} & Statistically Similar & $p$-values\\
    \hline
    \multirow{6}{*}{\rotatebox[origin=c]{90}{$1-20-1$}} &\multirow{3}{*}{\rotatebox[origin=c]{90}{MSE}}
    & MLP-Sin    & No & $1.143\cdot10^{-6}$ \\
    & & MLP-Tanh & Yes & $0.2003$ \\
    & & TBN      & No & $0.0046$ \\
    \cline{2-5}
    &\multirow{3}{*}{\rotatebox[origin=c]{90}{Epochs}}
    & MLP-Sin    & Yes & $0.2003$ \\
    & & MLP-Tanh & No & $0.0046$ \\
    & & TBN      & Yes & $0.7600$ \\
    \hline
    \multirow{6}{*}{\rotatebox[origin=c]{90}{$1-20-20-1$}} &\multirow{3}{*}{\rotatebox[origin=c]{90}{MSE}}
    & MLP-Sin    & Yes & $0.5372$ \\
    & & MLP-Tanh & No & $1.797\cdot10^{-14}$ \\
    & & TBN      & No & $4.644\cdot10^{-6}$ \\
    \cline{2-5}
    &\multirow{3}{*}{\rotatebox[origin=c]{90}{Epochs}}
    & MLP-Sin    & Yes & $0.5372$ \\
    & & MLP-Tanh & No & $1.755\cdot10^{-5}$ \\
    & & TBN      & No & $0.0046$ \\
    \hline
    \end{tabular}
    
\end{table}

The $\alpha$ weight distribution for the MLP-GLN with two hidden layers is shown  in Figure \ref{fig:SE_alphas_b}. The MLP-GLN gave approximately the same importance for both global and local components of its activation function, but with a slight dominance of the local component. 
 
\subsubsection{Sunspot Data set}

\begin{figure}[!htb]
     \centering
     \begin{subfigure}[b]{0.47\textwidth}
         \centering
         \includegraphics[width=\textwidth]{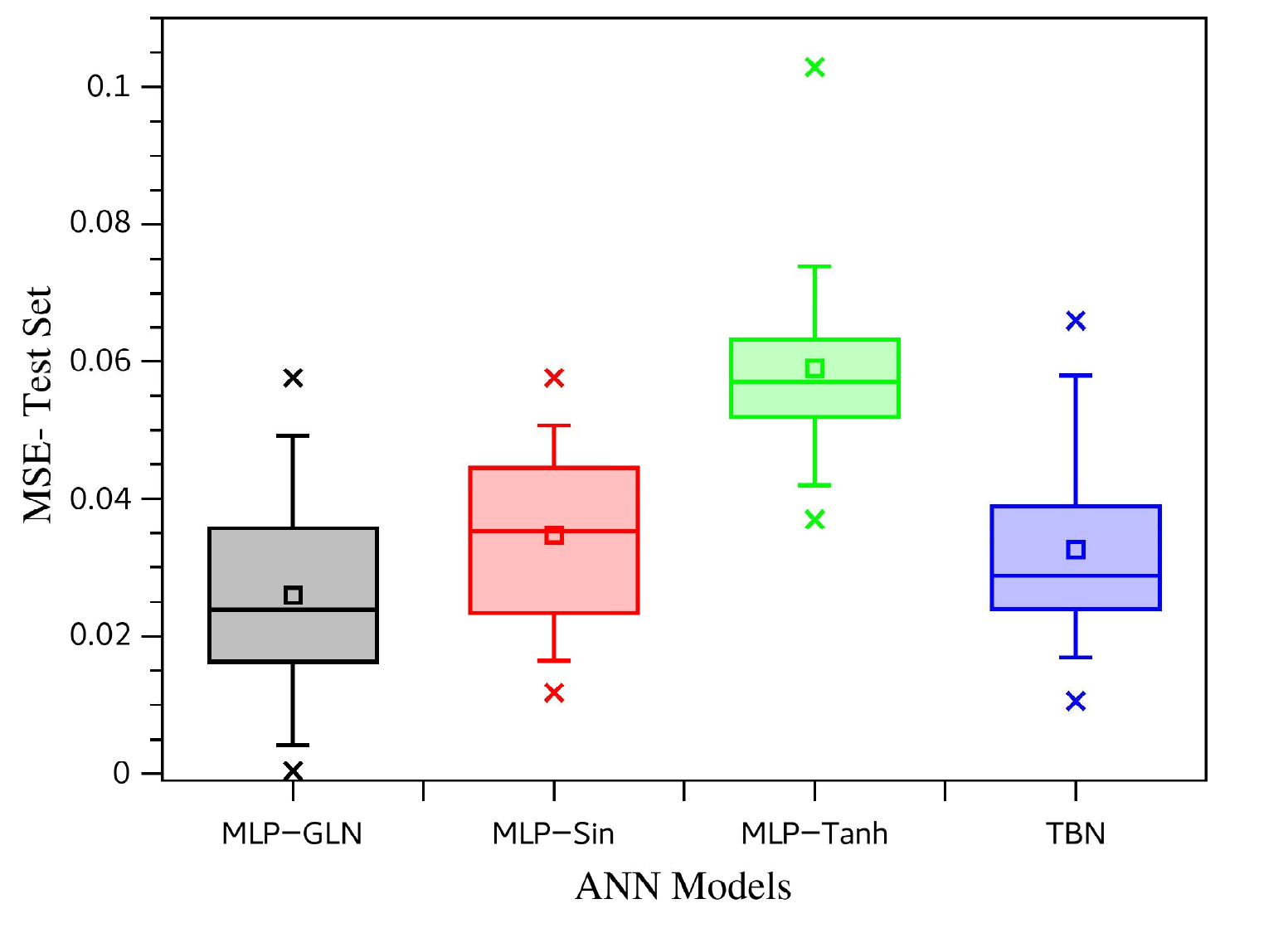}
         \caption{$1-20-1$ Architecture.}
         \label{fig:SunTestMSE_a}
     \end{subfigure}
     \\
     \vspace{20pt}    
     \begin{subfigure}[b]{0.47\textwidth}
         \centering
         \includegraphics[width=\textwidth]{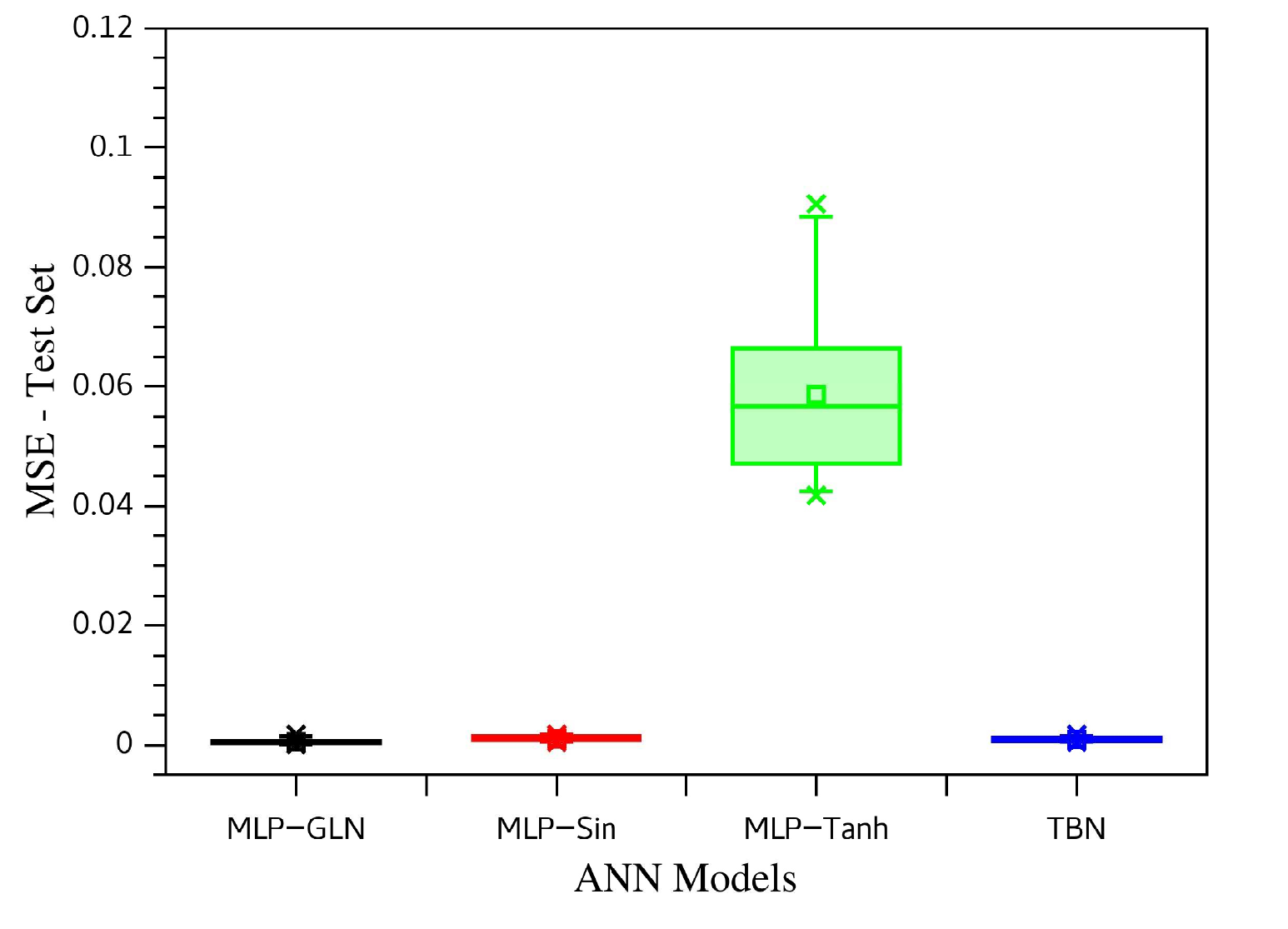}
         \caption{$1-20-20-1$ Architecture.}
         \label{fig:SunTestMSE_b}
     \end{subfigure}
     \hfill
     \begin{subfigure}[b]{0.47\textwidth}
         \centering
         \includegraphics[width=\textwidth]{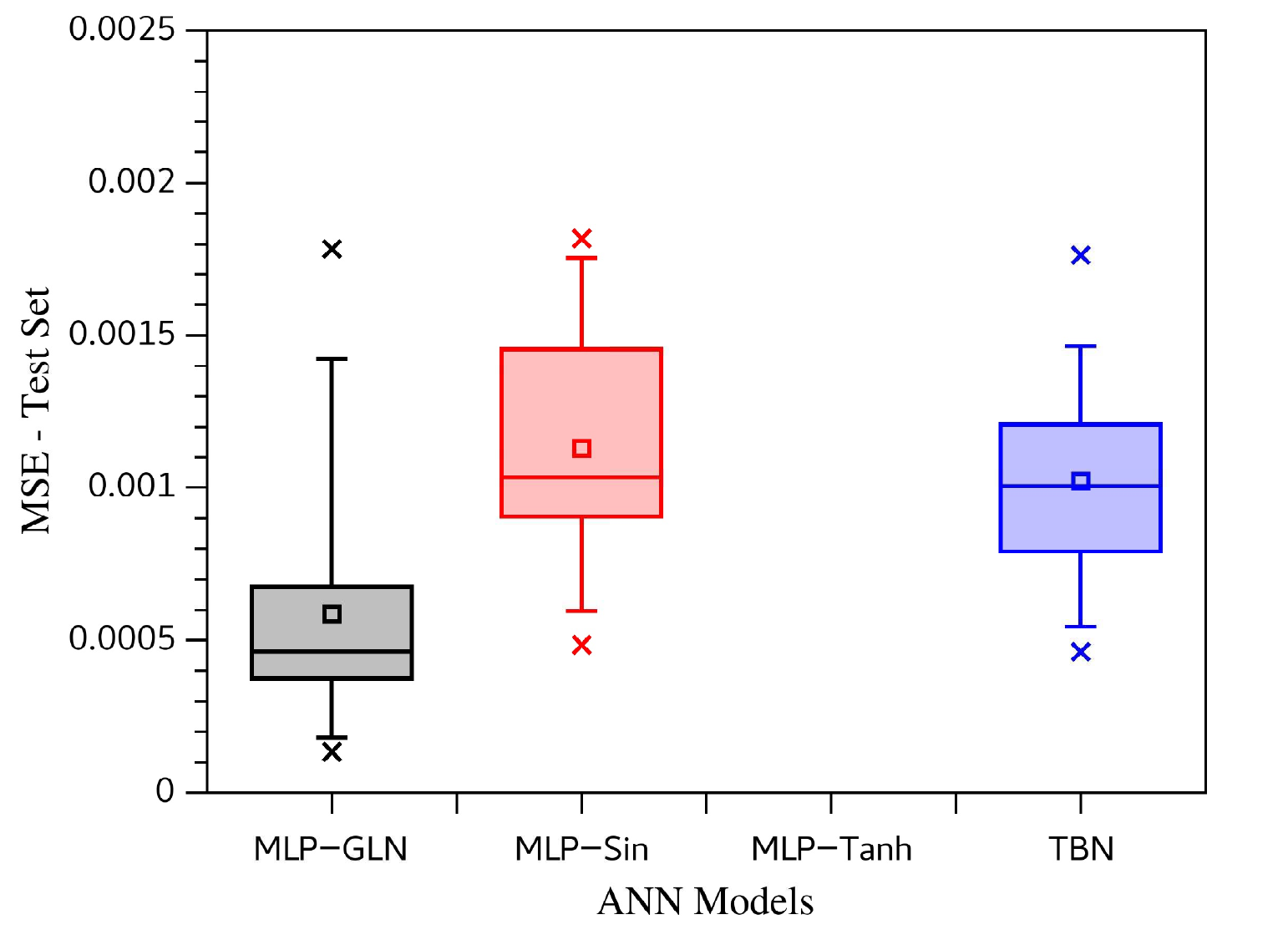}
         \caption{$1-20-20-1$ Architecture - Zoom scale.}
         \label{fig:SunTestMSE_c}
     \end{subfigure}
    \caption{The test set MSE box-plot for all ANN models and both architectures for the Sunspot data set. In (a) are presented the MSE distributions for the 30 repetitions for each ANN model, architecture $1-20-1$. In (b) and (c) are shown the results for architecture $1-20-20-1$ in regular and zoom scale, respectively.} \label{fig:SunTestMSE}
\end{figure}

\begin{figure}[!htb]
    \centering
     \begin{subfigure}[b]{0.47\textwidth}
         \centering
         \includegraphics[width=\textwidth]{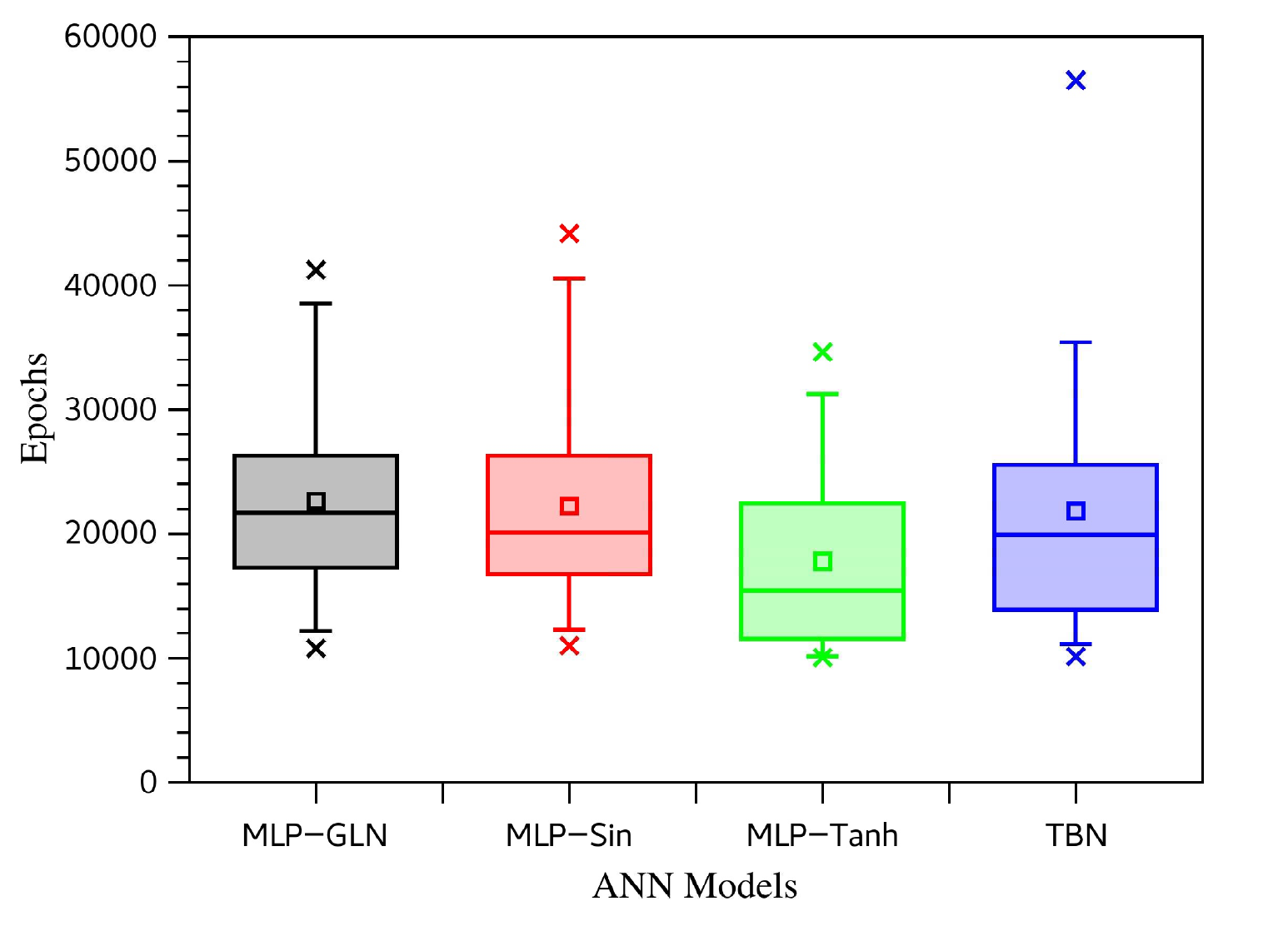}
         \caption{$1-20-1$ Architecture.}
         \label{fig:Sun_Epochs_a}
     \end{subfigure}
     \hfill
     \begin{subfigure}[b]{0.47\textwidth}
         \centering
         \includegraphics[width=\textwidth]{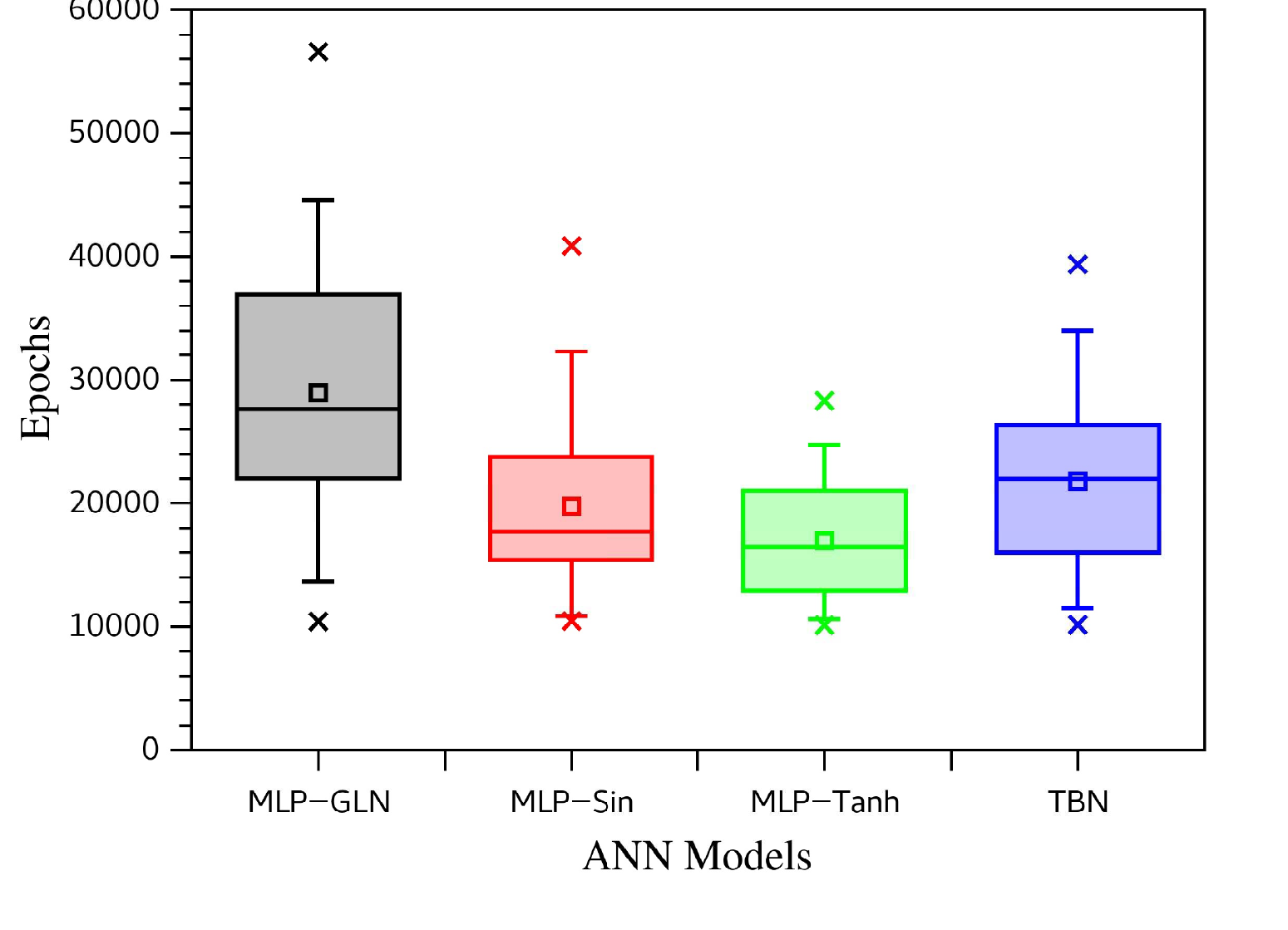}
         \caption{$1-20-20-1$ Architecture.}
         \label{fig:Sun_Epochs_b}
     \end{subfigure}  
     \caption{The epochs distributions for all ANN models -- Sunspot Data set. (a) is related to architecture $1-20-1$, and (b) is related to architecture $1-20-20-1$.} \label{fig:Sun_Epochs}
\end{figure}

The MLP-GLN model performance investigation on a real-world data set, the Sunspot data set, was done here. Figure \ref{fig:SunTestMSE} presents the MSE distribution for the analyzed models, and Figure \ref{fig:Sun_Epochs}, the distribution for the number of epochs.  

For the architecture $1-20-1$, although the MLP-GLN model has reached the best values of mean and median for MSE, presented in Table \ref{tab:SunTestMSE}, the statistical behavior of the MLP-Sin and TBN models is similar to the MLP-GLN model, for the MSE and epochs distributions. Table \ref{tab:KS_Sun} presents the results of the two-sample KS test at $5\%$ of significance level employed to verify whether the sample distribution of the analyzed models comes from the same population distribution or not. Only the MLP-Tanh was considered as a model with a distinct statistical behavior when compared with the MLP-GLN.  

\begin{figure}[!htb]
    \centering
     \begin{subfigure}[b]{0.47\textwidth}
         \centering
         \includegraphics[width=\textwidth]{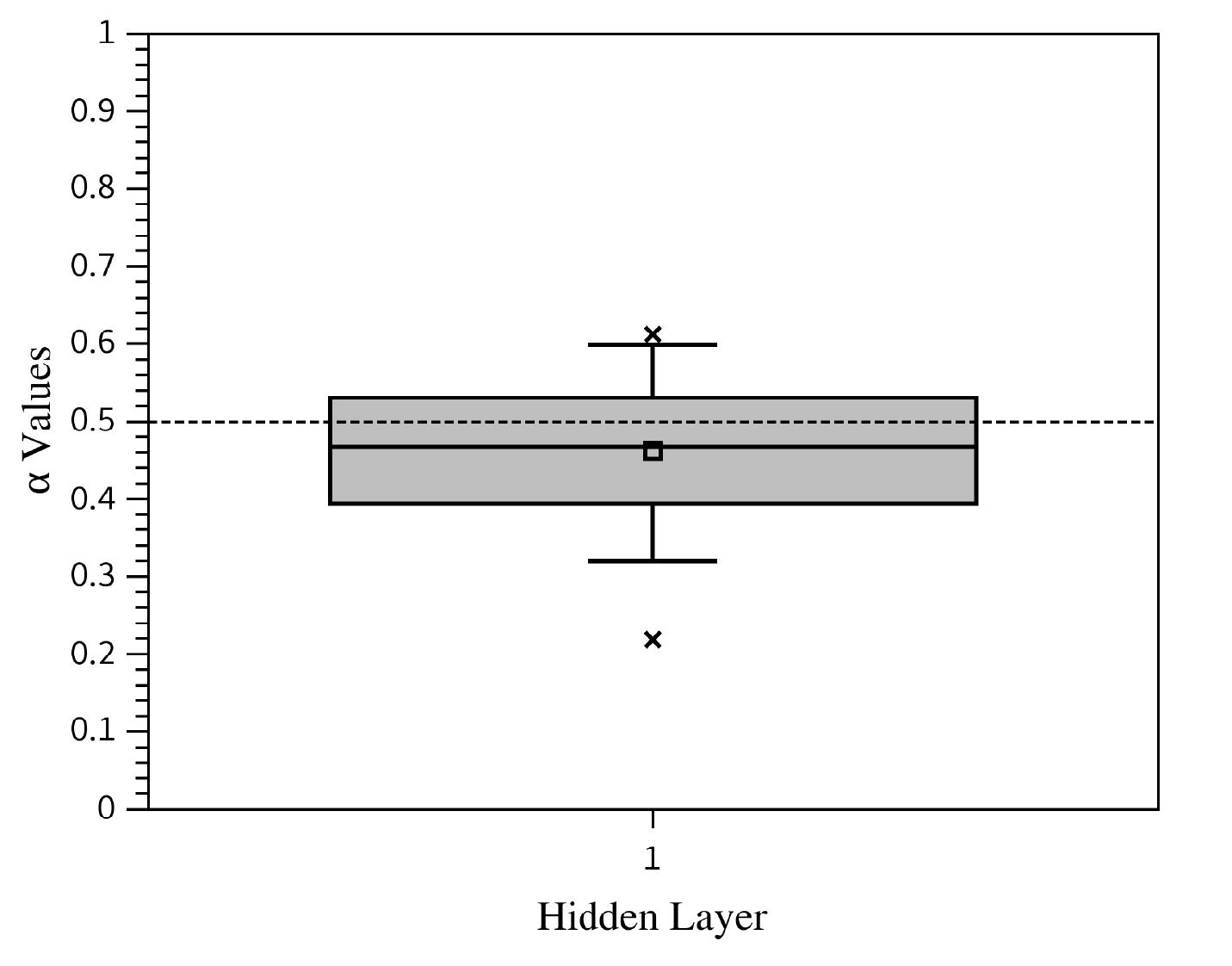}
         \caption{$1-20-1$ Architecture.}
         \label{fig:Sun_alphas_a}
     \end{subfigure}
     \hfill
     \begin{subfigure}[b]{0.47\textwidth}
         \centering
         \includegraphics[width=\textwidth]{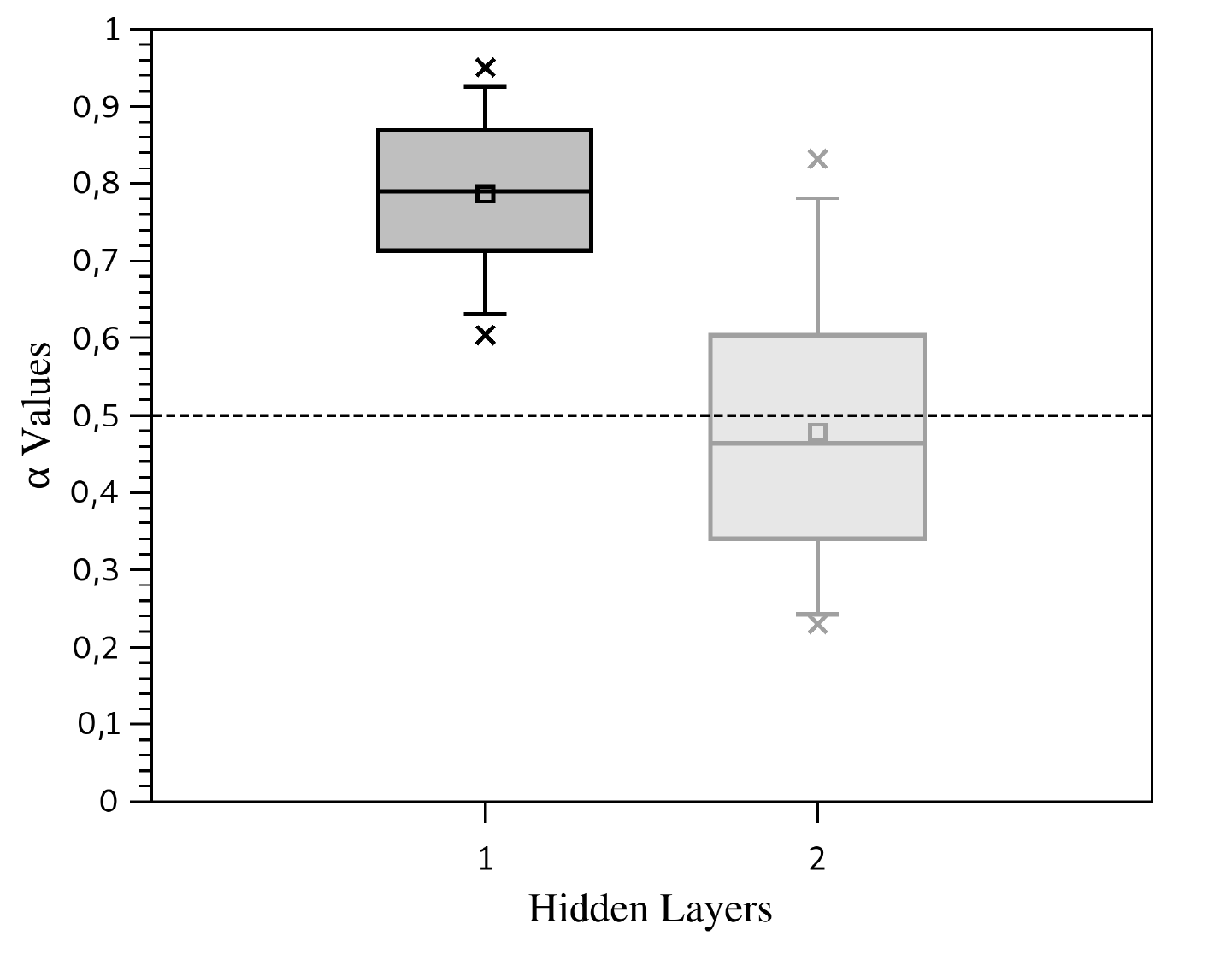}
         \caption{$1-20-20-1$ Architecture.}
         \label{fig:Sun_alphas_b}
     \end{subfigure}
     \caption{The $\alpha$ distributions for the MLP-GLN models after the training process -- Sunspot data set.}
     \label{fig:Sun_alphas}
\end{figure}

Figure \ref{fig:Sun_alphas_a} shows the $\alpha$ values distribution for the MLP-GLN model with one hidden layer. The mean and the median values of the $\alpha$ distribution are a little less than $0.5$, demonstrating an approximate equilibrium between the global and local components, with a slight predominance for the local component in the activation function.

\begin{table}[!htb]
    \centering
     \caption{The descriptive statistics for all ANN models analyzed. All MSE measures are relative to the Test Set of the Sunspot data set. The epochs measures are referent to the training process. The best results are highlighted in bold-face.}
    \label{tab:SunTestMSE}
    \begin{tabular}{ccccccc}
    \hline
    \multicolumn{3}{c}{\multirow{2}{*}{\textbf{Statistics}}} & \multicolumn{4}{c}{\textbf{ANN Models}}\\ 
    \cline{4-7}
     & & & MLP-GLN & MLP-Sin & MLP-Tanh & TBN \\
    \hline
    \multirow{12}{*}{\rotatebox[origin=c]{90}{$1-20-1$}} &\multirow{6}{*}{\rotatebox[origin=c]{90}{MSE}}
    & Min.     & $\mathbf{3.455\cdot10^{-4}}$ & $0.012$ & $0.037$ & $0.011$ \\
    & & Max.   & $\mathbf{0.058}$ & $\mathbf{0.058}$ & $0.103$ & $0.066$ \\
    & & Mean   & $\mathbf{0.026}$ & $0.034$ & $0.059$ & $0.033$ \\
    & & Median & $\mathbf{0.024}$ & $0.035$ & $0.057$ & $0.029$ \\
    & & Std.   & $0.015$ & $\mathbf{0.012}$ & $\mathbf{0.012}$ & $0.013$ \\
    & & CV     & $0.567$ & $0.346$ & $\mathbf{0.210}$ & $0.408$ \\
    \cline{2-7}
    &\multirow{6}{*}{\rotatebox[origin=c]{90}{Epochs}}
    & Min.     & $10785$ & $11013$ & $\mathbf{10048}$ & $10153$ \\
    & & Max.   & $41235$ & $44158$ & $\mathbf{34638}$ & $56472$ \\
    & & Mean   & $22668.833$ & $22222.633$ & $\mathbf{17807.600}$ & $21806.500$ \\
    & & Median & $21639.500$ & $20078.500$ & $\mathbf{15421.000}$ & $19949.000$ \\
    & & Std.   & $8082.901$ & $8454.458$ & $\mathbf{7384.870}$ & $9838.946$ \\
    & & CV     & $\mathbf{0.357}$ & $0.380$ & $0.415$ & $0.451$ \\
    \hline
    \multirow{12}{*}{\rotatebox[origin=c]{90}{$1-20-20-1$}} &\multirow{6}{*}{\rotatebox[origin=c]{90}{MSE}} 
    & Min.     & $\mathbf{1.342\cdot10^{-4}}$ & $4.857\cdot10^{-4}$ & $0.042$ & $4.625\cdot10^{-4}$ \\
    & & Max.   & $1.784\cdot{-3}$ & $1.818\cdot10^{-3}$ & $0.091$ & $\mathbf{1.763\cdot10^{-3}}$ \\
    & & Mean   & $\mathbf{5.863\cdot10^{-4}}$ & $1.129\cdot10^{-3}$ & $0.059$ & $1.021\cdot10^{-3}$ \\
    & & Median & $\mathbf{4.631\cdot10^{-4}}$ & $1.034\cdot10^{-3}$ & $0.057$ & $1.004\cdot10^{-3}$ \\
    & & Std.   & $3.880\cdot10^{-4}$ & $3.718\cdot10^{-4}$ & $0.014$ & $\mathbf{3.038\cdot10^{-4}}$ \\
    & & CV     & $0.662$ & $0.329$ & $\mathbf{0.244}$ & $0.298$ \\
    \cline{2-7}
    &\multirow{6}{*}{\rotatebox[origin=c]{90}{Epochs}}
    & Min.     & $10421$ & $10441$ & $\mathbf{10095}$ & $10158$ \\
    & & Max.   & $56591$ & $40855$ & $\mathbf{28286}$ & $39375$ \\
    & & Mean   & $28921.733$ & $19764.367$ & $\mathbf{16924.433}$ & $21772.967$ \\
    & & Median & $27608.500$ & $17675.00$ & $\mathbf{16430.500}$ & $21925.000$ \\
    & & Std.   & $10818.981$ & $7265.591$ & $\mathbf{5011.714}$ & $7412.880$ \\
    & & CV     & $0.374$ & $0.368$ & $\mathbf{0.296}$ & $0.340$ \\
    \hline
    \end{tabular}
\end{table}

\begin{table}[!htb]
    \centering
    \caption{Two-sample Kolmogorov-Smirnov Test at the $5\%$ significance level for the MSE and number of epochs distributions between the MLP-GLN and all other models for both architectures studied -- Sunspot data set.}
    \label{tab:KS_Sun}
    \begin{tabular}{ccccc}
    \hline
    \multicolumn{3}{c}{\textbf{Tested Model}} & \multicolumn{2}{c}{\textbf{KS Test Results}}\\ 
    \cline{4-5}
    \multicolumn{3}{c}{\textbf{whit MLP-GLN}} & Statistically Similar & $p$-values\\
    \hline
    \multirow{6}{*}{\rotatebox[origin=c]{90}{$1-20-1$}} &\multirow{3}{*}{\rotatebox[origin=c]{90}{MSE}}
    & MLP-Sin    & Yes & $0.055$ \\
    & & MLP-Tanh & No  & $3.520\cdot10^{-10}$ \\
    & & TBN      & Yes & $0.109$ \\
    \cline{2-5}
    &\multirow{3}{*}{\rotatebox[origin=c]{90}{Epochs}}
    & MLP-Sin    & Yes & $0.760$ \\
    & & MLP-Tanh & No & $0.011$ \\
    & & TBN      & Yes & $0.760$ \\
    \hline
    \multirow{6}{*}{\rotatebox[origin=c]{90}{$1-20-20-1$}} &\multirow{3}{*}{\rotatebox[origin=c]{90}{MSE}}
    & MLP-Sin    & No & $1.143\cdot10^{-6}$ \\
    & & MLP-Tanh & No & $1.797\cdot10^{-14}$ \\
    & & TBN      & No & $1.143\cdot10^{-6}$ \\
    \cline{2-5}
    &\multirow{3}{*}{\rotatebox[origin=c]{90}{Epochs}}
    & MLP-Sin    & No & $1.746\cdot10^{-3}$ \\
    & & MLP-Tanh & No & $4.644\cdot10^{-6}$ \\
    & & TBN      & Yes & $0.055$ \\
    \hline
    \end{tabular}
\end{table}

For the architecture $1-20-20-1$, the MLP-GLN had the best results for the MSE performance, except for the CV value, presenting the most prominent relative MSE distribution dispersion, as can be viewed in Table \ref{tab:SunTestMSE}. Applying the two-sample KS test to verify if the other models have the same population distribution of the MLP-GLN, the MLP-Sin presents the same statistical behavior for the MSE and number of epochs distribution, as it is demonstrated in Table \ref{tab:KS_Sun}. Observing the quantity of epochs to train the model,  MLP-GLN is computationally more costly than MLP-Tanh. It was needed about $79\%$ more epochs to train the MLP-GLN than to train the MLP-Tanh. However, when the MLP-GLN MSE distribution is compared with the MLP-Tanh MSE distribution, the MLP-GLN reached a mean MSE value $10^2$ times less than the MLP-Tanh model.

Figure \ref{fig:Sun_alphas_b} presents the $\alpha$ distributions for the two hidden layers. For the first hidden layer, the MLP-GLN gave more importance to the global component (sine function) than the local component, where the mean $\alpha$ value and the median reached were respectively $0.786$ and $0.790$. For the second hidden layer, the MLP-SN gave approximately the same importance for both activation function components but with a little tendency to the local component whit a mean $\alpha$ value of $0.478$ and a median of $0.463$. In this way, the first hidden layer makes an initial process highlighting the global data features. The second hidden layer receives this initial processing and improves the local component's importance to take a more accurate regression result. 


\subsection{Differential Equation Solving Problem}

In this second class of tests applied to assess the performance of the proposed neuron, we employed the same experimental structure used previously for the regression test of the architectures $1-20-1$ and $1-20-20-1$.  It was used 30 repetitions for each experiment with all ANN models, where the Adam training algorithm~\cite{kingma2014adam} with a learning rate of $10^{-3}$ was employed.

The experiments to solve the differential equations presented in Section \ref{sec:DiffEq} were implemented in Python3 with the library \texttt{NeuroDiffEq}\footnote{The NeuroDiffEq library is available in \textit{https://github.com/odegym/neurodiffeq}.} \cite{Giovanni2020}. For each differential equation solving experiment, a set of initial or boundary conditions is presented, as defined in Section \ref{sec:DiffEq}. The maximum number of epochs employed in the training process for each differential equation solving was defined based on the initial experimental investigations. With these parameters, the library \texttt{NeuroDiffEq} has trained the ANN models until the maximum number of epochs is set, returning the trained ANN models defined in the training point where the smallest validation loss occurs. The \texttt{NeuroDiffEq} library does not return information about the number of epochs used to select the ANN model. For this reason, any analysis of the number of epochs employed to train the ANN models was not done for the differential equations solving experiments.

\subsubsection{Exponential Decay Equation Results}

As presented in Section~\ref{sec:decayEq}, the Decay differential equation with initial condition $u(t=0)=1$ has analytical solution defined by $u(t) = e^{-t}$. The idea is to generate a solution $u(t)$ based on ANN models to describe this analytical solution combined with the initial condition of $u(t=0)=1$. All experiments used the domain of $t \in [0,3]$ with a maximum number of epochs of $500$ to training the ANN models.

\begin{figure}[!htb]
     \centering
     \begin{subfigure}[b]{0.47\textwidth}
         \centering
         \includegraphics[width=\textwidth]{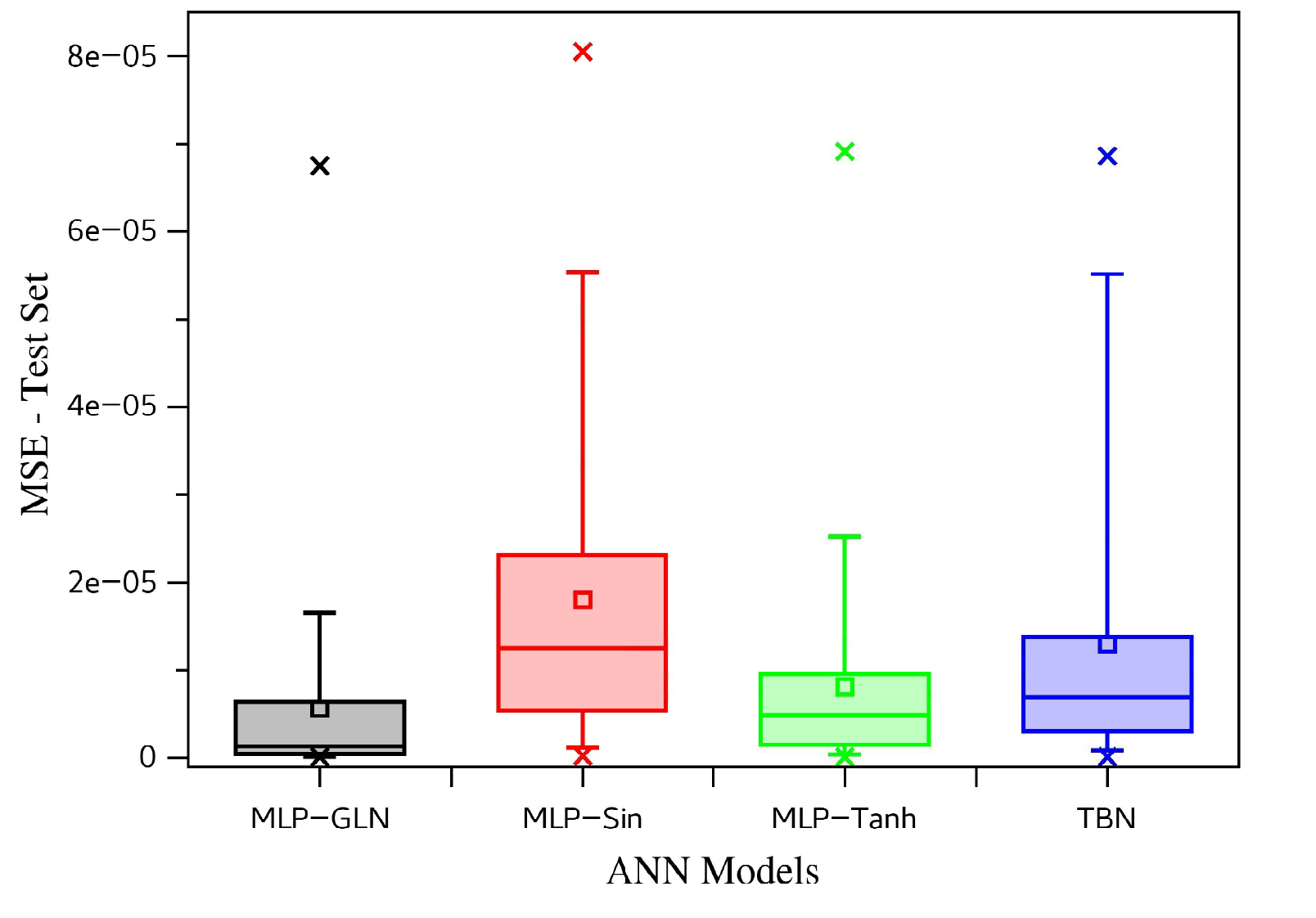}
         \caption{$1-20-1$ Architecture.}
         \label{fig:DeacyTestMSE_a}
     \end{subfigure}
     \\
     \vspace{20pt}    
     \begin{subfigure}[b]{0.47\textwidth}
         \centering
         \includegraphics[width=\textwidth]{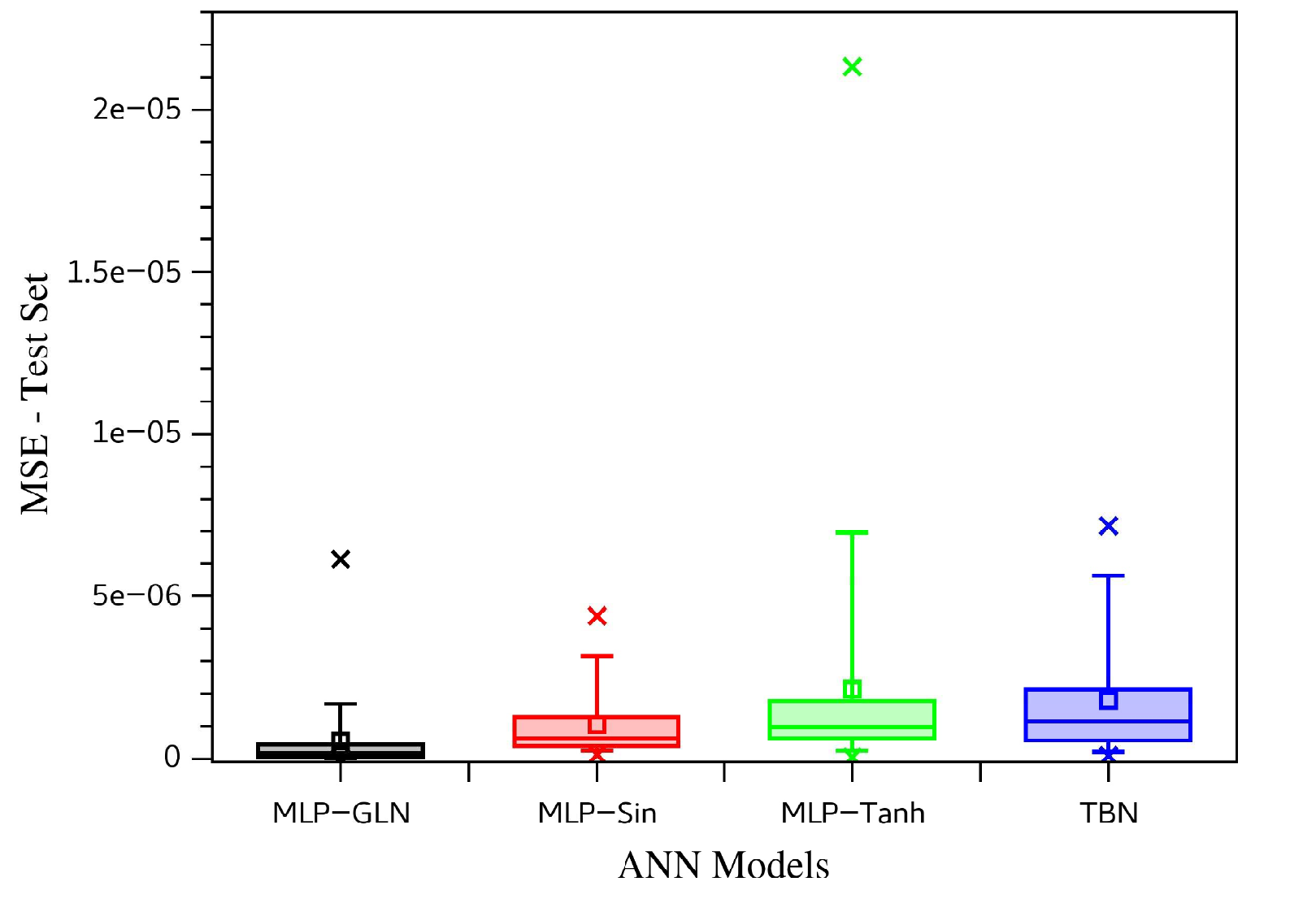}
         \caption{$1-20-20-1$ Architecture.}
         \label{fig:DecayTestMSE_b}
     \end{subfigure}
     \hfill
     \begin{subfigure}[b]{0.47\textwidth}
         \centering
         \includegraphics[width=\textwidth]{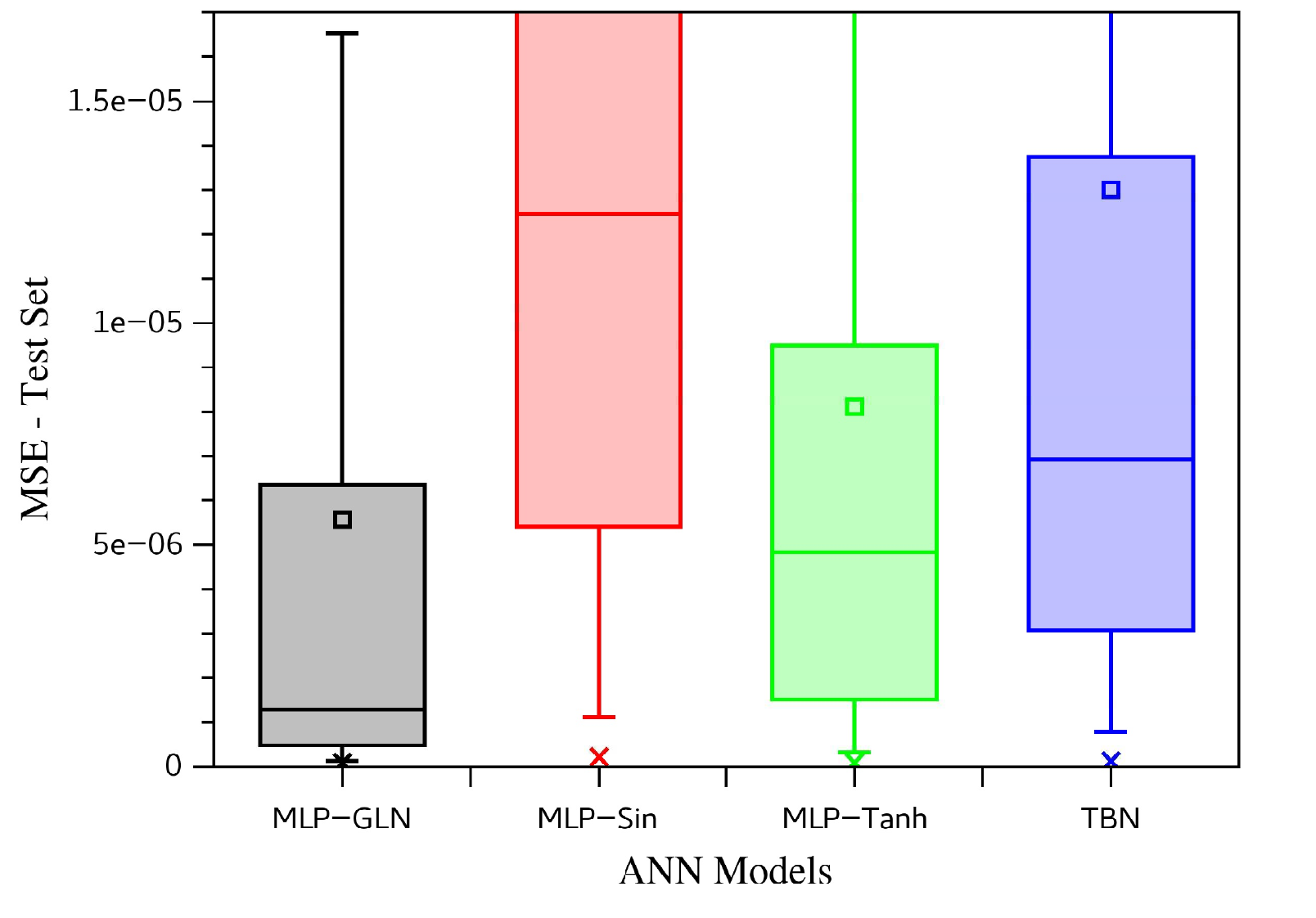}
         \caption{$1-20-20-1$ Architecture - Zoom scale.}
         \label{fig:DecayTestMSE_c}
     \end{subfigure}
    \caption{The test set MSE box plot for all ANN models studied and both architectures for solving the Exponential Decay differential equation. In (a) is presented the MSE distributions for the 30 repetitions for each ANN model for architectures $1-20-1$. In (b) and (c) are shown the results for architecture $1-20-20-1$ on regular and zoom scale.} \label{fig:DecayTestMSE}
\end{figure}

Figure \ref{fig:DecayTestMSE} shows the MSE distribution for Decay equation ANN solution, where Figure \ref{fig:DeacyTestMSE_a} is relative to the experiments with $1-20-1$ architecture, while Figures \ref{fig:DecayTestMSE_b} and \ref{fig:DecayTestMSE_c} are for the $1-20-20-1$ experiments. The descriptive statistics for these MSE distributions are presented in Table \ref{tab:DecayMSE}. All models were verified with the two-sample KS test, where the MLP-GLN is statistically distinct of all other ANN models tested, as demonstrated in Table \ref{tab:KS_Delay}.

In general, the MLP-GLN reaches better MSE performance than the other ANN models (MLP-Sin, MLP-Tanh, and TBN). The MLP-GLN has the smallest values for MSE mean and median distributions for both architectures but presents the largest CV value. However, even with the biggest CV value, the MLP-GLN MSE distribution has the maximum value less or the same order than the maximum MSE value of the other ANN models. Hence, the MLP-GLN is the best option in the context of  MSE performance measures.    
 
\begin{table}[!ht]
    \centering
     \caption{The descriptive statistics for all ANN models analyzed. All MSE measures are relative to the Exponential Decay solving. The best results are highlighted in bold-face.}
    \label{tab:DecayMSE}
    \begin{tabular}{cccccc}
    \hline
    \multicolumn{2}{c}{\multirow{2}{*}{\textbf{MSE}}} & \multicolumn{4}{c}{\textbf{ANN Models}}\\ 
    \cline{3-6}
     & & MLP-GLN & MLP-Sin & MLP-Tanh & TBN \\
    \hline
    \multirow{6}{*}{\rotatebox[origin=c]{90}{$1-20-1$}} 
    & Min.   & $9.773\cdot10^{-8}$ & $2.237\cdot10^{-7}$ & $\mathbf{8.909\cdot10^{-8}}$ & $1.241\cdot10^{-7}$ \\
    & Max.   & $\mathbf{6.748\cdot10^{-5}}$ & $8.051\cdot10^{-5}$ & $6.916\cdot10^{-5}$ & $6.863\cdot10^{-5}$ \\
    & Mean   & $\mathbf{5.567\cdot10^{-6}}$ & $1.799\cdot10^{-5}$ & $8.120\cdot10^{-6}$ & $1.300\cdot10^{-5}$ \\
    & Median & $\mathbf{1.276\cdot10^{-6}}$ & $1.246\cdot10^{-5}$ & $4.829\cdot10^{-6}$ & $6.921\cdot10^{-6}$ \\
    & Std.   & $\mathbf{1.258\cdot10^{-5}}$ & $1.855\cdot10^{-5}$ & $1.334\cdot10^{-5}$ & $1.720\cdot10^{-5}$ \\
    & CV     & $2.259$ & $\mathbf{1.031}$ & $1.643$ & $1.324$ \\
    \hline
    \multirow{6}{*}{\rotatebox[origin=c]{90}{$1-20-20-1$}}  
    & Min.   & $\mathbf{1.947\cdot10^{-9}}$ & $1.156\cdot10^{-7}$ & $3.530\cdot10^{-8}$ & $8.665\cdot10^{-8}$ \\
    & Max.   & $6.129\cdot10^{-6}$ & $\mathbf{4.398\cdot10^{-6}}$ & $2.132\cdot10^{-5}$ & $7.165\cdot10^{-6}$ \\
    & Mean   & $\mathbf{5.218\cdot10^{-7}}$ & $1.046\cdot10^{-6}$ & $2.144\cdot10^{-6}$ & $1.786\cdot10^{-6}$ \\
    & Median & $\mathbf{1.550\cdot10^{-7}}$ & $6.065\cdot10^{-7}$ & $9.607\cdot10^{-7}$ & $1.142\cdot10^{-6}$ \\
    & Std.   & $1.155\cdot10^{-6}$ & $\mathbf{1.046\cdot10^{-6}}$ & $4.025\cdot10^{-6}$ & $1.825\cdot10^{-6}$ \\
    & CV     & $2.212$ & $\mathbf{1.000}$ & $1.877$ & $1.022$ \\
    \hline
    \end{tabular}
\end{table}

\begin{table}
    \centering
    \caption{Two-sample Kolmogorov-Smirnov Test at the $5\%$ significance level for the MSE distributions between the MLP-GLN and all other models for both architectures studied for solving the Exponential Decay differential equation.}
    \label{tab:KS_Delay}
    \setlength{\extrarowheight}{5.5pt}
    \begin{tabular}{ccccc}
    \hline
    \multicolumn{3}{c}{\textbf{Tested Model}} & \multicolumn{2}{c}{\textbf{KS Test Results}}\\ 
    \cline{4-5}
    \multicolumn{3}{c}{\textbf{whit MLP-GLN}} & Statistically Similar & $p$-values\\
    \hline
    \multirow{3}{*}{\scriptsize \rotatebox[origin=c]{90}{$1-20-1$} }  &\multirow{3}{*}{\rotatebox[origin=c]{90}{MSE}}
    & MLP-Sin    & No & $6.174\cdot10^{-5}$ \\
    & & MLP-Tanh & No & $0.0259$ \\
    & & TBN      & No & $0.0113$ \\
    \hline
    \multirow{3}{*}{\scriptsize \rotatebox[origin=c]{90}{$1-20-20-1$} }  &\multirow{3}{*}{\rotatebox[origin=c]{90}{MSE}}
    & MLP-Sin    & No & $4.644\cdot10^{-6}$ \\
    & & MLP-Tanh & No & $1.143\cdot10^{-6}$ \\
    & & TBN      & No & $1.755\cdot10^{-5}$ \\
    \hline
    \end{tabular}
\end{table}

\begin{figure}[!htb]
    \centering
     \begin{subfigure}[b]{0.47\textwidth}
         \centering
         \includegraphics[width=\textwidth]{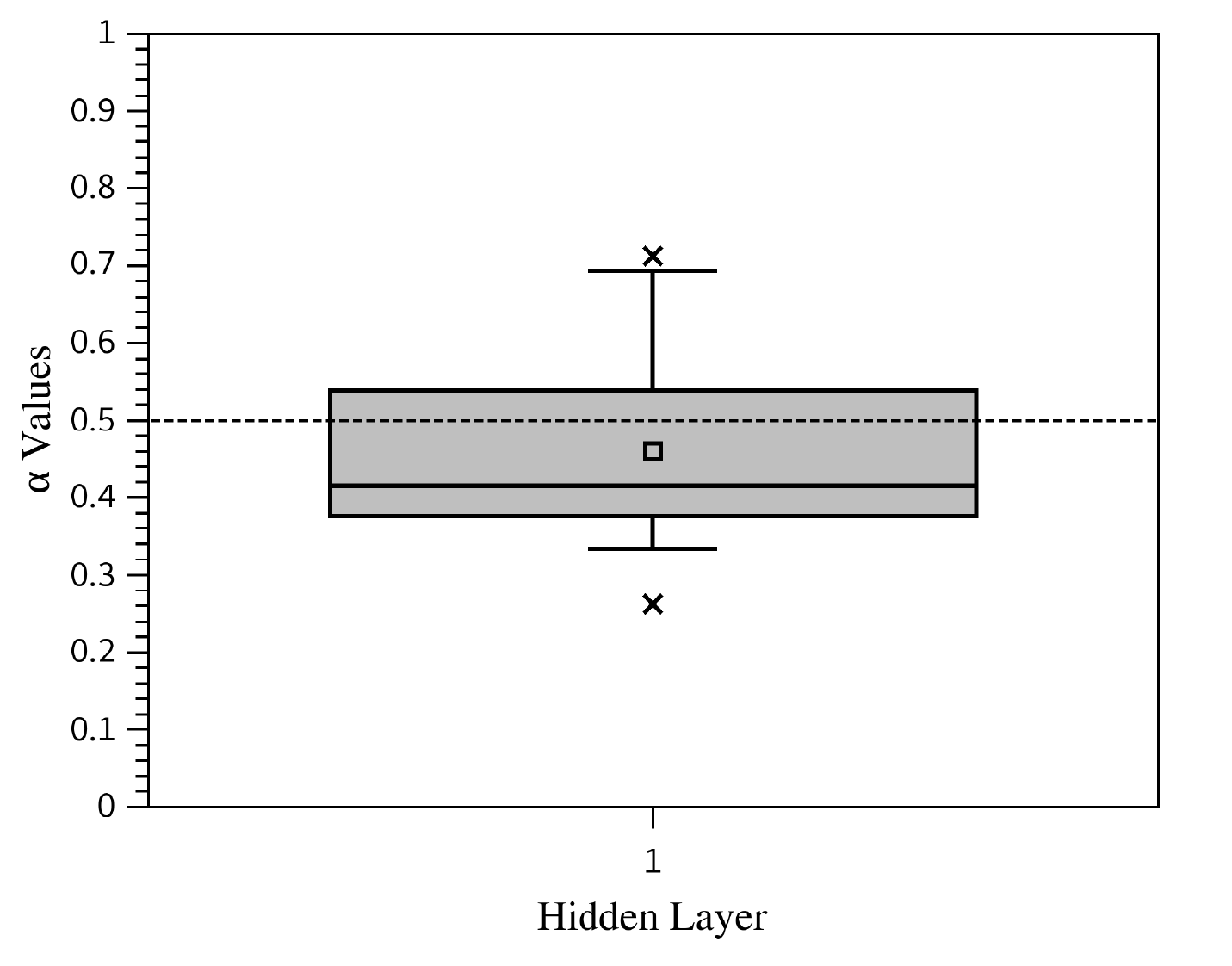}
         \caption{$1-20-1$ Architecture.}
         \label{fig:Decay_alphas_a}
     \end{subfigure}
     \hfill
     \begin{subfigure}[b]{0.47\textwidth}
         \centering
         \includegraphics[width=\textwidth]{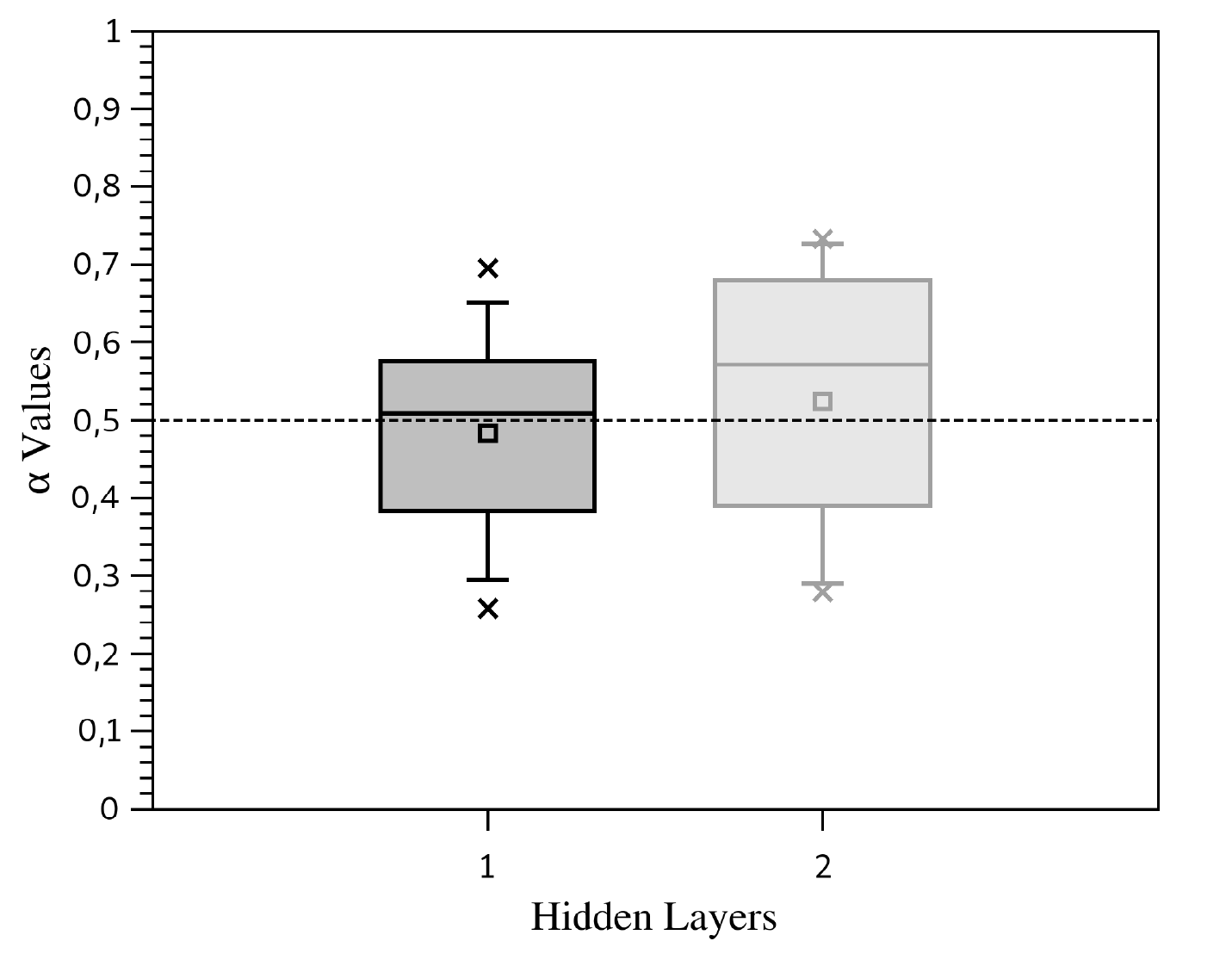}
         \caption{$1-20-20-1$ Architecture.}
         \label{fig:Decay_alphas_b}
     \end{subfigure}
     \caption{The $\alpha$ distributions for the MLP-GLN models after the training process for solving the Exponential Decay differential equation.}
     \label{fig:Decay_alphas}
\end{figure}

The $\alpha$ value distributions for both architectures are shown in Figure \ref{fig:Decay_alphas}. Figure \ref{fig:Decay_alphas_a} refers to the $1-20-1$ architecture, where the mean value of $0.460$ and median value of $0.416$ indicate a slight tendency to the local component. Figure \ref{fig:Decay_alphas_b} corresponds to the $1-20-20-1$ architecture, where at the first hidden layer the $\alpha$ mean and median are, respectively, $0.483$ and $0.508$, and for the second layer are $0.525$ and $0.571$.

\subsubsection{Catenary Equation Results}

The catenary curve is the differential equation solution presented in Section \ref{sec:Catenary}. The ANN models for both $1-20-1$ and $1-20-20-1$ architectures were employed to solve the catenary differential equation (Equation \ref{eqn:catenary}). For all experiments, the domain of $x \in [-2, 2]$ was used to discover an approximate solution of the differential equation. The maximum number of epochs of $700$ was employed for all ANN models training phases.  

\begin{figure}
     \centering
     \begin{subfigure}[b]{0.47\textwidth}
         \centering
         \includegraphics[width=\textwidth]{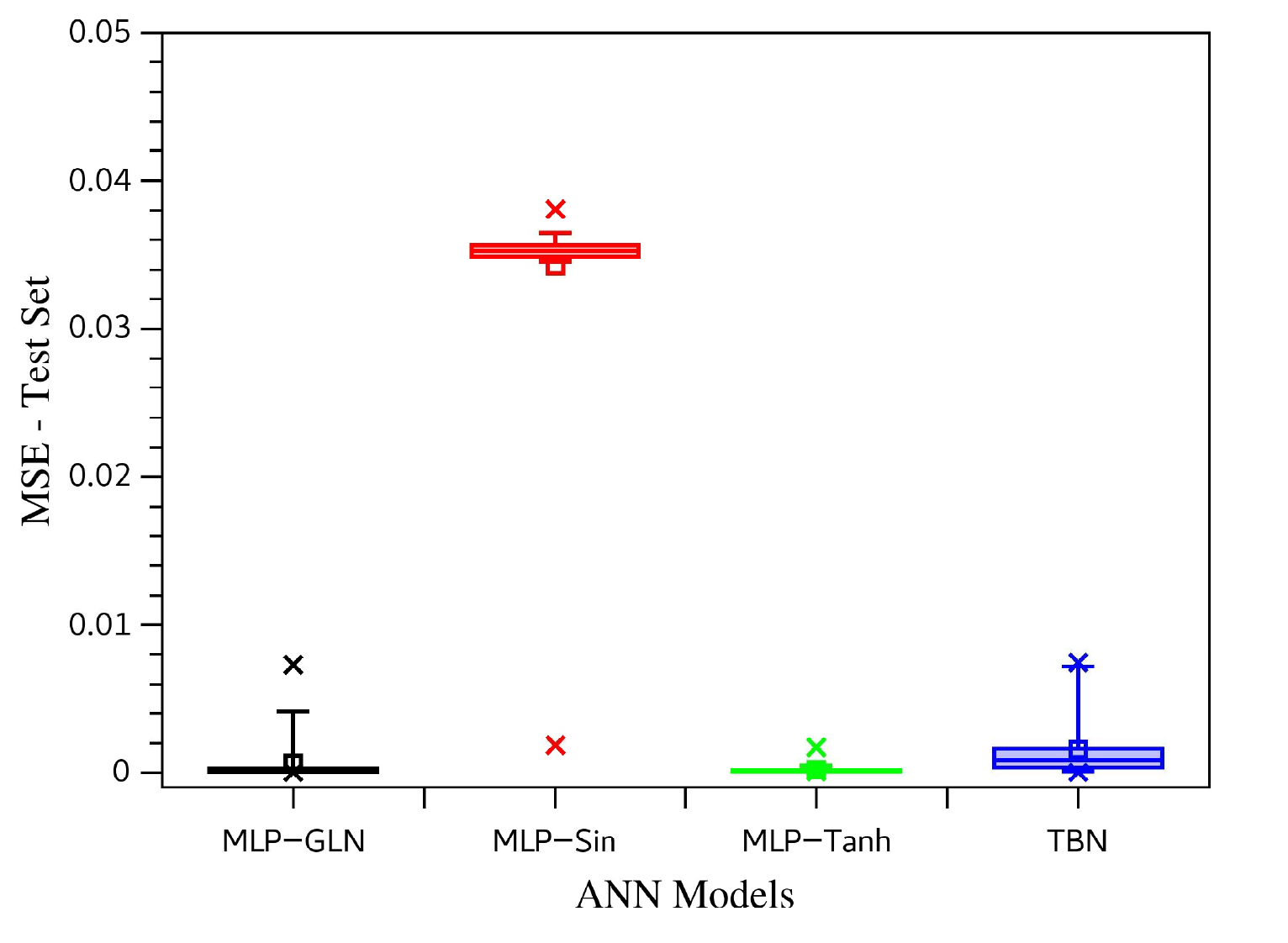}
         \caption{$1-20-1$ Architecture.}
         \label{fig:CatenaryTestMSE_a}
     \end{subfigure}
     \hfill
     \begin{subfigure}[b]{0.47\textwidth}
         \centering
         \includegraphics[width=\textwidth]{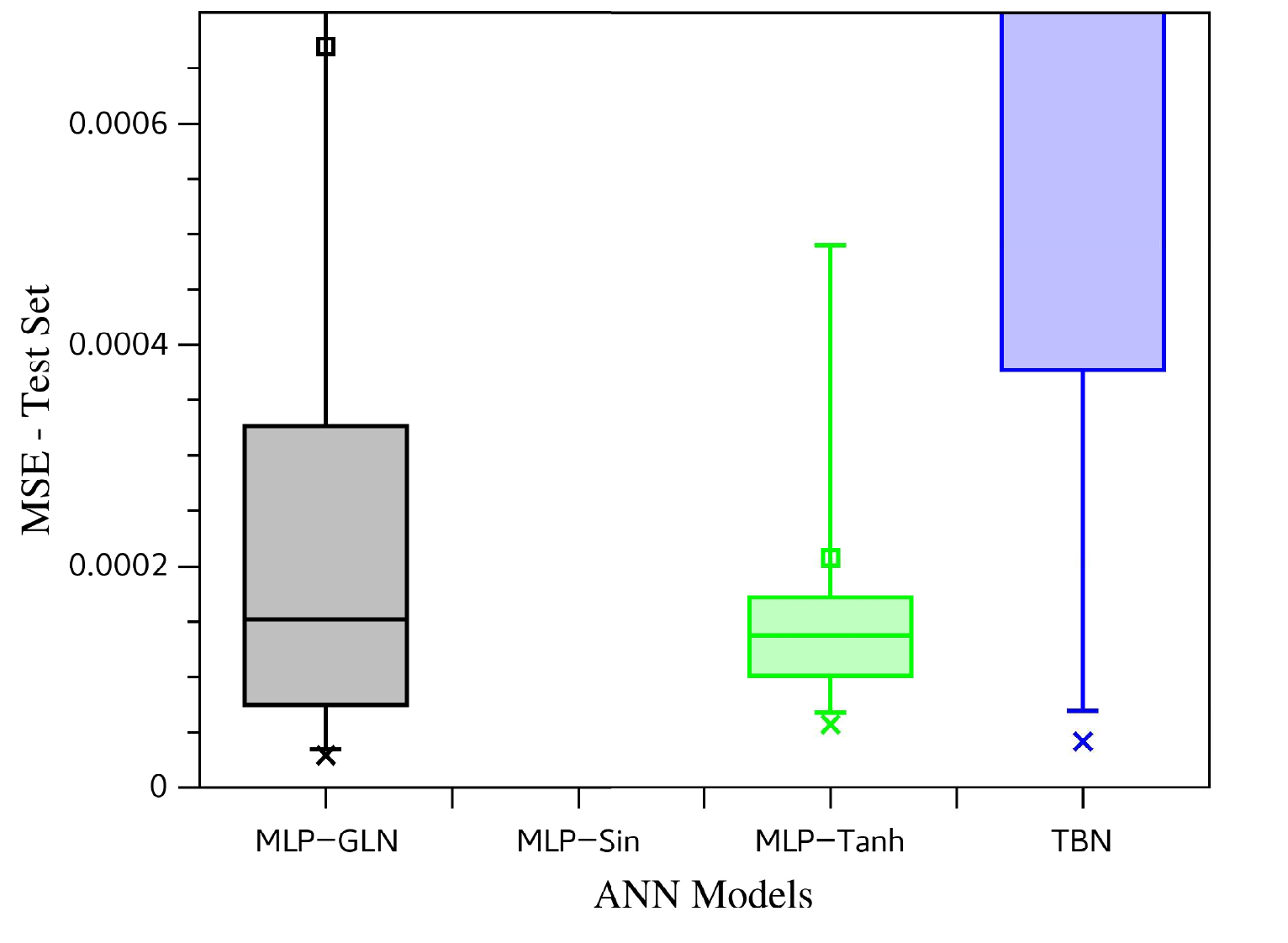}
         \caption{$1-20-1$ Architecture.}
         \label{fig:CatenaryTestMSE_b}
     \end{subfigure}
     \hfill
     
     \vspace{20pt}    
     \begin{subfigure}[b]{0.47\textwidth}
         \centering
         \includegraphics[width=\textwidth]{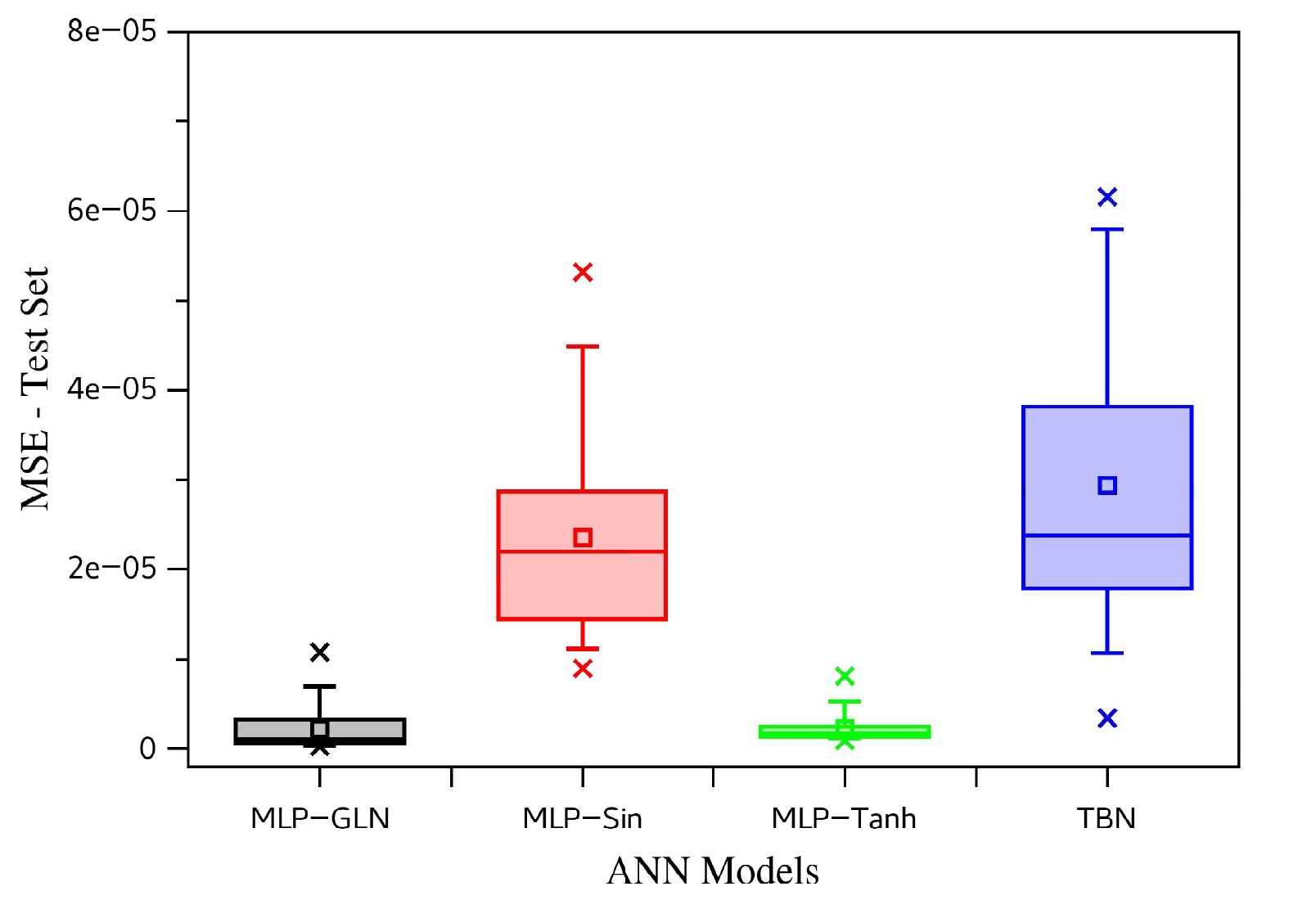}
         \caption{$1-20-20-1$ Architecture.}
         \label{fig:CatenaryTestMSE_c}
     \end{subfigure}
     \hfill
     \begin{subfigure}[b]{0.47\textwidth}
         \centering
         \includegraphics[width=\textwidth]{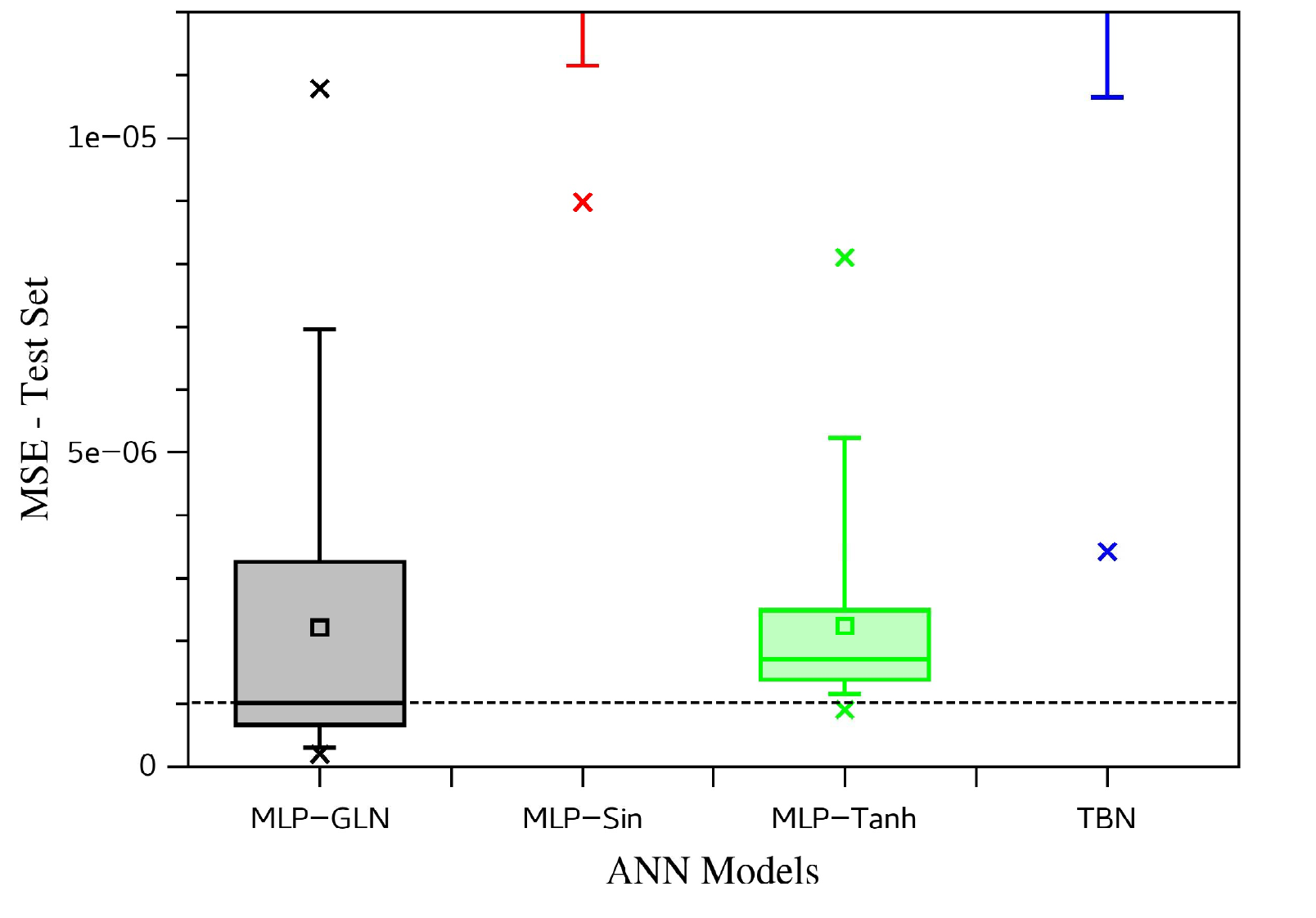}
         \caption{$1-20-20-1$ Architecture - Zoom scale.}
         \label{fig:CatenaryTestMSE_d}
     \end{subfigure}
    \caption{The MSE test set box-plot for all ANN models studied and both architectures to solve the Catenary differential equation. In (a) and (b) are presented the MSE distributions for the 30 repetitions for each ANN model for architectures $1-20-1$ in standard and zoom scale. In (c) and (d) are shown the results for architecture $1-20-20-1$ in standard and zoom scale.} \label{fig:CatenaryTestMSE}
\end{figure}

Figure \ref{fig:CatenaryTestMSE} shows the MSE distribution for all analyzed ANN models. Figures \ref{fig:CatenaryTestMSE_a}  and \ref{fig:CatenaryTestMSE_b} are relative to the $1-20-1$ architecture, while Figures \ref{fig:CatenaryTestMSE_c} and \ref{fig:CatenaryTestMSE_d} correspond to the $1-20-20-1$ architecture. For the architecture with one hidden layer, the best model was the MLP-Tanh. However, the MLP-GLN reached the best mean and median values of the MSE distribution for the architecture with two hidden layers. All descriptive statistics of these MSE distributions can be viewed in Table \ref{tab:CatenaryMSE}. In general, although the MLP-GLN model does not always show the best mean and median MSE values, at least this model has a MSE's mean performance similar to the best model.

\begin{table}[!ht]
    \centering
     \caption{The descriptive statistics for ANN models analyzed. All MSE measures are relative to the Catenary differential equation solving. The best results are highlighted in bold-face.}
    \label{tab:CatenaryMSE}
    \begin{tabular}{cccccc}
    \hline
    \multicolumn{2}{c}{\multirow{2}{*}{\textbf{MSE}}} & \multicolumn{4}{c}{\textbf{ANN Models}}\\ 
    \cline{3-6}
     & & MLP-GLN & MLP-Sin & MLP-Tanh & TBN \\
    \hline
    \multirow{6}{*}{\rotatebox[origin=c]{90}{$1-20-1$}} 
    & Min.   & $\mathbf{2.907\cdot10^{-5}}$ & $1.883\cdot10^{-3}$ & $5.671\cdot10^{-5}$ & $4.194\cdot10^{-5}$ \\
    & Max.   & $7.299\cdot10^{-3}$ & $3.807\cdot10^{-2}$ & $\mathbf{1.747\cdot10^{-3}}$ & $7.434\cdot10^{-3}$ \\
    & Mean   & $6.696\cdot10^{-4}$ & $3.427\cdot10^{-2}$ & $\mathbf{2.072\cdot10^{-4}}$ & $1.602\cdot10^{-3}$ \\
    & Median & $1.520\cdot10^{-4}$ & $3.522\cdot10^{-2}$ & $\mathbf{1.370\cdot10^{-4}}$ & $8.473\cdot10^{-4}$ \\
    & Std.   & $1.731\cdot10^{-3}$ & $6.162\cdot10^{-3}$ & $\mathbf{3.120\cdot10^{-4}}$ & $2.112\cdot10^{-3}$ \\
    & CV     & $2.586$ & $\mathbf{0.180}$ & $1.506$ & $1.318$ \\
    \hline
    \multirow{6}{*}{\rotatebox[origin=c]{90}{$1-20-20-1$}}  
    & Min.   & $\mathbf{2.056\cdot10^{-7}}$ & $8.980\cdot10^{-6}$ & $9.024\cdot10^{-7}$ & $3.4237\cdot10^{-6}$ \\
    & Max.   & $1.079\cdot10^{-5}$ & $5.320\cdot10^{-5}$ & $\mathbf{8.103\cdot10^{-6}}$ & $6.156\cdot10^{-5}$ \\
    & Mean   & $\mathbf{2.2135\cdot10^{-6}}$ & $2.358\cdot10^{-5}$ & $2.235\cdot10^{-6}$ & $2.939\cdot10^{-5}$ \\
    & Median & $\mathbf{1.005\cdot10^{-6}}$ & $2.196\cdot10^{-5}$ & $1.713\cdot10^{-6}$ & $2.377\cdot10^{-5}$ \\
    & Std.   & $2.466\cdot10^{-6}$ & $1.062\cdot10^{-5}$ & $\mathbf{1.533\cdot10^{-6}}$ & $1.566\cdot10^{-5}$ \\
    & CV     & $1.114$ & $\mathbf{0.450}$ & $0.686$ & $0.533$ \\
    \hline
    \end{tabular}
\end{table}

For this differential equation, the MLP-Sin, MLP-Tanh, and TBN models were not statistically similar to the MLP-GLN when tested with the two-sample KS test. Table \ref{tab:KS_Catenary} exhibits the results of the KS test at $5\%$ significance, indicating that MLP-GLN has a different population distribution from all other ANN models' population distributions.  

\begin{table}
    \centering
    \caption{Two-sample Kolmogorov-Smirnov Test at the $5\%$ significance level for the MSE distributions between the MLP-GLN and all other models for both architectures studied for the Catenary differential equation solving.}
    \label{tab:KS_Catenary}
    \setlength{\extrarowheight}{5.5pt}
    \begin{tabular}{ccccc}
    \hline
    \multicolumn{3}{c}{\textbf{Tested Model}} & \multicolumn{2}{c}{\textbf{KS Test Results}}\\ 
    \cline{4-5}
    \multicolumn{3}{c}{\textbf{whit MLP-GLN}} & Statistically Similar & $p$-values\\
    \hline
    \multirow{3}{*}{\scriptsize \rotatebox[origin=c]{90}{$1-20-1$} }  &\multirow{3}{*}{\rotatebox[origin=c]{90}{MSE}}
    & MLP-Sin    & No & $1.498\cdot10^{-13}$ \\
    & & MLP-Tanh & No & $0.026$ \\
    & & TBN      & No & $1.755\cdot10^{-5}$ \\
    \hline
    \multirow{3}{*}{\scriptsize \rotatebox[origin=c]{90}{$1-20-20-1$} }  &\multirow{3}{*}{\rotatebox[origin=c]{90}{MSE}}
    & MLP-Sin    & No & $ 1.498\cdot10^{-13}$ \\
    & & MLP-Tanh & No & $6.158\cdot10^{-4}$ \\
    & & TBN      & No & $1.162\cdot10^{-12}$ \\
    \hline
    \end{tabular}
\end{table}

\begin{figure}
    \centering
     \begin{subfigure}[b]{0.47\textwidth}
         \centering
         \includegraphics[width=\textwidth]{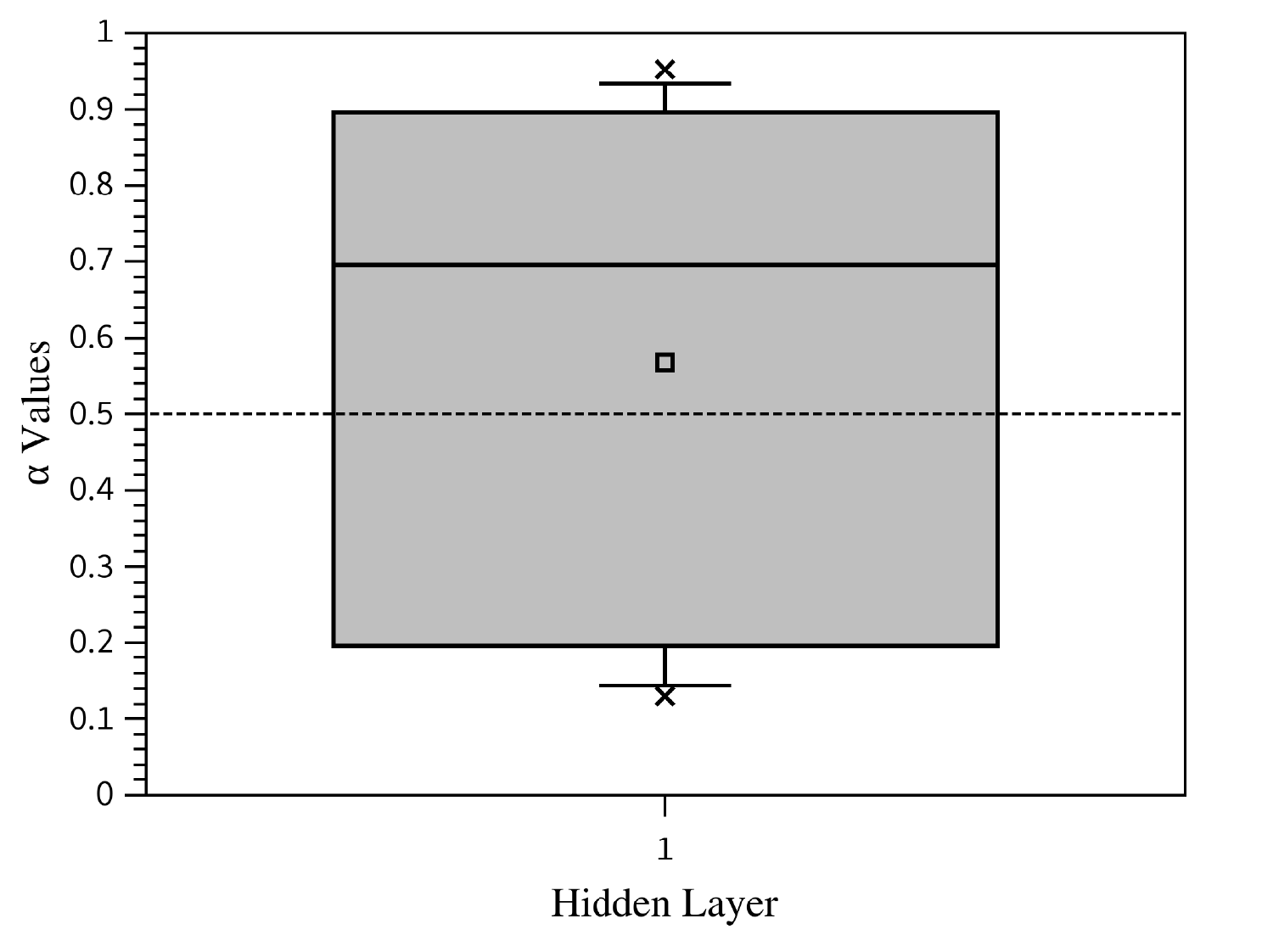}
         \caption{$1-20-1$ Architecture.}
         \label{fig:Catenary_alphas_a}
     \end{subfigure}
     \hfill
     \begin{subfigure}[b]{0.47\textwidth}
         \centering
         \includegraphics[width=\textwidth]{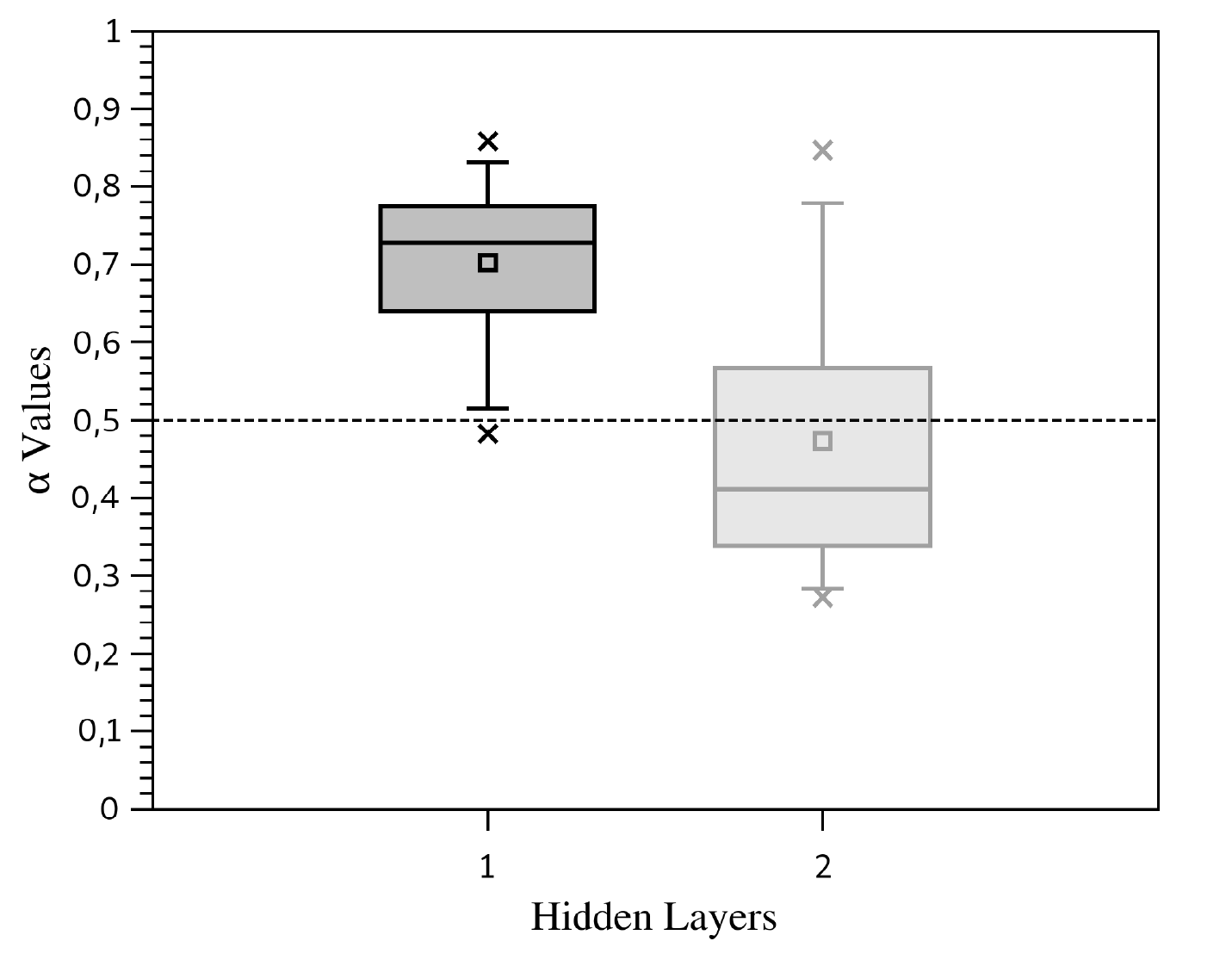}
         \caption{$1-20-20-1$ Architecture.}
         \label{fig:Catenary_alphas_b}
     \end{subfigure}
     \caption{The $\alpha$ distributions for the MLP-GLN models after the training process for solving the Catenary differential equation.}
     \label{fig:Catenary_alphas}
\end{figure}

Figure \ref{fig:Catenary_alphas} presents the $\alpha$ weight value distributions. Figure \ref{fig:Catenary_alphas_a} is for the $1-20-1$ architecture. It is possible to note a significant dispersion for the $\alpha$ weight values, with a total variation from $0.1304$ to $0.9530$. The mean and median values reached were $0.568$ and $0.695$, respectively, indicating a global behavior tendency. This behavior in the selected $\alpha$ values can tell that both components have similar importance. The MLP-GLN found a solution with three possible activation function behavior, convergence, purely local, purely global, and a mix of these components.      

For the architecture $1-20-20-1$, the $\alpha$ values distributions were more stable. Figure \ref{fig:Catenary_alphas_b} shows the $\alpha$ weight distributions for both hidden layers. The first hidden layer selected values of $\alpha$ greater than $0.5$ indicating that the global component of the activation function has great importance in this layer. The second layer selected values of $\alpha$ in general less than $0.5$ indicating a tendency to give more importance to the local component of the activation function. Thus, for these two hidden layer architecture, the ANN initially selected the global information with the sine function, and after that, the ANN gave more importance to the local component in the second hidden layer. This behavior gives the idea that the ANN first makes a global evaluation of the information and then checks the information locally.

\subsubsection{Simple Harmonic Oscillator Equation Results}

The differential equation for a simple harmonic oscillator, presented in Section \ref{sec:SimpleOscillator}, has an analytical solution described by a global function, like a sine. Thus, it is expected that ANN models with sine activation function are more appropriate to describe this differential equation's solution.

For all experiments, the differential equation solutions were sought in the domain $t \in [0, 3\pi]$, and all ANN models were trained with maximum epoch number of $700$.

\begin{figure}
     \centering
     \begin{subfigure}[b]{0.47\textwidth}
         \centering
         \includegraphics[width=\textwidth]{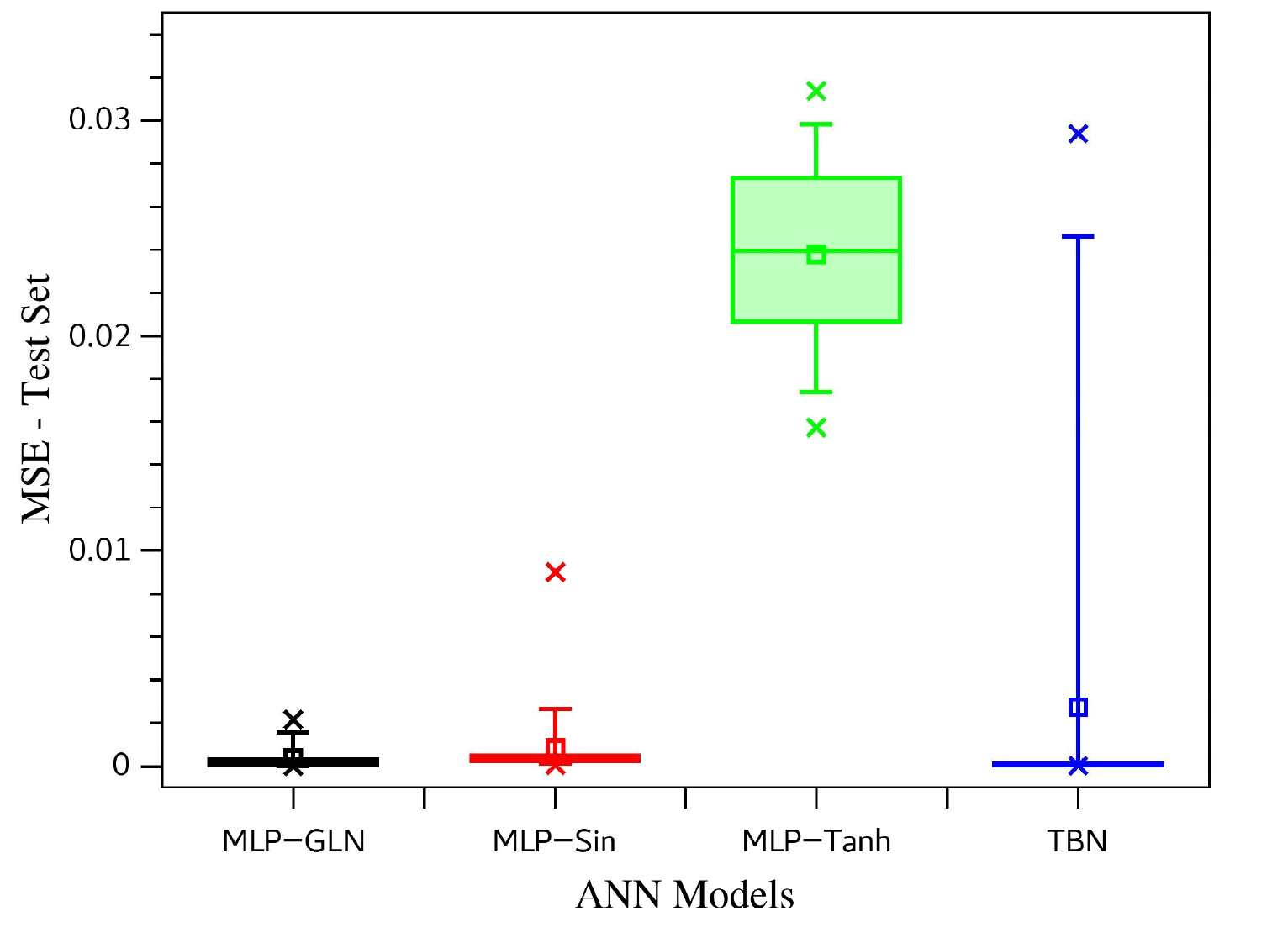}
         \caption{$1-20-1$ Architecture.}
         \label{fig:OscillatorTestMSE_a}
     \end{subfigure}
     \hfill
     \begin{subfigure}[b]{0.47\textwidth}
         \centering
         \includegraphics[width=\textwidth]{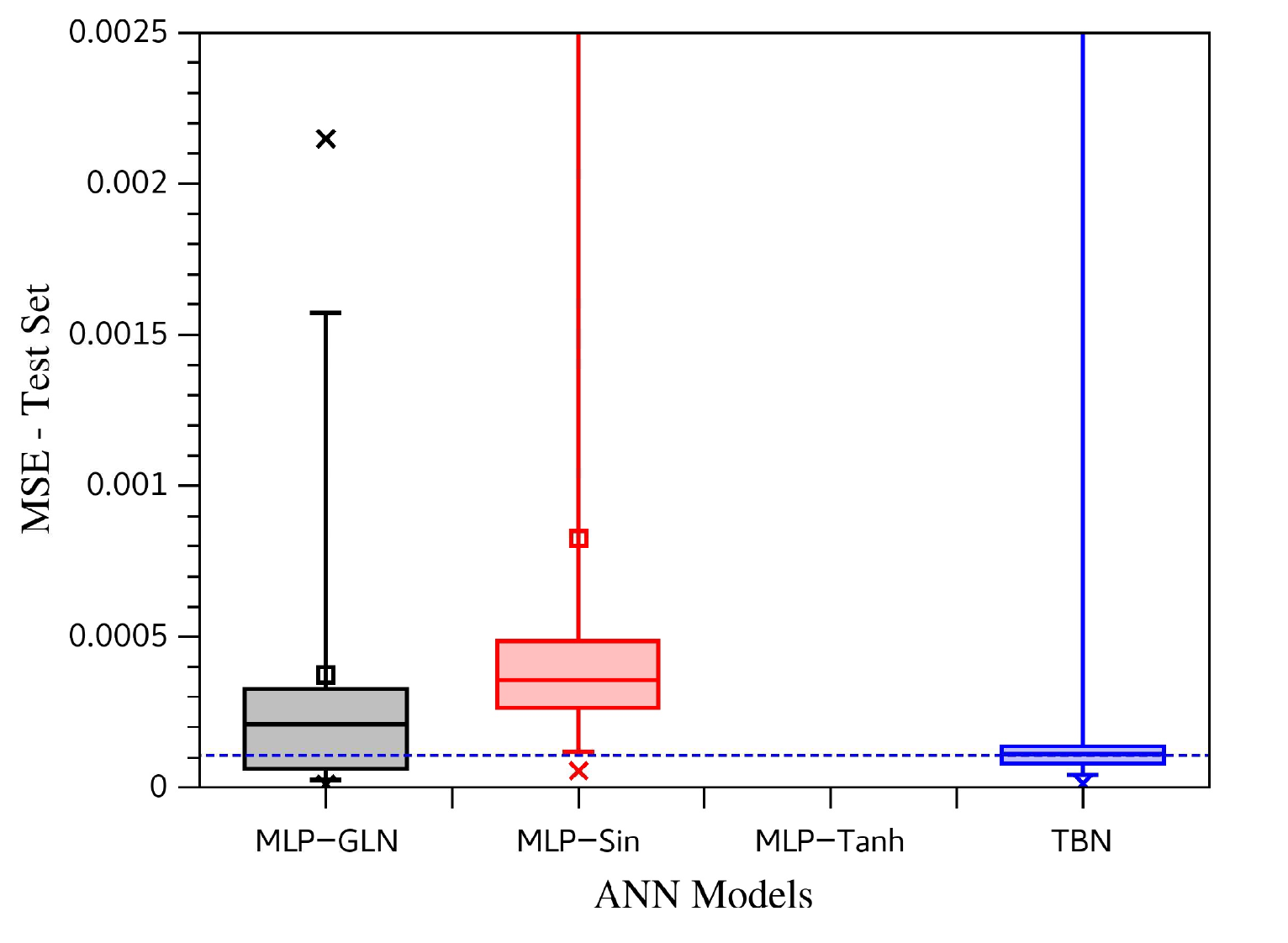}
         \caption{$1-20-1$ Architecture - Zoom scale.}
         \label{fig:OscillatorTestMSE_b}
     \end{subfigure}
     \vspace{20pt}    
     \begin{subfigure}[b]{0.47\textwidth}
         \centering
         \includegraphics[width=\textwidth]{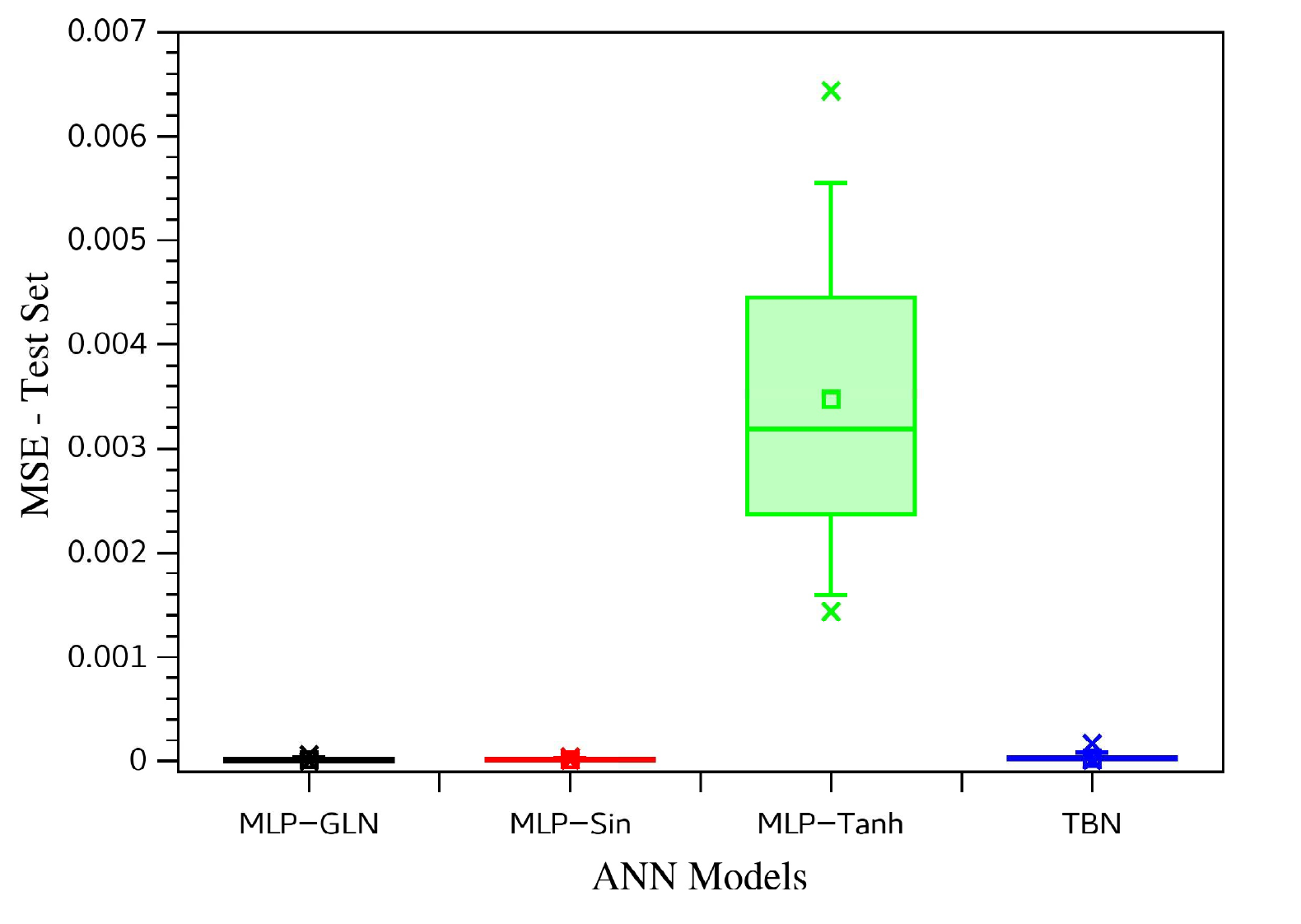}
         \caption{$1-20-20-1$ Architecture.}
         \label{fig:OscillatorTestMSE_c}
     \end{subfigure}
     \hfill
     \begin{subfigure}[b]{0.47\textwidth}
         \centering
         \includegraphics[width=\textwidth]{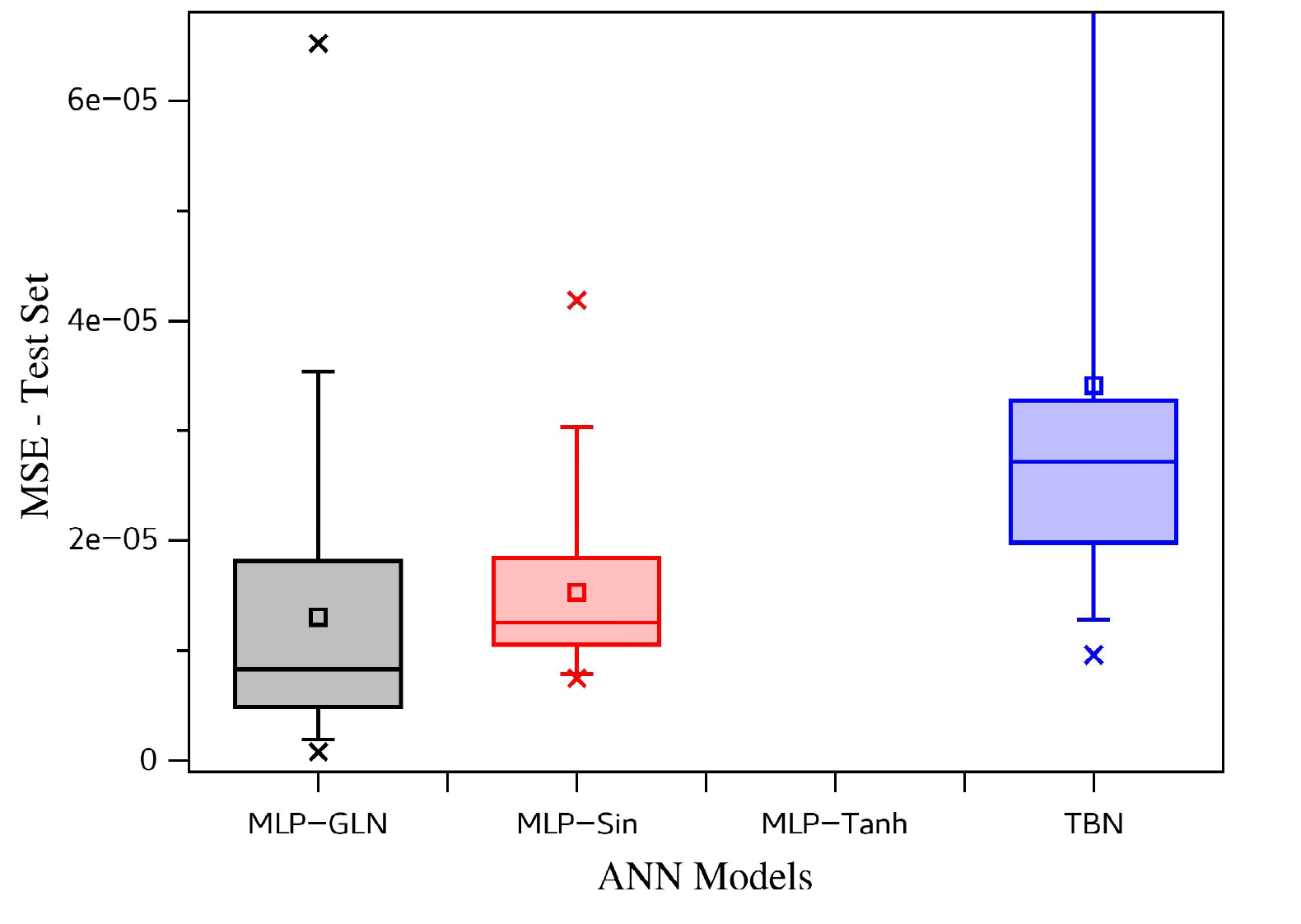}
         \caption{$1-20-20-1$ Architecture - Zoom scale.}
         \label{fig:OscillatorTestMSE_d}
     \end{subfigure}
    \caption{The test set MSE box-plot for all ANN models studied and both architectures for the Harmonic Oscillator differential equation solving. In (a) and (b) are presented the MSE distributions for the 30 repetitions for each ANN model for architecture $1-20-1$. In (c) and (d) is shown the results for architecture $1-20-20-1$.} \label{fig:OscillatorTestMSE}
\end{figure}

Figure \ref{fig:OscillatorTestMSE} presents the test set MSE distributions for all ANN models. Figures \ref{fig:OscillatorTestMSE_a} and \ref{fig:OscillatorTestMSE_b} are referent to $1-20-1$ architecture, and Figures \ref{fig:OscillatorTestMSE_c} and \ref{fig:OscillatorTestMSE_d} are referent to $1-20-20-1$ architecture.  All descriptive statistics of these MSE distributions are shown in Table \ref{tab:OscillatorMSE}.

For the architecture $1-20-1$, the models MLP-GLN, MLP-Sin, and TBN obtained the best MSE performances. Looking at the mean values of the MSE distributions, the MLP-GLN model was the best, with a value of $3.736\cdot 10^{-4}$. However, observing the MSE distributions' median, we read that the TBN is the best model, with a value of $1.113 \cdot 10^{-4}$, and with the best results for the second and third quarterlies. The MLP-Sin performance stayed between these two models. If the MSE distribution dispersion is observed, the MLP-GLN presents a CV of $1.420$ against the MLP-Sin and TBN CV values of $2.034$ and $2.925$. In this way, the MLP-GLN reached the best stability with the smallest MSE distribution dispersion for these three models. The best individual model MSE also has reached by the MLP-GLN, as can be viewed in Table \ref{tab:OscillatorMSE} in the Min. MSE field.  

\begin{table}[!ht]
    \centering
     \caption{The descriptive statistics for all ANN models analyzed. All MSE measures are relative to the Simple Harmonic Oscillator differential equation solving. The best results are highlighted in bold-face.}
    \label{tab:OscillatorMSE}
    \begin{tabular}{cccccc}
    \hline
    \multicolumn{2}{c}{\multirow{2}{*}{\textbf{MSE}}} & \multicolumn{4}{c}{\textbf{ANN Models}}\\ 
    \cline{3-6}
     & & MLP-GLN & MLP-Sin & MLP-Tanh & TBN \\
    \hline
    \multirow{6}{*}{\rotatebox[origin=c]{90}{$1-20-1$}} 
    & Min.   & $\mathbf{1.0719\cdot10^{-5}}$ & $5.620\cdot10^{-5}$ & $1.574\cdot10^{-2}$ & $1.304\cdot10^{-5}$ \\
    & Max.   & $\mathbf{2.149\cdot10^{-3}}$ & $9.000\cdot10^{-3}$ & $3.137\cdot10^{-2}$ & $2.938\cdot10^{-2}$ \\
    & Mean   & $\mathbf{3.736\cdot10^{-4}}$ & $8.255\cdot10^{-4}$ & $2.377\cdot10^{-2}$ & $2.732\cdot10^{-3}$ \\
    & Median & $2.086\cdot10^{-4}$ & $3.575\cdot10^{-4}$ & $2.393\cdot10^{-2}$ & $\mathbf{1.113\cdot10^{-4}}$ \\
    & Std.   & $\mathbf{5.307\cdot10^{-4}}$ & $1.671\cdot10^{-3}$ & $4.323\cdot10^{-3}$ & $7.991\cdot10^{-3}$ \\
    & CV     & $1.420$ & $2.024$ & $\mathbf{0.182}$ & $2.925$ \\
    \hline
    \multirow{6}{*}{\rotatebox[origin=c]{90}{$1-20-20-1$}}  
    & Min.   & $\mathbf{7.890\cdot10^{-7}}$ & $7.480\cdot10^{-6}$ & $1.437\cdot10^{-3}$ & $9.606\cdot10^{-6}$ \\
    & Max.   & $6.518\cdot10^{-5}$ & $\mathbf{4.183\cdot10^{-5}}$ & $6.436\cdot10^{-3}$ & $1.704\cdot10^{-4}$ \\
    & Mean   & $\mathbf{1.304\cdot10^{-5}}$ & $1.532\cdot10^{-5}$ & $3.472\cdot10^{-3}$ & $3.410\cdot10^{-5}$ \\
    & Median & $\mathbf{8.267\cdot10^{-6}}$ & $1.256\cdot10^{-5}$ & $3.183\cdot10^{-3}$ & $2.711\cdot10^{-5}$ \\
    & Std.   & $1.363\cdot10^{-5}$ & $\mathbf{7.556\cdot10^{-6}}$ & $1.366\cdot10^{-3}$ & $3.033\cdot10^{-5}$ \\
    & CV     & $1.046$ & $0.493$ & $\mathbf{0.394}$ & $0.889$ \\
    \hline
    \end{tabular}
\end{table}

For the architecture $1-20-20-1$, there was a similar behavior, the MLP-GLN, MLP-Sin, and TBN models have the best MSE performance. For this architecture with two hidden layers, the MLP-GLN reached the best MSE performance for both mean and median measures, whereas the MLP-Sin presented the smallest CV measure among these three models. 

It worth noting that the MLP-Sin was a better option than the MLP-Tanh. This result was already expected since trigonometric functions directly describe this differential equation's solution. However, the ANN functional approximator process was more efficient with the use of the MLP-GLN double activation function with local and global components.

\begin{table}[!htb]
    \centering
    \caption{Two-sample Kolmogorov-Smirnov Test at the $5\%$ significance level for the MSE distributions between the MLP-GLN and all other models for both architectures studied for solving the Catenary differential equation.}
    \label{tab:KS_Oscilator}
    \setlength{\extrarowheight}{5.5pt}
    \begin{tabular}{ccccc}
    \hline
    \multicolumn{3}{c}{\textbf{Tested Model}} & \multicolumn{2}{c}{\textbf{KS Test Results}}\\ 
    \cline{4-5}
    \multicolumn{3}{c}{\textbf{whit MLP-GLN}} & Statistically Similar & $p$-values\\
    \hline
    \multirow{3}{*}{\scriptsize \rotatebox[origin=c]{90}{$1-20-1$} }  &\multirow{3}{*}{\rotatebox[origin=c]{90}{MSE}}
    & MLP-Sin    & No & $0.005$ \\
    & & MLP-Tanh & No & $1.797\cdot10^{-14} $ \\
    & & TBN      & No & $0.011$ \\
    \hline
    \multirow{3}{*}{\scriptsize \rotatebox[origin=c]{90}{$1-20-20-1$} }  &\multirow{3}{*}{\rotatebox[origin=c]{90}{MSE}}
    & MLP-Sin    & No & $2.021\cdot10^{-4}$ \\
    & & MLP-Tanh & No & $1.797\cdot10^{-14}$ \\
    & & TBN      & No & $4.644\cdot10^{-6}$ \\
    \hline
    \end{tabular}
\end{table}

The MLP-GLN was tested against all other ANN models to verify if there is some ANN model with the same MSE population distribution. The two-sample KS test was applied, and the MLP-GLN presents a different MSE populational distribution. The two-sample KS test results at $5\%$ significance are shown in Table \ref{tab:KS_Oscilator}.

\begin{figure}
    \centering
     \begin{subfigure}[b]{0.47\textwidth}
         \centering
         \includegraphics[width=\textwidth]{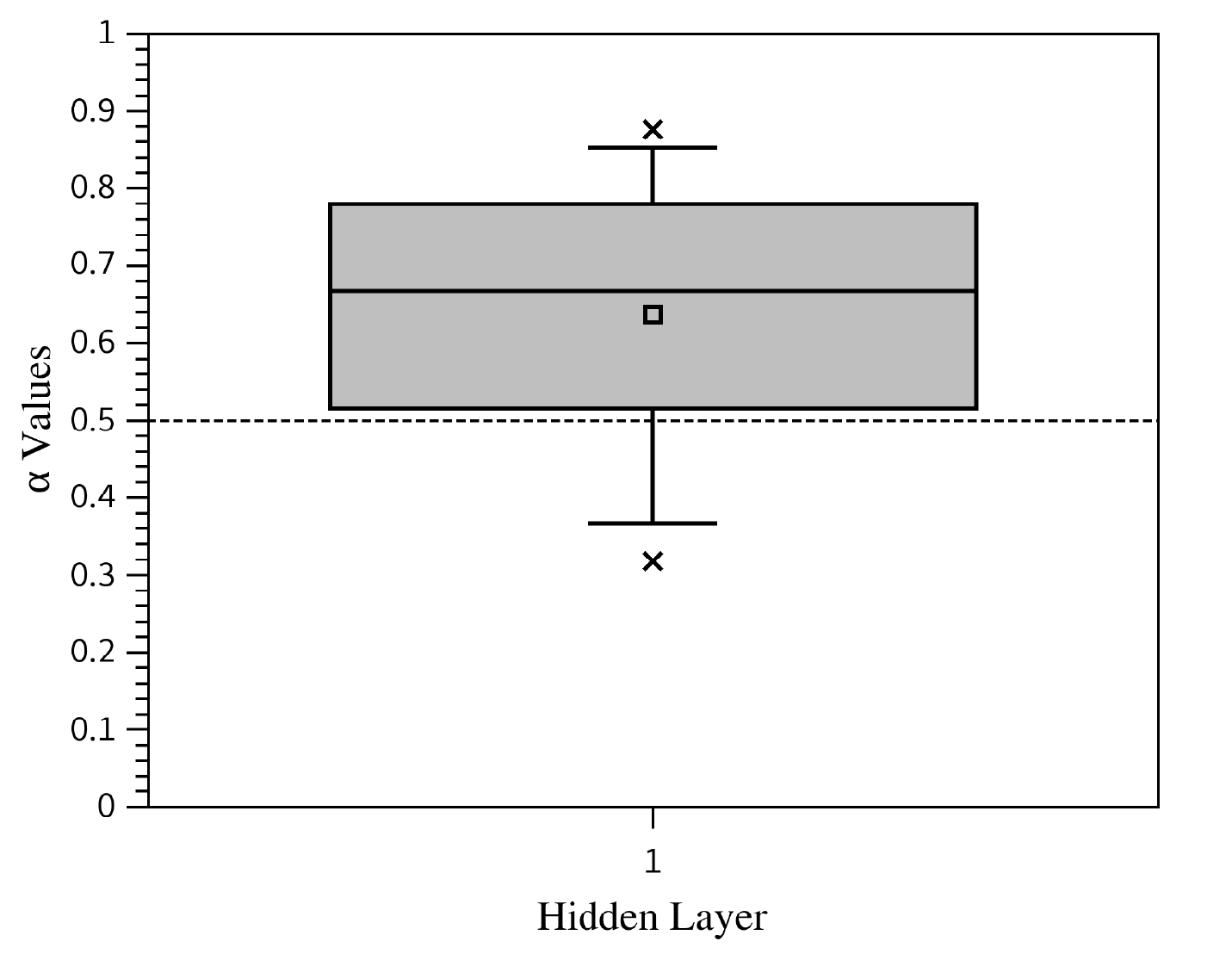}
         \caption{$1-20-1$ Architecture.}
         \label{fig:Oscillator_alphas_a}
     \end{subfigure}
     \hfill
     \begin{subfigure}[b]{0.47\textwidth}
         \centering
         \includegraphics[width=\textwidth]{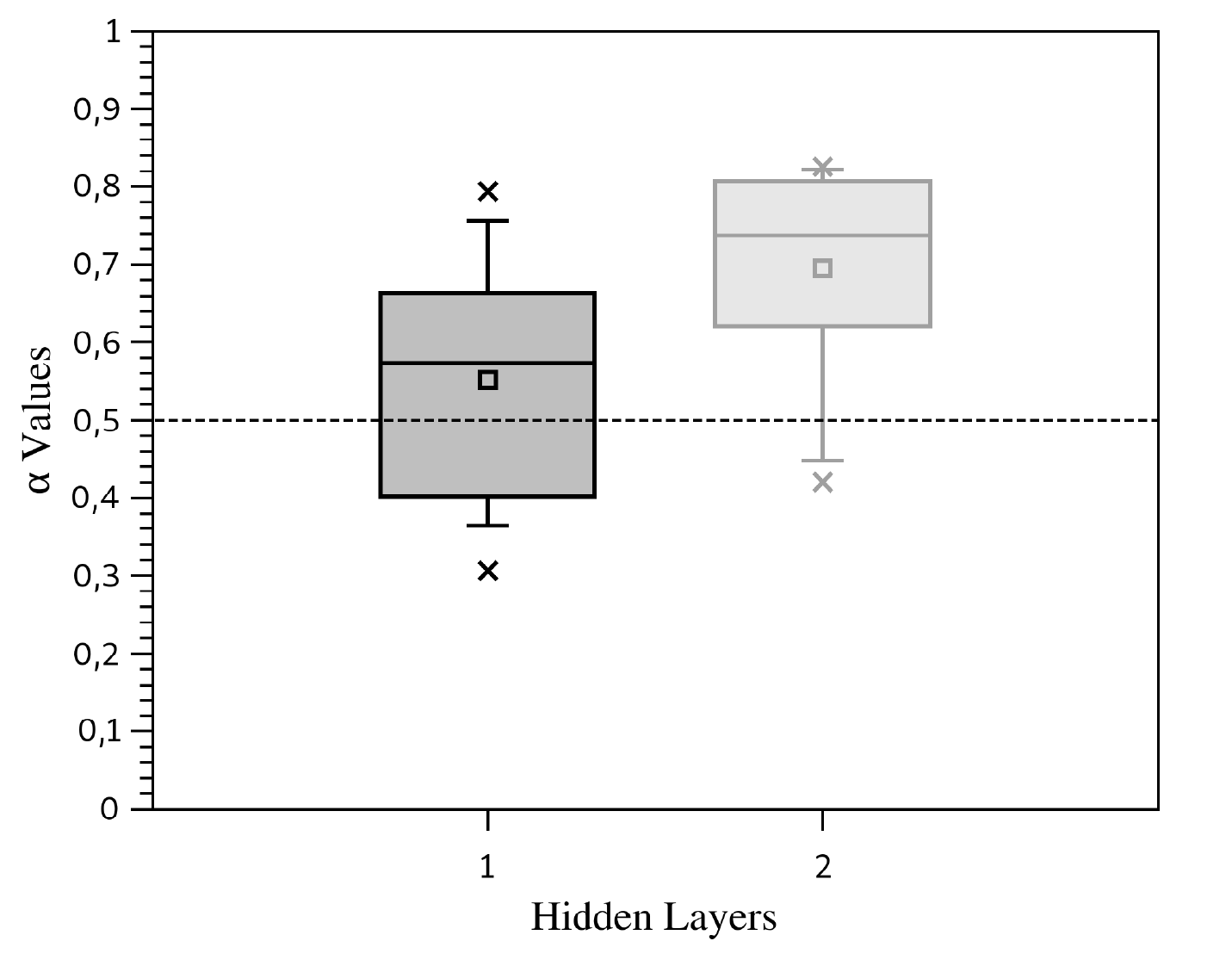}
         \caption{$1-20-20-1$ Architecture.}
         \label{fig:Oscillator_alphas_b}
     \end{subfigure}
     \caption{The $\alpha$ distributions for the MLP-GLN models after the training process for solving the Simple Harmonic Oscillator differential equation.}
     \label{fig:Oscillator_alphas}
\end{figure}

Figure \ref{fig:Oscillator_alphas} shows the $\alpha$ weight values distribution for both MLP-GLN architectures. Observing the Figure~\ref{fig:Oscillator_alphas_a}, the $\alpha$ values distribution reached by the architecture $1-20-1$ has a mean value of $0.637$, and a median value of $0.667$. In general, this architecture, with one hidden layer, gave more importance to the sine component, as it was expected. Looking at Figure \ref{fig:Oscillator_alphas_b} for the architecture $1-20-20-1$, the both hidden layers obtained mean $\alpha$ values greater than $0.5$ (mean values of $0.552$ and $0.695$, median values of $0.573$ and $0.737$, respectively). These mean values corroborate with the expected behavior, but it worth noting that the first hidden layer has a bigger contribution of the local component ($\tanh$) than the second hidden layer. This behavior can indicate that, like in the case of the catenary equation solving, the MLP-GLN sets the first hidden layer as a generic information representation and the second hidden layer as a specific or more appropriate information representation for the problem.  

\subsubsection{Damped Harmonic Oscillator Equation Results}

The differential equation for a damped harmonic oscillator, as defined in Section \ref{sec:DampedOsc}, has an analytical solution formed by a mathematical product of two functions, an exponential and a trigonometric function. All the ANN models were employed to solve the damped harmonic oscillator equation in the domain $t \in [0,6\pi]$, where it was applied the number of epochs of $1000$ to execute all ANN training processes.  

\begin{figure}[!htb]
     \centering
     \begin{subfigure}[b]{0.47\textwidth}
         \centering
         \includegraphics[width=\textwidth]{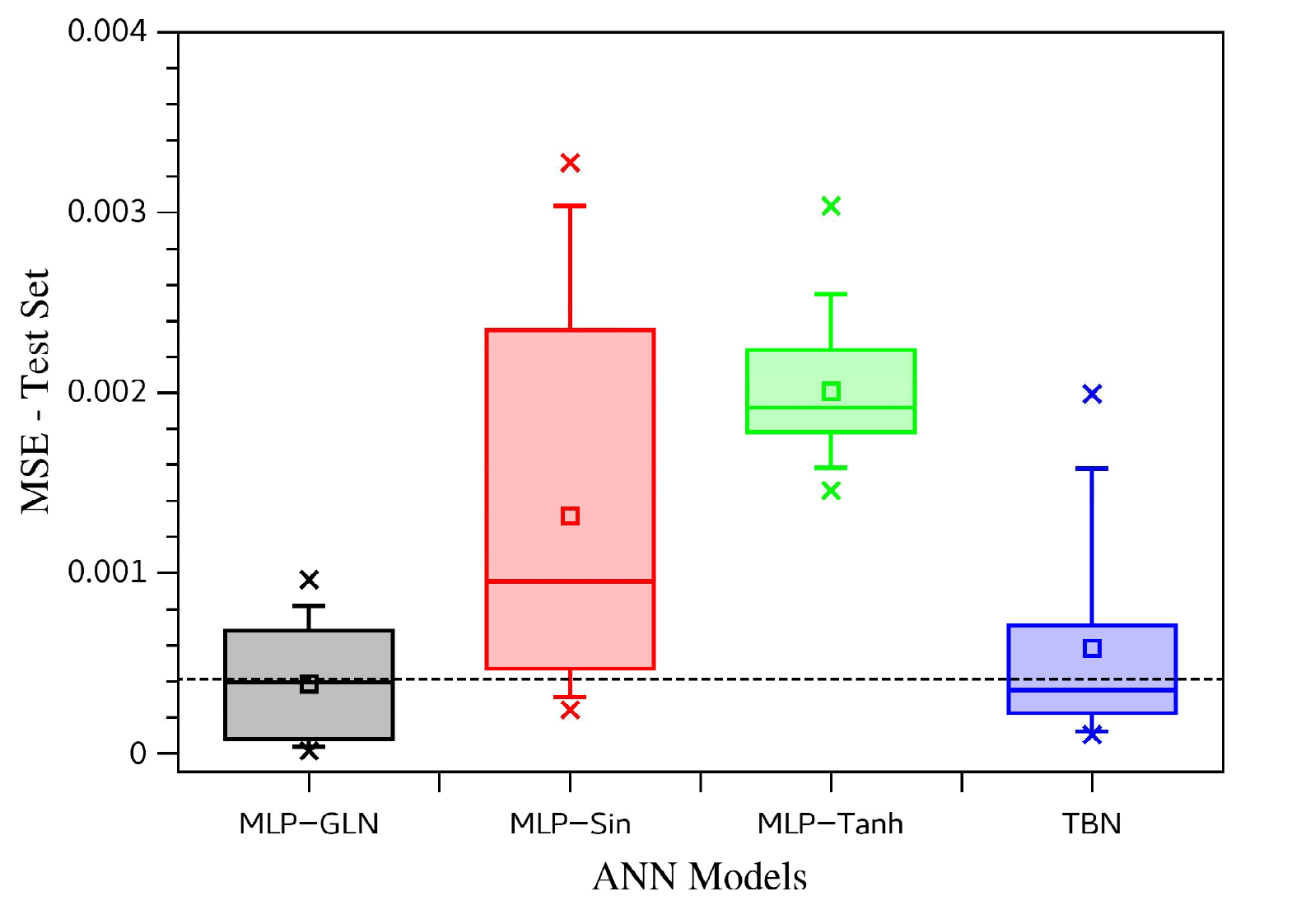}
         \caption{$1-20-1$ Architecture.}
         \label{fig:DampedTestMSE_a}
     \end{subfigure}
     \\
     \vspace{20pt}    
     \begin{subfigure}[b]{0.47\textwidth}
         \centering
         \includegraphics[width=\textwidth]{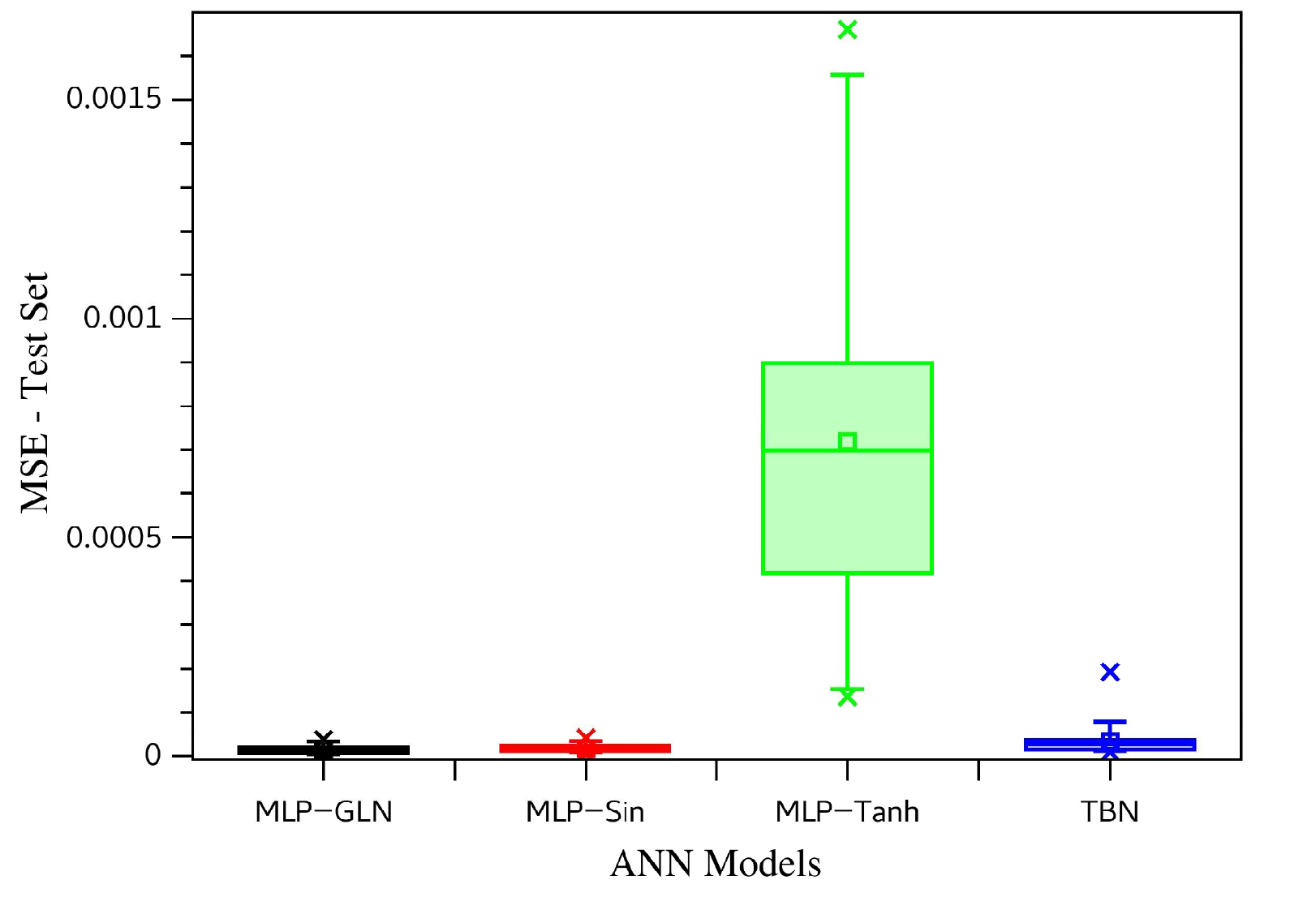}
         \caption{$1-20-20-1$ Architecture.}
         \label{fig:DampedTestMSE_b}
     \end{subfigure}
     \hfill
     \begin{subfigure}[b]{0.47\textwidth}
         \centering
         \includegraphics[width=\textwidth]{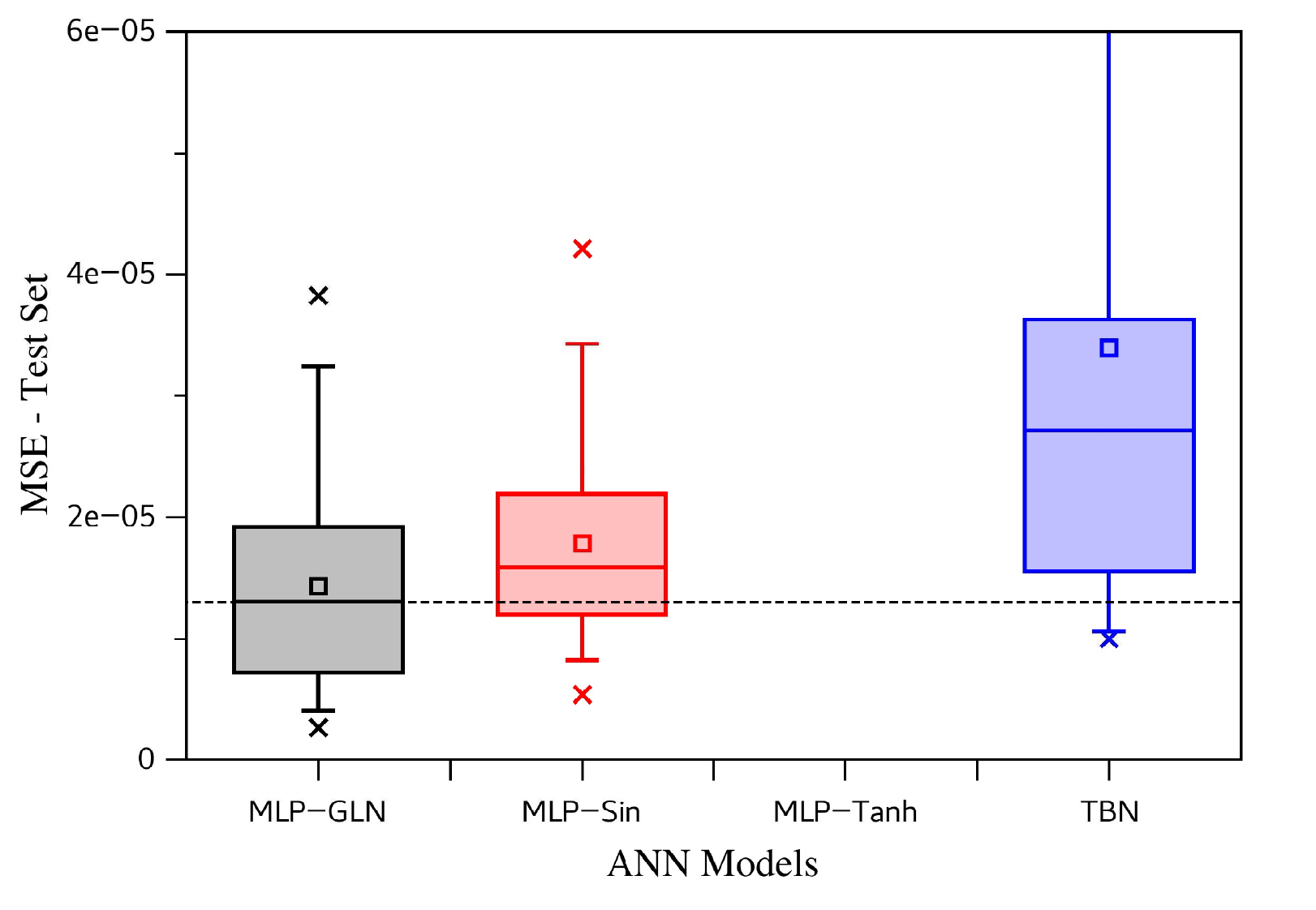}
         \caption{$1-20-20-1$ Architecture - Zoom scale.}
         \label{fig:DampedTestMSE_c}
     \end{subfigure}
    \caption{The MSE test set box-plot for all ANN models studied and both architectures for the Damped Harmonic Oscillator differential equation solving. In (a) is presented the MSE distribution for the 30 repetitions for each ANN model for architectures $1-20-1$. In (b) and (c) are shown the results for architecture $1-20-20-1$ in standard and zoom scale.} \label{fig:DampedTestMSE}
\end{figure}

Figure \ref{fig:DampedTestMSE} shows the MSE distributions for all ANN models. The MSE distributions for architecture $1-20-1$ are exhibited in Figure \ref{fig:DampedTestMSE_a}, and for architecture $1-20-20-1$ the MSE distributions are presented in Figures \ref{fig:DampedTestMSE_b}  and \ref{fig:DampedTestMSE_c}. All descriptive statistics for these MSE distributions are presented in Table \ref{tab:DampedMSE}.

\begin{table}[!htb]
    \centering
     \caption{The descriptive statistics for all ANN models analyzed. All MSE measures are relative to the Damped Harmonic Oscillator differential equation solving. The best results are highlighted in bold-face.}
    \label{tab:DampedMSE}
    \begin{tabular}{cccccc}
    \hline
    \multicolumn{2}{c}{\multirow{2}{*}{\textbf{MSE}}} & \multicolumn{4}{c}{\textbf{ANN Models}}\\ 
    \cline{3-6}
     & & MLP-GLN & MLP-Sin & MLP-Tanh & TBN \\
    \hline
    \multirow{6}{*}{\rotatebox[origin=c]{90}{$1-20-1$}} 
    & Min.   & $\mathbf{1.312\cdot10^{-5}}$ & $2.414\cdot10^{-4}$ & $1.457\cdot10^{-3}$ & $1.051\cdot10^{-4}$ \\
    & Max.   & $\mathbf{9.645\cdot10^{-4}}$ & $3.275\cdot10^{-3}$ & $3.034\cdot10^{-3}$ & $1.994\cdot10^{-3}$ \\
    & Mean   & $\mathbf{3.891\cdot10^{-4}}$ & $1.317\cdot10^{-3}$ & $2.009\cdot10^{-3}$ & $5.845\cdot10^{-4}$ \\
    & Median & $3.945\cdot10^{-4}$ & $9.568\cdot10^{-4}$ & $1.917\cdot10^{-3}$ & $\mathbf{3.505\cdot10^{-4}}$ \\
    & Std.   & $\mathbf{3.051\cdot10^{-4}}$ & $1.002\cdot10^{-3}$ & $3.456\cdot10^{-4}$ & $5.241\cdot10^{-4}$ \\
    & CV     & $0.784$ & $0.760$ & $\mathbf{0.172}$ & $0.897$ \\
    \hline
    \multirow{6}{*}{\rotatebox[origin=c]{90}{$1-20-20-1$}}  
    & Min.   & $\mathbf{2.669\cdot10^{-6}}$ & $5.383\cdot10^{-6}$ & $1.362\cdot10^{-4}$ & $9.993\cdot10^{-6}$ \\
    & Max.   & $\mathbf{3.823\cdot10^{-5}}$ & $4.213\cdot10^{-5}$ & $1.660\cdot10^{-3}$ & $1.926\cdot10^{-4}$ \\
    & Mean   & $\mathbf{1.432\cdot10^{-5}}$ & $1.784\cdot10^{-5}$ & $7.184\cdot10^{-4}$ & $3.394\cdot10^{-5}$ \\
    & Median & $\mathbf{1.303\cdot10^{-5}}$ & $1.588\cdot10^{-5}$ & $6.983\cdot10^{-4}$ & $2.713\cdot10^{-5}$ \\
    & Std.   & $9.484\cdot10^{-6}$ & $\mathbf{8.684\cdot10^{-6}}$ & $4.161\cdot10^{-4}$ & $3.471\cdot10^{-5}$ \\
    & CV     & $0.662$ & $\mathbf{0.487}$ & $0.579$ & $1.023$ \\
    \hline
    \end{tabular}
\end{table}

For the architecture $1-20-1$, the two best ANN models were the MLP-GLN and the TBN, demonstrating that the local and global function combinations have a better performance than the ANN with a single activation function for this problem. The MLP-GLN reached the best MSE mean value, while the TBN had better MSE median result. Both these ANN models had very similar statistical central measures results, but the MLP-GLN had a CV value of $0.784$ against a TBN CV value of $0.897$. In this way, looking at Figure \ref{fig:DampedTestMSE_a}, the MLP-GLN is the best option given its smaller dispersion. However, as shown in Table \ref{tab:KS_Damped}, the two-sample KS test, at $5\%$ significance level, indicates that both ANN models are equivalent. By the KS test, the MLP-GLN and TBN MSE distributions come than the same MSE populational distribution.    

For the architecture $1-20-20-1$, three ANN models stood out, the MLP-GLN, MLP-Sin, and TBN, as it can be viewed in Figure \ref{fig:DampedTestMSE_b}. The MLP-GLN reached the better MSE performance from these three ANN models, as viewed in Table \ref{tab:DampedMSE}. For this architecture, the MLP-GLN has a statistically different MSE distribution, as the two-sample KS test affirms in Table \ref{tab:KS_Damped}. Thus, the MLP-GLN reached better MSE performance from these three highlighted ANN models. 

\begin{table}[!htb]
    \centering
    \caption{Two-sample Kolmogorov-Smirnov Test at the $5\%$ significance level for the MSE distributions between the MLP-GLN and all other models for both architectures studied for the Damped Harmonic Oscillator differential equation solving.}
    \label{tab:KS_Damped}
    \setlength{\extrarowheight}{5.5pt}
    \begin{tabular}{ccccc}
    \hline
    \multicolumn{3}{c}{\textbf{Tested Model}} & \multicolumn{2}{c}{\textbf{KS Test Results}}\\ 
    \cline{4-5}
    \multicolumn{3}{c}{\textbf{whit MLP-GLN}} & Statistically Similar & $p$-values\\
    \hline
    \multirow{3}{*}{\scriptsize \rotatebox[origin=c]{90}{$1-20-1$} }  &\multirow{3}{*}{\rotatebox[origin=c]{90}{MSE}}
    & MLP-Sin    & No  & $2.021\cdot10^{-4}$ \\
    & & MLP-Tanh & No  & $1.797\cdot10^{-14}$ \\
    & & TBN      & Yes & $0.055$ \\
    \hline
    \multirow{3}{*}{\scriptsize \rotatebox[origin=c]{90}{$1-20-20-1$} }  &\multirow{3}{*}{\rotatebox[origin=c]{90}{MSE}}
    & MLP-Sin    & No & $0.026$ \\
    & & MLP-Tanh & No & $1.797\cdot10^{-14}$ \\
    & & TBN      & No & $0.002$ \\
    \hline
    \end{tabular}
\end{table}

\begin{figure}[!htb]
    \centering
     \begin{subfigure}[b]{0.47\textwidth}
         \centering
         \includegraphics[width=\textwidth]{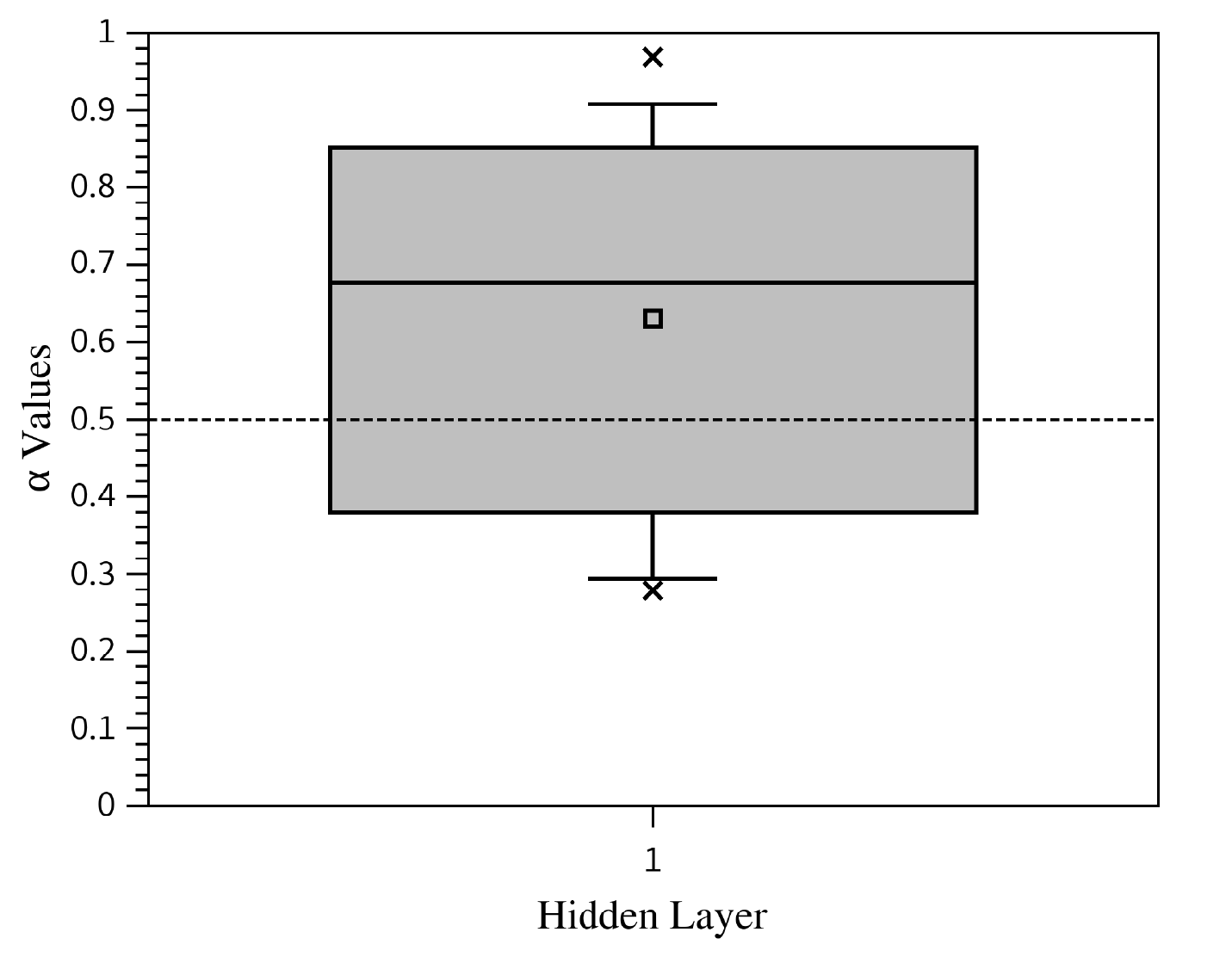}
         \caption{$1-20-1$ Architecture.}
         \label{fig:Damped_alphas_a}
     \end{subfigure}
     \hfill
     \begin{subfigure}[b]{0.47\textwidth}
         \centering
         \includegraphics[width=\textwidth]{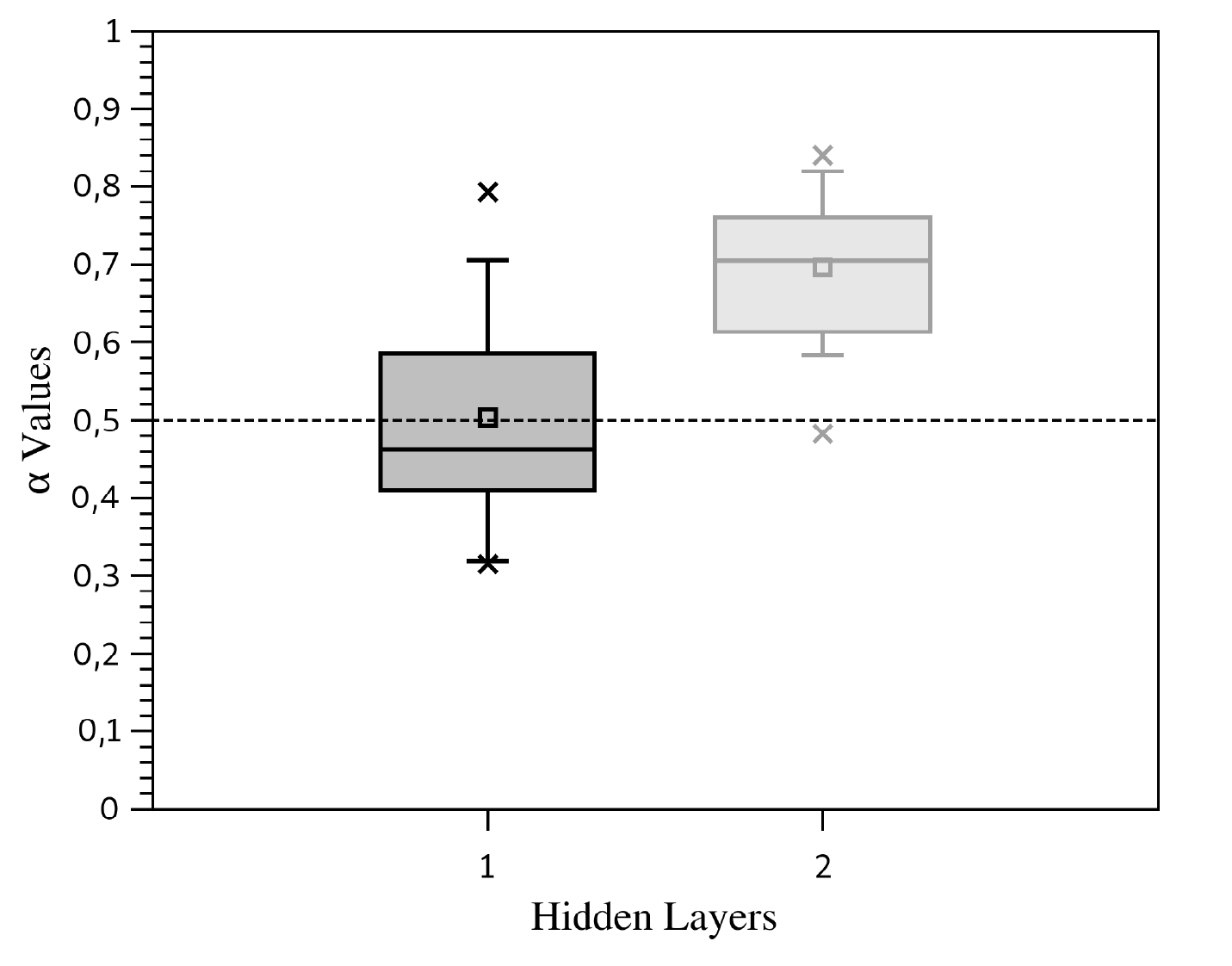}
         \caption{$1-20-20-1$ Architecture.}
         \label{fig:Damped_alphas_b}
     \end{subfigure}
     \caption{The $\alpha$ distributions for the MLP-GLN models after the training process for the Damped Harmonic Oscillator differential equation solving.}
     \label{fig:Damped_alphas}
\end{figure}

The $\alpha$ values distribution for the MLP-GLN can be viewed in Figure \ref{fig:Damped_alphas}. For the architecture with a hidden layer, Figure \ref{fig:Damped_alphas_a} shows an $\alpha$ distribution with mean value of $0.631$, and median of $0.676$. In this case, the MLP-GLN, in general, gives more importance to the global component than the local component. For the architecture $1-20-20-1$, Figure \ref{fig:Damped_alphas_b}, the first hidden layer has given approximately the same importance for both local and global components, although with a very slight tendency for the local component, with an $\alpha$ mean value of $0.503$, and a median value of $0.462$. For the second hidden layer, the global component had more important than the local component, presenting an $\alpha$ values distribution with a mean of $0.696$ and a median of $0.705$.

Let us look at the $\alpha$ values distributions for both damped and simple harmonic oscillators. It is possible to note that, for the $1-20-1$ case, there was a modification of the distribution shape. The mean and median values for both differential equation cases have the same behavior, but for the damped oscillator case the $\alpha$ distribution presented an accentuated asymmetric, extending its tail for smaller values of $\alpha$. This behavior indicates that the MLP-GLN introduces a more significant contribution to the local component in the activation function composition for the damped harmonic oscillator. This behavior makes sense because there is an exponential component in the damped oscillator solution, whereas for a simple oscillator, there is only the sine component. For the architecture $1-20-20-1$, the $\alpha$ values distributions for both damped and simple oscillators have similar behavior. However, for the damped oscillator ANN solver, the $\alpha$ distribution has less dispersion, indicating a finer specialization for each hidden layer, where the hyperbolic tangent contribution was increased in the first hidden layer and the sine contribution in the second hidden layer. One more time, this behavior of the $\alpha$ values points to more local information representation in the first hidden layer and a more global representation in the second hidden layer. 

\subsubsection{Laplace Equation Results}

The Laplace differential equation has an analytical solution given by Equation \ref{eqn:AnaSolLaplace}, as defined in Section \ref{sec:Laplace}. This solution is a mathematical product between sine and hyperbolic sines. The analytical solution is a $2$-dimensional function, $u(x,y)$. The mathematical domain analyzed here was $x \in [-2,2]$ and $y \in [-2,2]$. For all experiments, it was employed $300$ epochs in the training process.

\begin{figure}
     \centering
     \begin{subfigure}[b]{0.47\textwidth}
         \centering
         \includegraphics[width=\textwidth]{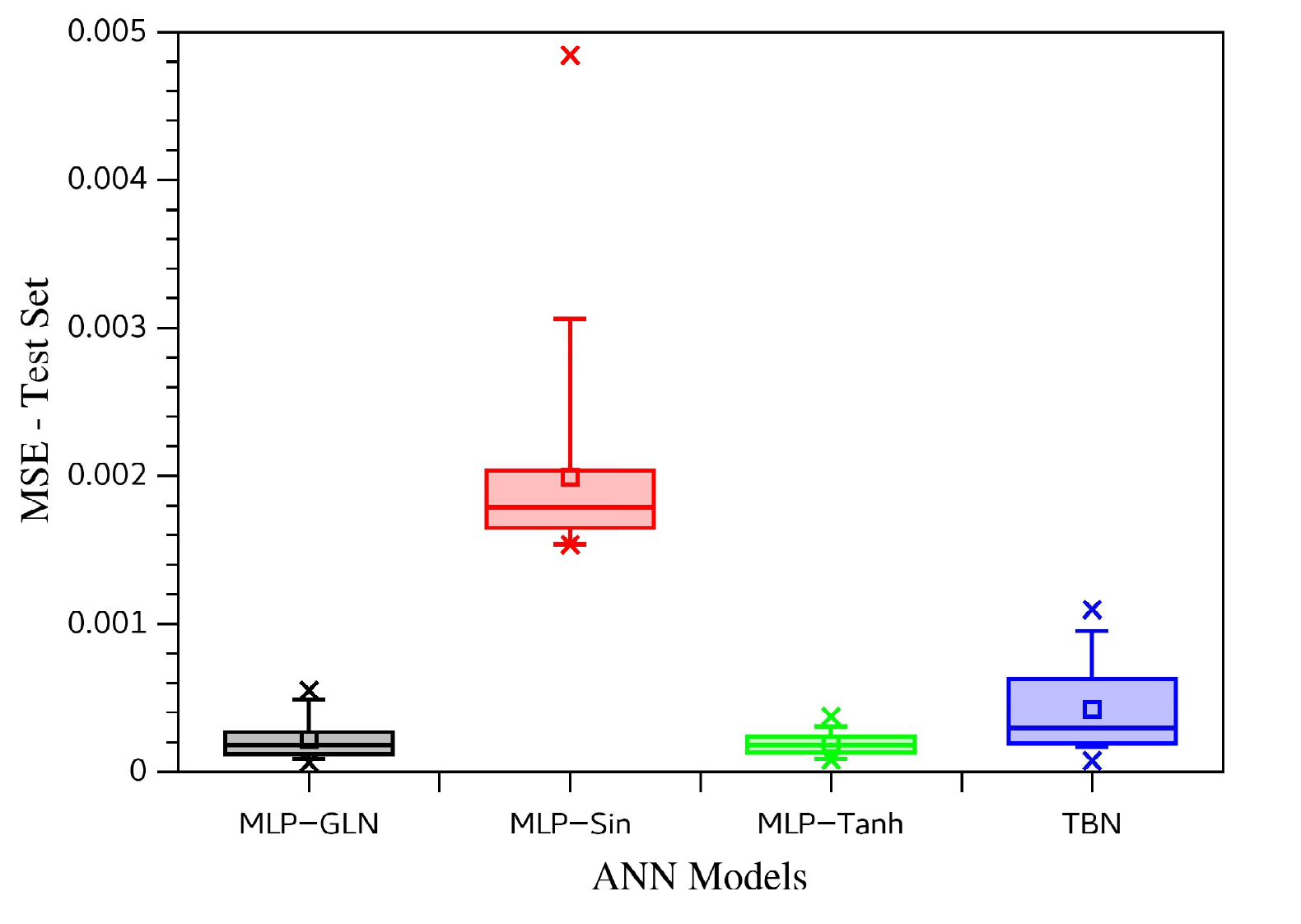}
         \caption{$1-20-1$ Architecture.}
         \label{fig:LaplaceTestMSE_a}
     \end{subfigure}
     \hfill
     \begin{subfigure}[b]{0.47\textwidth}
         \centering
         \includegraphics[width=\textwidth]{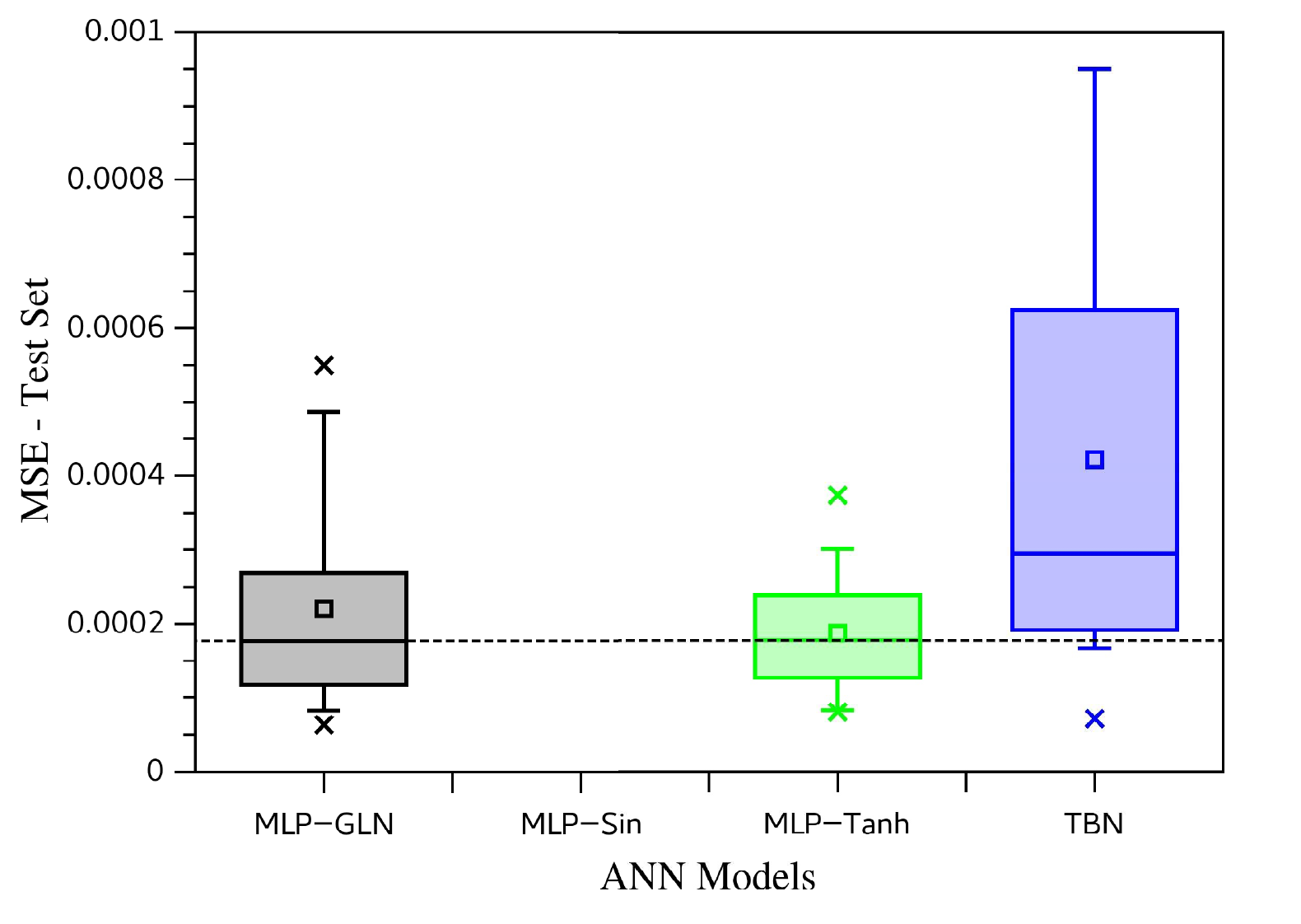}
         \caption{$1-20-1$ Architecture - Zoom scale.}
         \label{fig:LaplaceTestMSE_b}
     \end{subfigure}
     \vspace{20pt}    
     \begin{subfigure}[b]{0.47\textwidth}
         \centering
         \includegraphics[width=\textwidth]{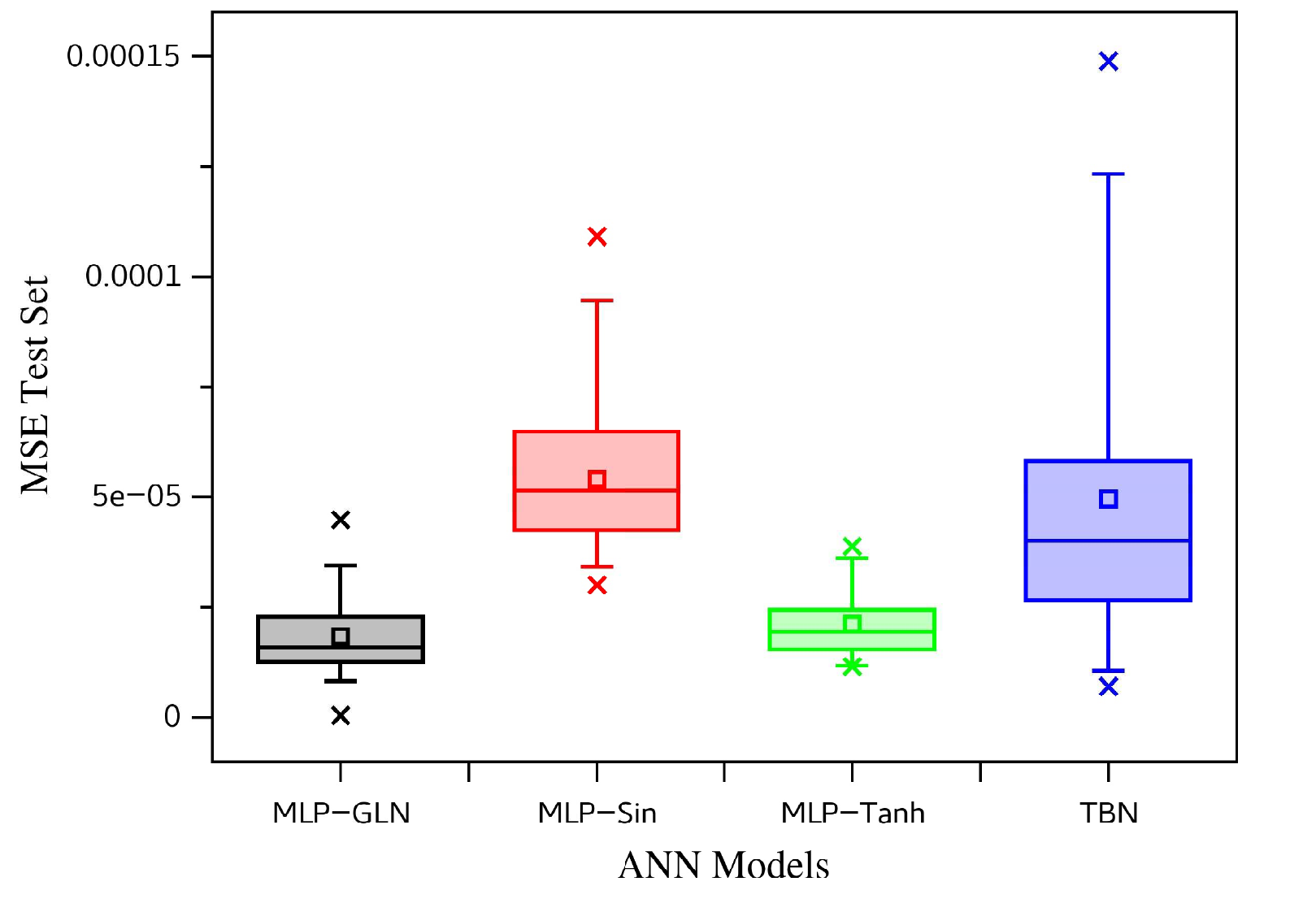}
         \caption{$1-20-20-1$ Architecture.}
         \label{fig:LaplaceTestMSE_c}
     \end{subfigure}
     \hfill
     \begin{subfigure}[b]{0.47\textwidth}
         \centering
         \includegraphics[width=\textwidth]{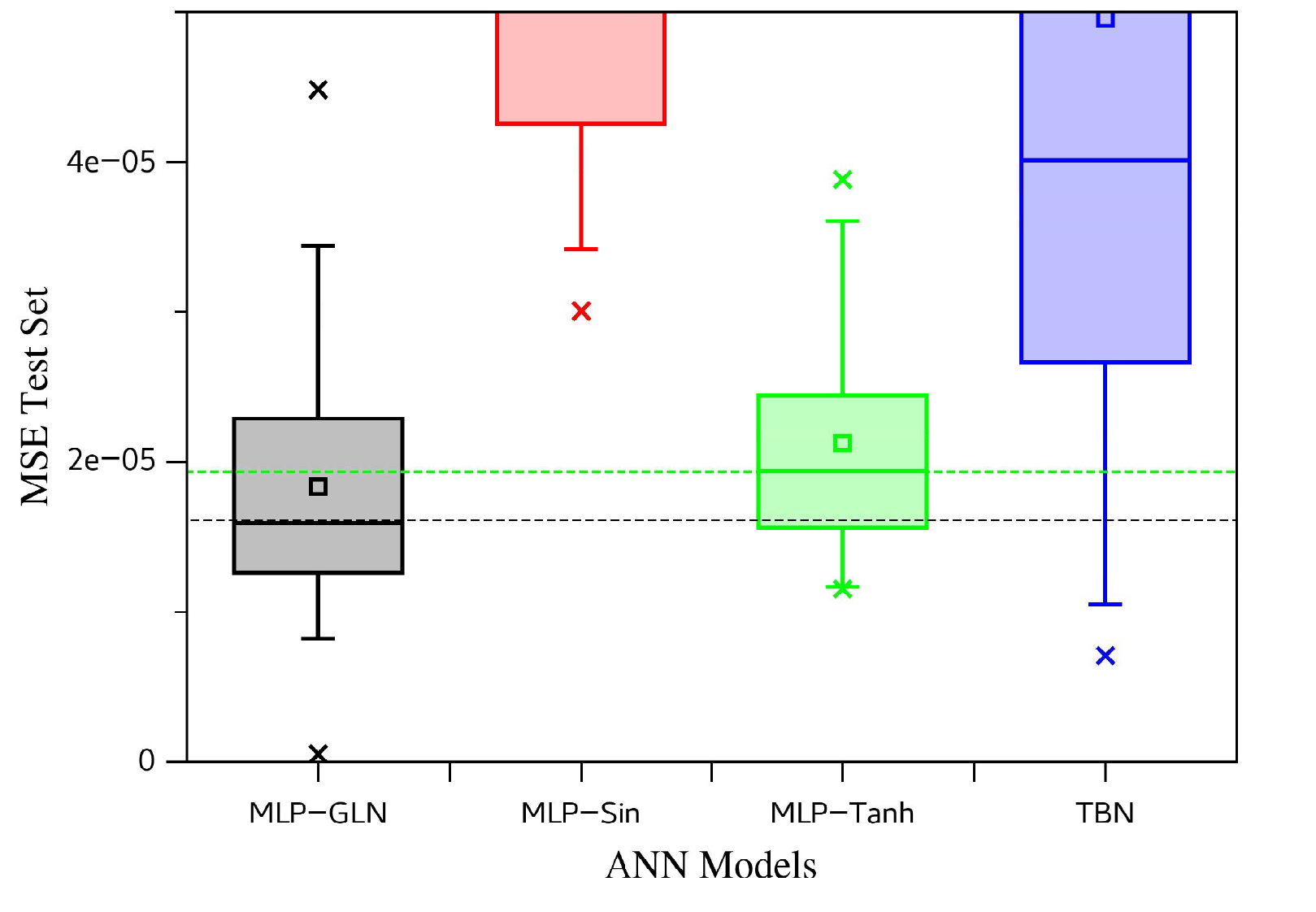}
         \caption{$1-20-20-1$ Architecture - Zoom scale.}
         \label{fig:LaplaceTestMSE_d}
     \end{subfigure}
    \caption{The test set MSE box-plot for all ANN models studied and both architectures for the Laplace differential equation solving. In (a) and (b) are presented the MSE distributions for the 30 repetitions for each ANN model for architectures $1-20-1$. In (c) and (d) is shown the results for architecture $1-20-20-1$. The dashed lines are the median values references.} \label{fig:LaplaceTestMSE}
\end{figure}

Figure \ref{fig:LaplaceTestMSE} shows the MSE distributions for all tested ANN models. Figures \ref{fig:LaplaceTestMSE_a} and \ref{fig:LaplaceTestMSE_b} highlight the MSE distribution for the experiment with the architecture $1-20-1$. It is possible to note that the MLP-GLN and MLP-Tanh are the two models with the best MSE performance. The MLP-GLN reached an MSE distribution with a mean value of $2.208 \cdot 10^{-4}$ and a median of $1.762 \cdot 10^{-4}$,  the MLP-Tanh a mean of $1.873 \cdot 10^{-4}$ and a media of $1.773 \cdot 10^{-4}$. All descriptive statistics are presented in Table \ref{tab:LaplaceMSE}. Looking at Figure \ref{fig:LaplaceTestMSE_b}, apparently the MLP-Tanh has better MSE performance, since its MSE distribution CV is $0.382$, while the CV of the MLP-GLN is $0.602$. However, applying the two-sample KS test, as presented in Table \ref{tab:KS_Laplace}, the conclusion is the MLP-GLN and MLP-Tanh have the same  MSE populational distribution. The MLP-GLN and MLP-Tanh have statistically the same behavior.

\begin{table}[!htb]
    \centering
     \caption{The descriptive statistics for all ANN models analyzed. All MSE measures are relative to the Laplace differential equation solving. The best results are highlighted in bold-face.}
    \label{tab:LaplaceMSE}
    \begin{tabular}{cccccc}
    \hline
    \multicolumn{2}{c}{\multirow{2}{*}{\textbf{MSE}}} & \multicolumn{4}{c}{\textbf{ANN Models}}\\ 
    \cline{3-6}
     & & MLP-GLN & MLP-Sin & MLP-Tanh & TBN \\
    \hline
    \multirow{6}{*}{\rotatebox[origin=c]{90}{$1-20-1$}} 
    & Min.   & $\mathbf{6.304\cdot10^{-5}}$ & $1.535\cdot10^{-3}$ & $8.083\cdot10^{-5}$ & $7.186\cdot10^{-5}$ \\
    & Max.   & $5.497\cdot10^{-4}$ & $4.844\cdot10^{-3}$ & $\mathbf{3.737\cdot10^{-4}}$ & $1.096\cdot10^{-3}$ \\
    & Mean   & $2.208\cdot10^{-4}$ & $1.990\cdot10^{-3}$ & $\mathbf{1.873\cdot10^{-4}}$ & $4.217\cdot10^{-4}$ \\
    & Median & $\mathbf{1.761\cdot10^{-4}}$ & $1.790\cdot10^{-3}$ & $1.773\cdot10^{-4}$ & $2.949\cdot10^{-4}$ \\
    & Std.   & $1.329\cdot10^{-4}$ & $6.659\cdot10^{-4}$ & $\mathbf{7.156\cdot10^{-5}}$ & $2.937\cdot10^{-4}$ \\
    & CV     & $0.602$ & $\mathbf{0.335}$ & $0.382$ & $0.696$ \\
    \hline
    \multirow{6}{*}{\rotatebox[origin=c]{90}{$1-20-20-1$}}  
    & Min.   & $\mathbf{4.945\cdot10^{-7}}$ & $3.007\cdot10^{-5}$ & $1.152\cdot10^{-5}$ & $7.084\cdot10^{-6}$ \\
    & Max.   & $4.486\cdot10^{-5}$ & $1.092\cdot10^{-4}$ & $\mathbf{3.884\cdot10^{-5}}$ & $1.489\cdot10^{-4}$ \\
    & Mean   & $\mathbf{1.834\cdot10^{-5}}$ & $5.414\cdot10^{-5}$ & $2.124\cdot10^{-5}$ & $4.958\cdot10^{-5}$ \\
    & Median & $\mathbf{1.590\cdot10^{-5}}$ & $5.140\cdot10^{-5}$ & $1.939\cdot10^{-5}$ & $4.009\cdot10^{-5}$ \\
    & Std.   & $9.408\cdot10^{-6}$ & $\mathbf{1.921\cdot10^{-5}}$ & $7.597\cdot10^{-6}$ & $3.572\cdot10^{-5}$ \\
    & CV     & $0.513$ & $\mathbf{0.355}$ & $0.358$ & $0.720$ \\
    \hline
    \end{tabular}
\end{table}

\begin{table}[!htb]
    \centering
    \caption{Two-sample Kolmogorov-Smirnov Test at the $5\%$ significance level for the MSE distributions between the MLP-GLN and all other models for both architectures studied for the Laplace differential equation solving.}
    \label{tab:KS_Laplace}
    \setlength{\extrarowheight}{5.5pt}
    \begin{tabular}{ccccc}
    \hline
    \multicolumn{3}{c}{\textbf{Tested Model}} & \multicolumn{2}{c}{\textbf{KS Test Results}}\\ 
    \cline{4-5}
    \multicolumn{3}{c}{\textbf{whit MLP-GLN}} & Statistically Similar & $p$-values\\
    \hline
    \multirow{3}{*}{\scriptsize \rotatebox[origin=c]{90}{$1-20-1$} }  &\multirow{3}{*}{\rotatebox[origin=c]{90}{MSE}}
    & MLP-Sin    & No & $1.797\cdot10^{-14}$ \\
    & & MLP-Tanh & Yes & $0.760$ \\
    & & TBN      & No & $0.0259$ \\
    \hline
    \multirow{3}{*}{\scriptsize \rotatebox[origin=c]{90}{$1-20-20-1$} }  &\multirow{3}{*}{\rotatebox[origin=c]{90}{MSE}}
    & MLP-Sin    & No & $8.384\cdot10^{-12}$ \\
    & & MLP-Tanh & Yes & $0.055$ \\
    & & TBN      & No & $1.755\cdot10^{-5}$ \\
    \hline
    \end{tabular}
\end{table}

The MSE distributions for the architecture $1-20-20-1$ are presented in Figures \ref{fig:LaplaceTestMSE_c} and \ref{fig:LaplaceTestMSE_d}. The two best ANN models were the MLP-GLN and MLP-Tanh. As shown in Table \ref{tab:LaplaceMSE}, the MLP-GLN reached better mean and median MSE distribution values than the MLP-Tanh. However, one more time, both ANN models' MSE distributions come from the same population distribution as indicated by the two-sample KS test Table \ref{tab:KS_Laplace}. In this way, the MLP-GLN and MLP-Tanh are statistically similar for the MSE performance for both architectures. 

\begin{figure}
    \centering
     \begin{subfigure}[b]{0.47\textwidth}
         \centering
         \includegraphics[width=\textwidth]{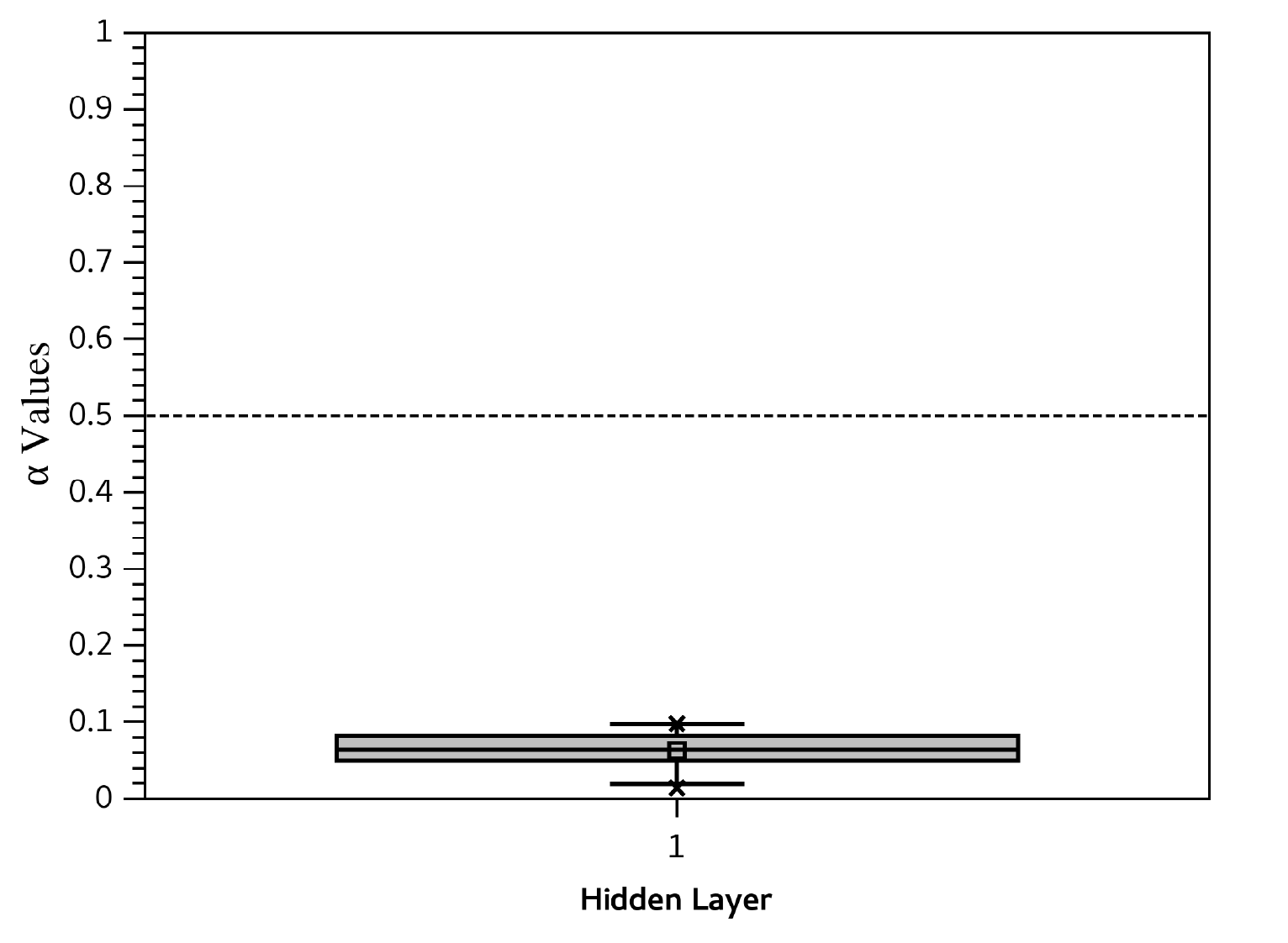}
         \caption{$1-20-1$ Architecture.}
         \label{fig:Laplace_alphas_a}
     \end{subfigure}
     \hfill
     \begin{subfigure}[b]{0.47\textwidth}
         \centering
         \includegraphics[width=\textwidth]{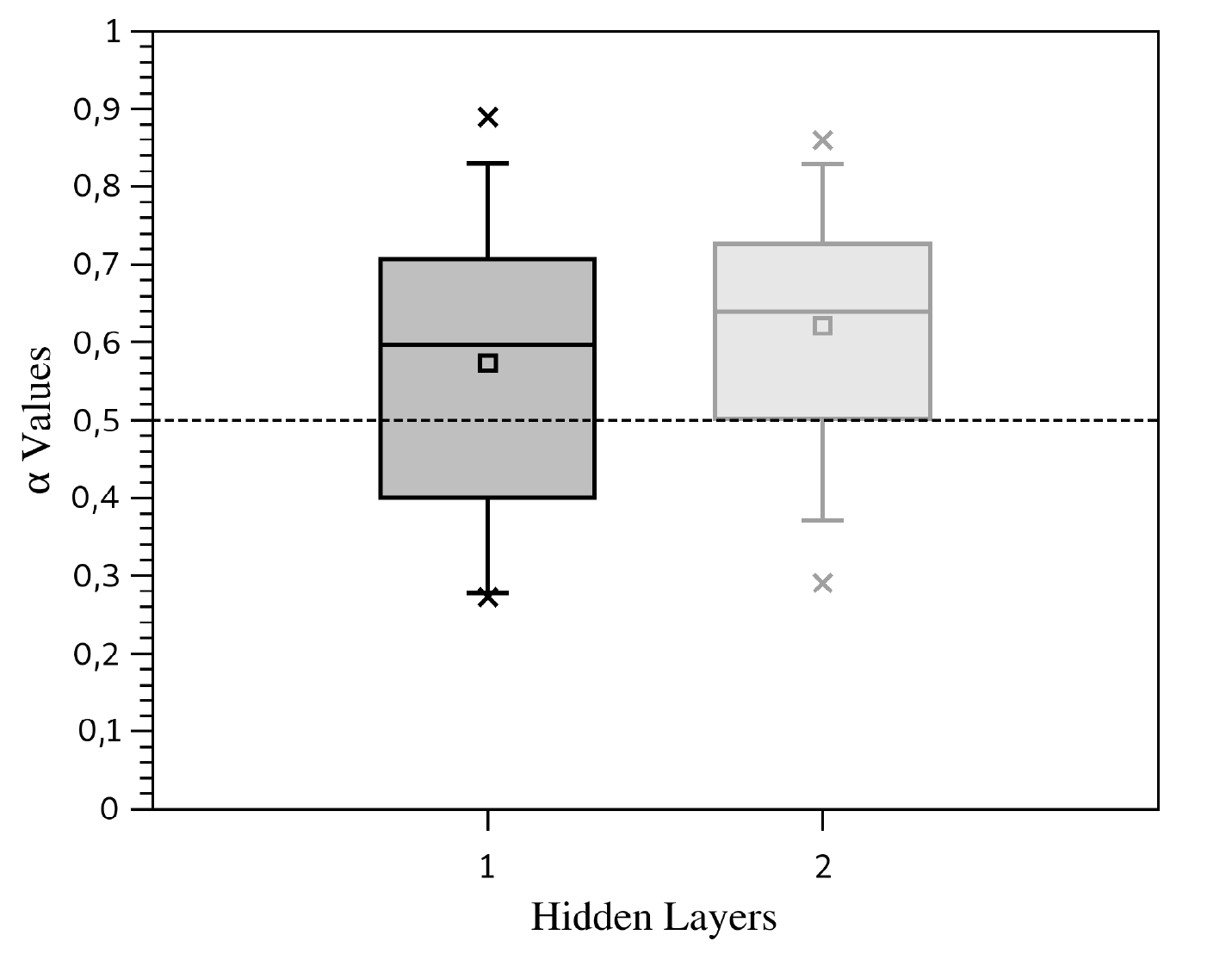}
         \caption{$1-20-20-1$ Architecture.}
         \label{fig:Laplace_alphas_b}
     \end{subfigure}
     \caption{The $\alpha$ distributions for the MLP-GLN models after the training process for the Laplace differential equation solving.}
     \label{fig:Laplace_alphas}
\end{figure}

Looking at Figure \ref{fig:Laplace_alphas}, it is possible to observe the $\alpha$ values distribution reached by MLP-GLN for both architectures. Figure \ref{fig:Laplace_alphas_a} for architecture $1-20-1$, and Figure \ref{fig:Laplace_alphas_b} for architecture $1-20-20-1$. For the architecture with one hidden layer, the $\alpha$ values distribution obtained both mean and median values of $0.063$. These values demonstrate that the MLP-GLN has selected almost only the local component, justifying similar MLP-Tanh like behavior. However, for the architecture, $1-20-20-1$, the $\alpha$ values distribution was dominated by values greater than $0.5$. For both hidden layers, the global component of the activation function had more importance than the local component, where the mean and median values for the first hidden layer were $0.573$ and $0.596$, and for the second hidden layer were $0.621$ and $0.639$. Although the MLP-GLN converges to a model where the sine function is dominant, the MSE performance reached by MLP-GLN was statistically similar to the MLP-Tanh, indicating the GLN versatility.

\subsubsection{Heat Equation Results}

The heat differential equation defined in Section \ref{sec:Heat} with respective boundary conditions, has a possible analytical solution presented in Equation \ref{eqn:solHeat}. This solution is a mathematical product of the sine and exponential functions, where in the context of this article, a solution with global and local components. For the experiments, the parameters used were a medium diffusivity $k=0.3$, a size of the heat propagation medium $L=2.0$, a time-domain of $t \in [0,2]$ and a space-domain of $x \in [0,L]$. All ANN models were trained with the number of epochs of $300$.

\begin{figure}
     \centering
     \begin{subfigure}[b]{0.47\textwidth}
         \centering
         \includegraphics[width=\textwidth]{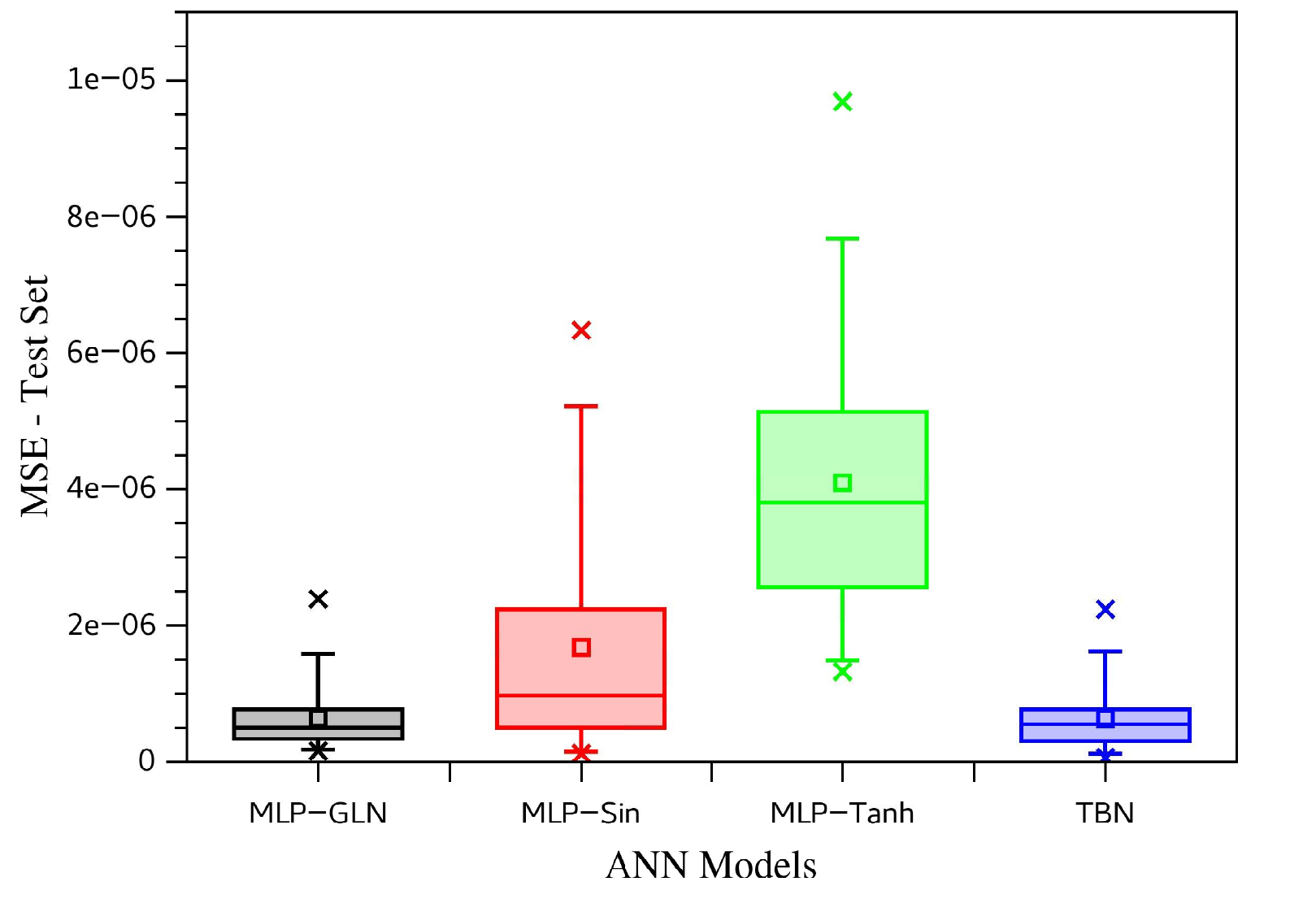}
         \caption{$1-20-1$ Architecture.}
         \label{fig:HeatTestMSE_a}
     \end{subfigure}
     \hfill
     \begin{subfigure}[b]{0.47\textwidth}
         \centering
         \includegraphics[width=\textwidth]{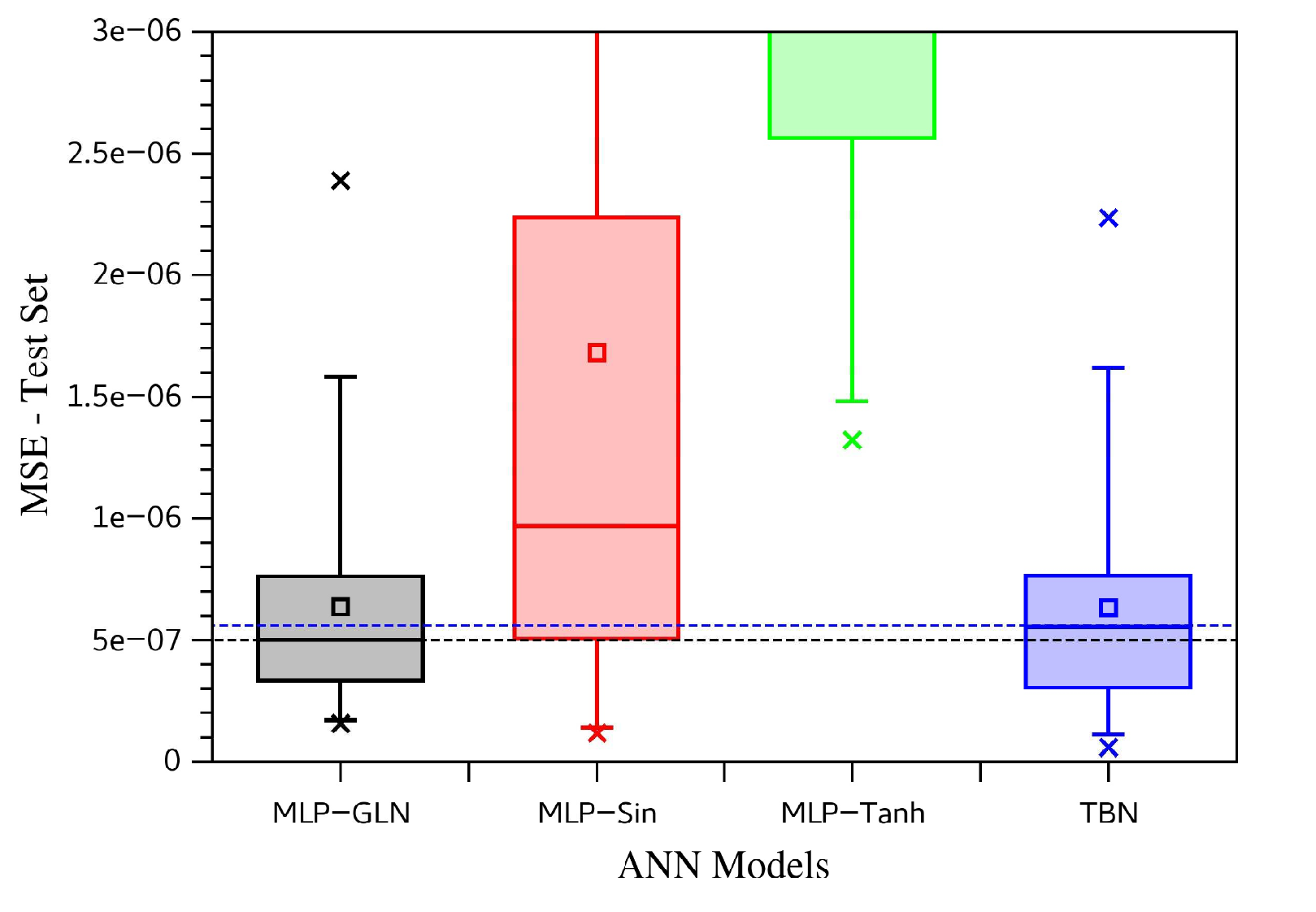}
         \caption{$1-20-1$ Architecture - Zoom scale.}
         \label{fig:HeatTestMSE_b}
     \end{subfigure}
     \vspace{20pt}    
     \begin{subfigure}[b]{0.47\textwidth}
         \centering
         \includegraphics[width=\textwidth]{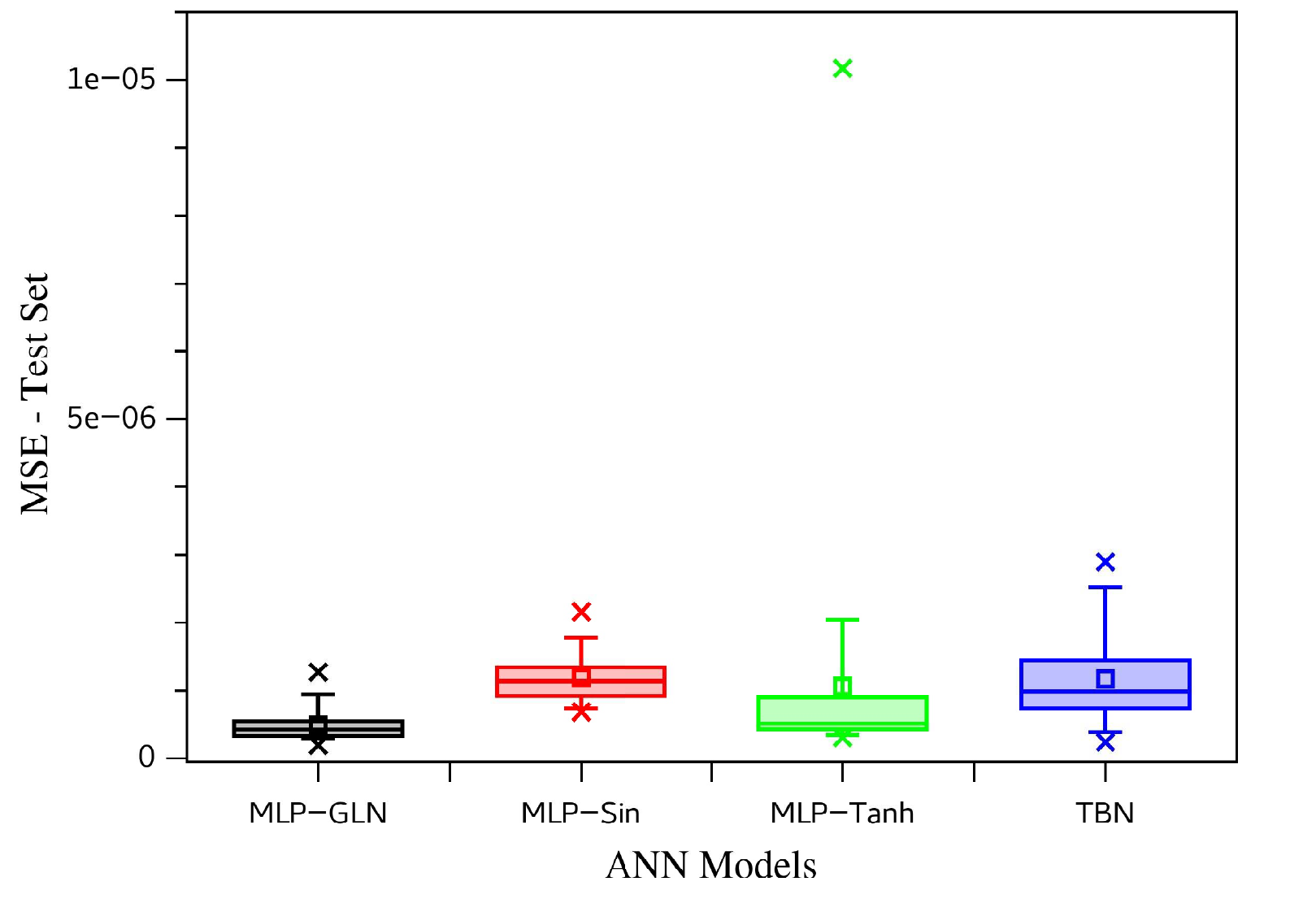}
         \caption{$1-20-20-1$ Architecture.}
         \label{fig:HeatTestMSE_c}
     \end{subfigure}
     \hfill
     \begin{subfigure}[b]{0.47\textwidth}
         \centering
         \includegraphics[width=\textwidth]{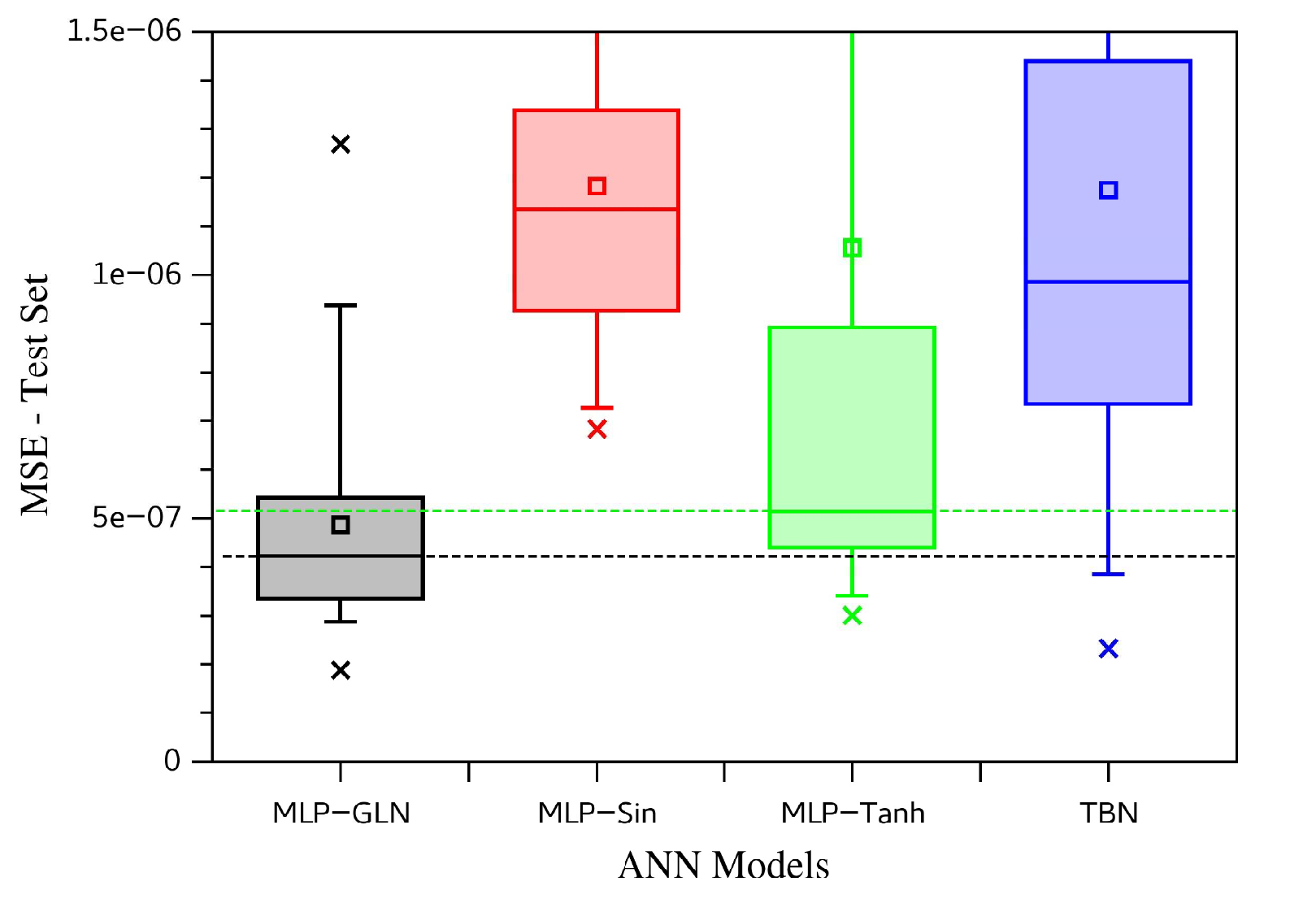}
         \caption{$1-20-20-1$ Architecture - Zoom scale.}
         \label{fig:HeatTestMSE_d}
     \end{subfigure}
    \caption{The MSE test set box-plot for all ANN models studied and both architectures for the Heat differential equation solving. In (a) and (b) are presented the MSE distributions for the 30 repetitions for each ANN model for architectures $1-20-1$, where the dashed lines are the median values references. In (c) and (d) are shown the results for architecture $1-20-20-1$.} \label{fig:HeatTestMSE}
\end{figure}

\begin{table}[!ht]
    \centering
     \caption{The descriptive statistics for all ANN models analyzed. All MSE measures are relative to the Heat differential equation solving. The best results are highlighted in bold-face.}
    \label{tab:HeatMSE}
    \begin{tabular}{cccccc}
    \hline
    \multicolumn{2}{c}{\multirow{2}{*}{\textbf{MSE}}} & \multicolumn{4}{c}{\textbf{ANN Models}}\\ 
    \cline{3-6}
     & & MLP-GLN & MLP-Sin & MLP-Tanh & TBN \\
    \hline
    \multirow{6}{*}{\rotatebox[origin=c]{90}{$1-20-1$}} 
    & Min.   & $1.563\cdot10^{-7}$ & $1.176\cdot10^{-7}$ & $1.322\cdot10^{-6}$ & $\mathbf{5.918\cdot10^{-8}}$ \\
    & Max.   & $2.388\cdot10^{-6}$ & $6.330\cdot10^{-6}$ & $9.684\cdot10^{-6}$ & $\mathbf{2.235\cdot10^{-6}}$ \\
    & Mean   & $6.367\cdot10^{-7}$ & $1.680\cdot10^{-6}$ & $4.092\cdot10^{-6}$ & $\mathbf{6.316\cdot10^{-7}}$ \\
    & Median & $\mathbf{4.988\cdot10^{-7}}$ & $9.670\cdot10^{-7}$ & $3.800\cdot10^{-6}$ & $5.526\cdot10^{-7}$ \\
    & Std.   & $5.044\cdot10^{-7}$ & $1.690\cdot10^{-6}$ & $1.988\cdot10^{-6}$ & $\mathbf{4.929\cdot10^{-7}}$ \\
    & CV     & $0.792$ & $1.006$ & $\mathbf{0.486}$ & $0.780$ \\
    \hline
    \multirow{6}{*}{\rotatebox[origin=c]{90}{$1-20-20-1$}}  
    & Min.   & $\mathbf{1.880\cdot10^{-7}}$ & $6.837\cdot10^{-7}$ & $3.011\cdot10^{-7}$ & $2.331\cdot10^{-7}$ \\
    & Max.   & $\mathbf{1.269\cdot10^{-6}}$ & $2.156\cdot10^{-6}$ & $1.017\cdot10^{-5}$ & $2.891\cdot10^{-6}$ \\
    & Mean   & $\mathbf{4.866\cdot10^{-7}}$ & $1.183\cdot10^{-6}$ & $1.055\cdot10^{-6}$ & $1.175\cdot10^{-6}$ \\
    & Median & $\mathbf{4.216\cdot10^{-7}}$ & $1.134\cdot10^{-6}$ & $5.133\cdot10^{-7}$ & $9.849\cdot10^{-7}$ \\
    & Std.   & $\mathbf{2.343\cdot10^{-7}}$ & $3.431\cdot10^{-7}$ & $1.794\cdot10^{-6}$ & $6.694\cdot10^{-7}$ \\
    & CV     & $0.482$ & $\mathbf{0.290}$ & $1.700$ & $0.570$ \\
    \hline
    \end{tabular}
\end{table}

Figure \ref{fig:HeatTestMSE} shows the MSE distributions. For the architecture $1-20-1$, Figures \ref{fig:HeatTestMSE_a} and \ref{fig:HeatTestMSE_b}, the MLP-GLN and TBN models had a MSE performance better than pure ANN models. Table \ref{tab:HeatMSE} presents all descriptive statistics for those MSE distributions. The MLP-GLN and TBN MSE distributions came from the same population distribution according to the two-sample KS test Table \ref{tab:KS_Heat}. The MLP-GLN and TBN reached statistical central measures very similar, with MSE mean values of $6.367 \cdot 10^{-7}$ and $6.316\cdot 10^{-7}$, respectively, and a median of  $4.988\cdot 10^{-7}$ and $5.526 \cdot 10^{-7}$.

For the architecture $1-20-20-1$, the MLP-GLN presented the best MSE performance, as demonstrated by Figures \ref{fig:HeatTestMSE_c} and \ref{fig:HeatTestMSE_d}. For this architecture, the MLP-GLN MSE distribution came from a distinct population distribution when compared with the other analyzed ANN models, as can be viewed in Table \ref{tab:KS_Heat} with the two sample KS test results. 

\begin{table}[!htb]
    \centering
    \caption{Two-sample Kolmogorov-Smirnov Test at the $5\%$ significance level for the MSE distributions between the MLP-GLN and all other models for both architectures studied for the Heat differential equation solving.}
    \label{tab:KS_Heat}
    \setlength{\extrarowheight}{5.5pt}
    \begin{tabular}{ccccc}
    \hline
    \multicolumn{3}{c}{\textbf{Tested Model}} & \multicolumn{2}{c}{\textbf{KS Test Results}}\\ 
    \cline{4-5}
    \multicolumn{3}{c}{\textbf{whit MLP-GLN}} & Statistically Similar & $p$-values\\
    \hline
    \multirow{3}{*}{\scriptsize \rotatebox[origin=c]{90}{$1-20-1$} }  &\multirow{3}{*}{\rotatebox[origin=c]{90}{MSE}}
    & MLP-Sin    & No & $0.011$ \\
    & & MLP-Tanh & No & $8.383\cdot10^{-12}$ \\
    & & TBN      & Yes & $0.936$ \\
    \hline
    \multirow{3}{*}{\scriptsize \rotatebox[origin=c]{90}{$1-20-20-1$} }  &\multirow{3}{*}{\rotatebox[origin=c]{90}{MSE}}
    & MLP-Sin    & No & $5.631\cdot10^{-11}$ \\
    & & MLP-Tanh & No & $0.011$ \\
    & & TBN      & No & $1.143\cdot10^{-6}$ \\
    \hline
    \end{tabular}
\end{table}

\begin{figure}[!htb]
    \centering
     \begin{subfigure}[b]{0.47\textwidth}
         \centering
         \includegraphics[width=\textwidth]{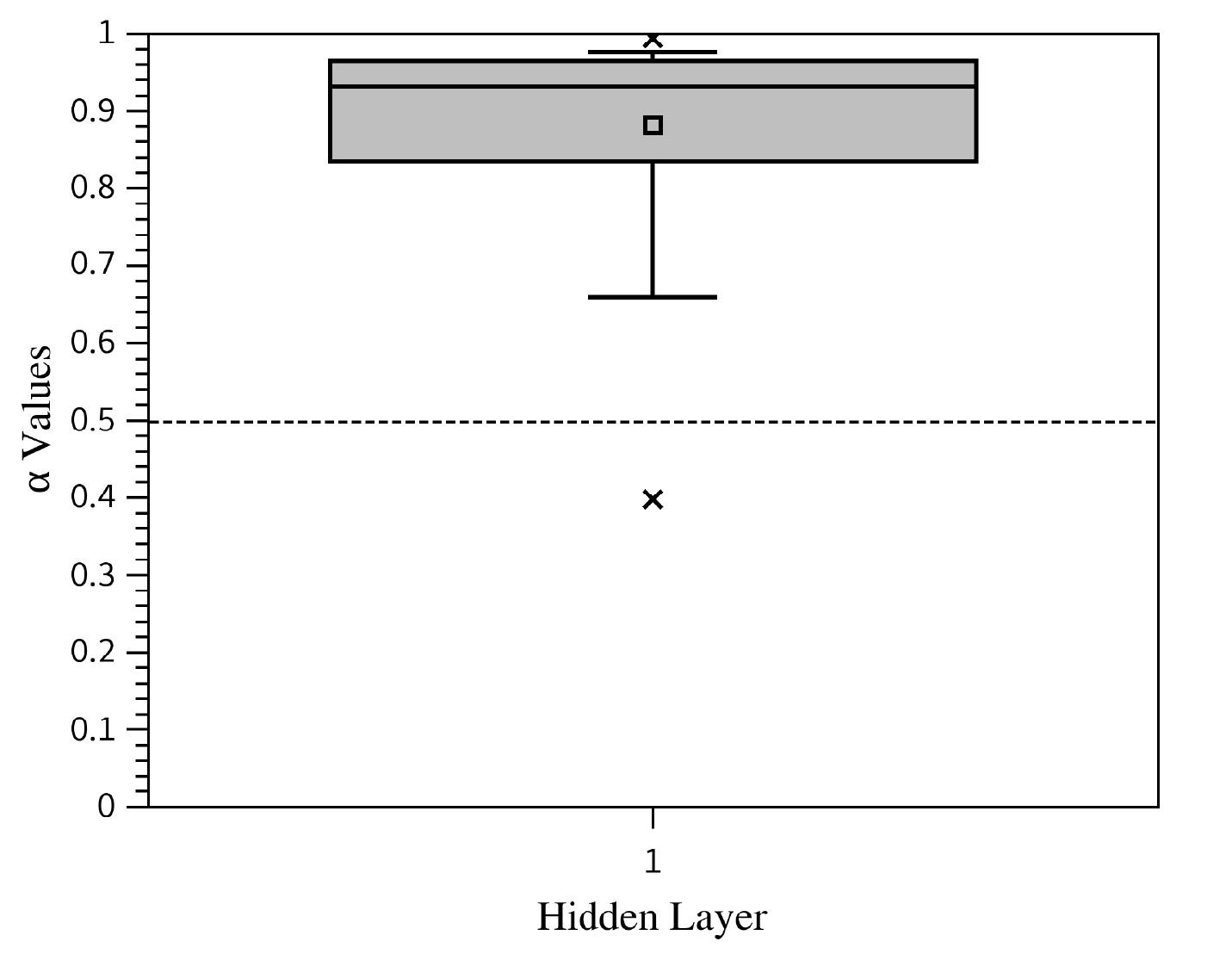}
         \caption{$1-20-1$ Architecture.}
         \label{fig:Heat_alphas_a}
     \end{subfigure}
     \hfill
     \begin{subfigure}[b]{0.47\textwidth}
         \centering
         \includegraphics[width=\textwidth]{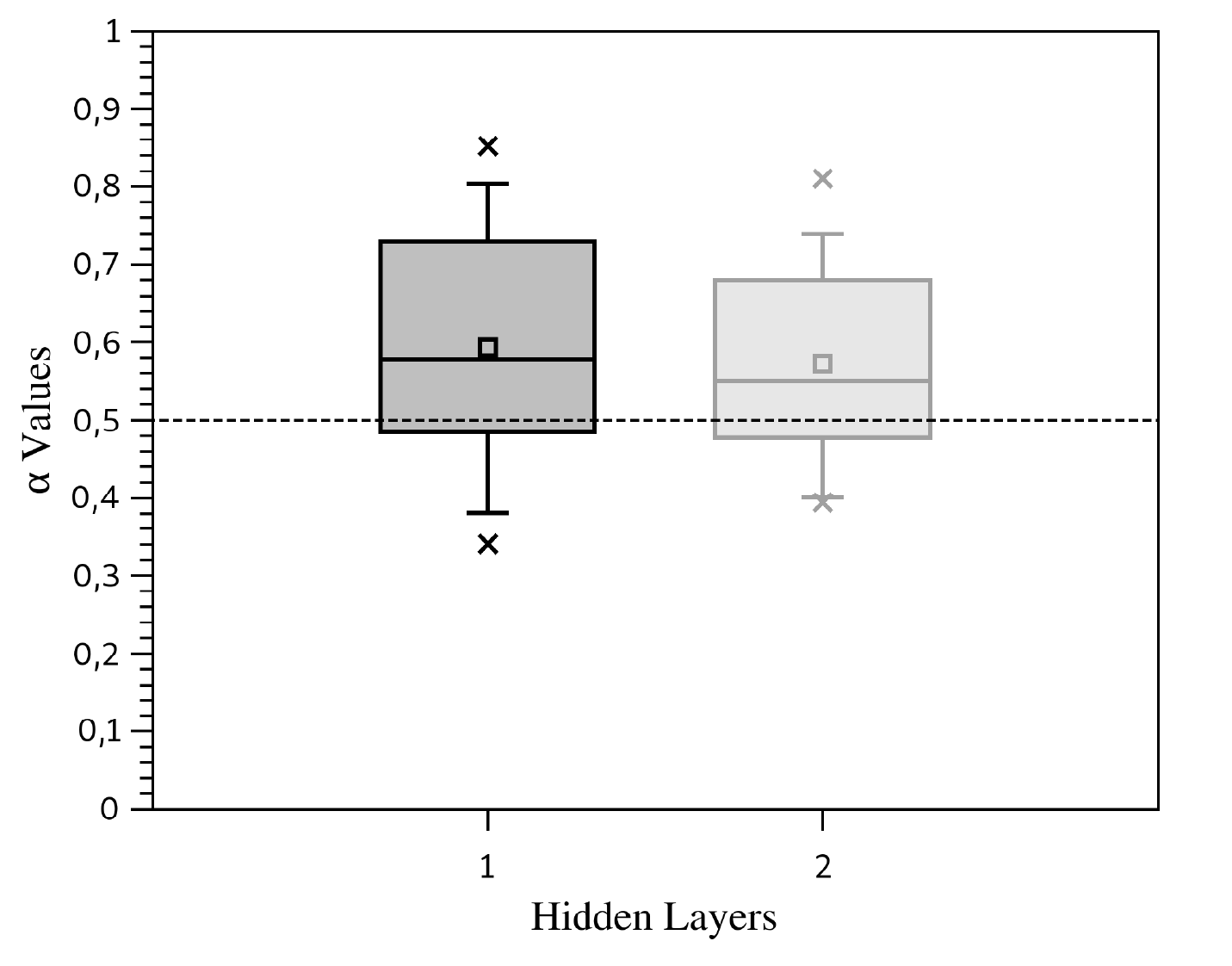}
         \caption{$1-20-20-1$ Architecture.}
         \label{fig:Heat_alphas_b}
     \end{subfigure}
     \caption{The $\alpha$ distributions for the MLP-GLN models after the training process for the Heat differential equation solving.}
     \label{fig:Heat_alphas}
\end{figure}

Observing the $\alpha$ values distribution, Figure \ref{fig:Heat_alphas}, it is possible to verify the expectation for the activation function components combination. Figure \ref{fig:Heat_alphas_a} shows the $\alpha$ values distribution for the architecture with one hidden layer. In this case, the MLP-GLN gave more importance to the global component, with an $\alpha$ mean value of $0.881$ and a median value of $0.931$. For the architecture $1-20-20-1$, Figure \ref{fig:Heat_alphas_b}, both hidden layers presented a tendency to enhance the global component, with a mean value of $0.594$ and a median value of $0.578$ for the first hidden layer, and $0.572$ and $0.550$ for the second hidden layer. 

\subsubsection{Kuramoto-Sivashinsky Equation Results}

The Kuramoto-Sivashinsky differential equation presented in Section \ref{sec:KuSiEq} is a very complex nonlinear differential equation. The analysis employed here, given the set of initial and boundary conditions, compares the MSE performance of the pure ANN models (MLP-Sin and MLP-Tanh) and the ANN composite models (MLP-GLN, and TBN). The experiments were executed in the domain of $x \in [-40,40]$ and $t \in [0,20]$. The number of epochs of $1500$ was employed for the training process for all ANN models. 

\begin{figure}[!thb]
     \centering
     \begin{subfigure}[b]{0.47\textwidth}
         \centering
         \includegraphics[width=\textwidth]{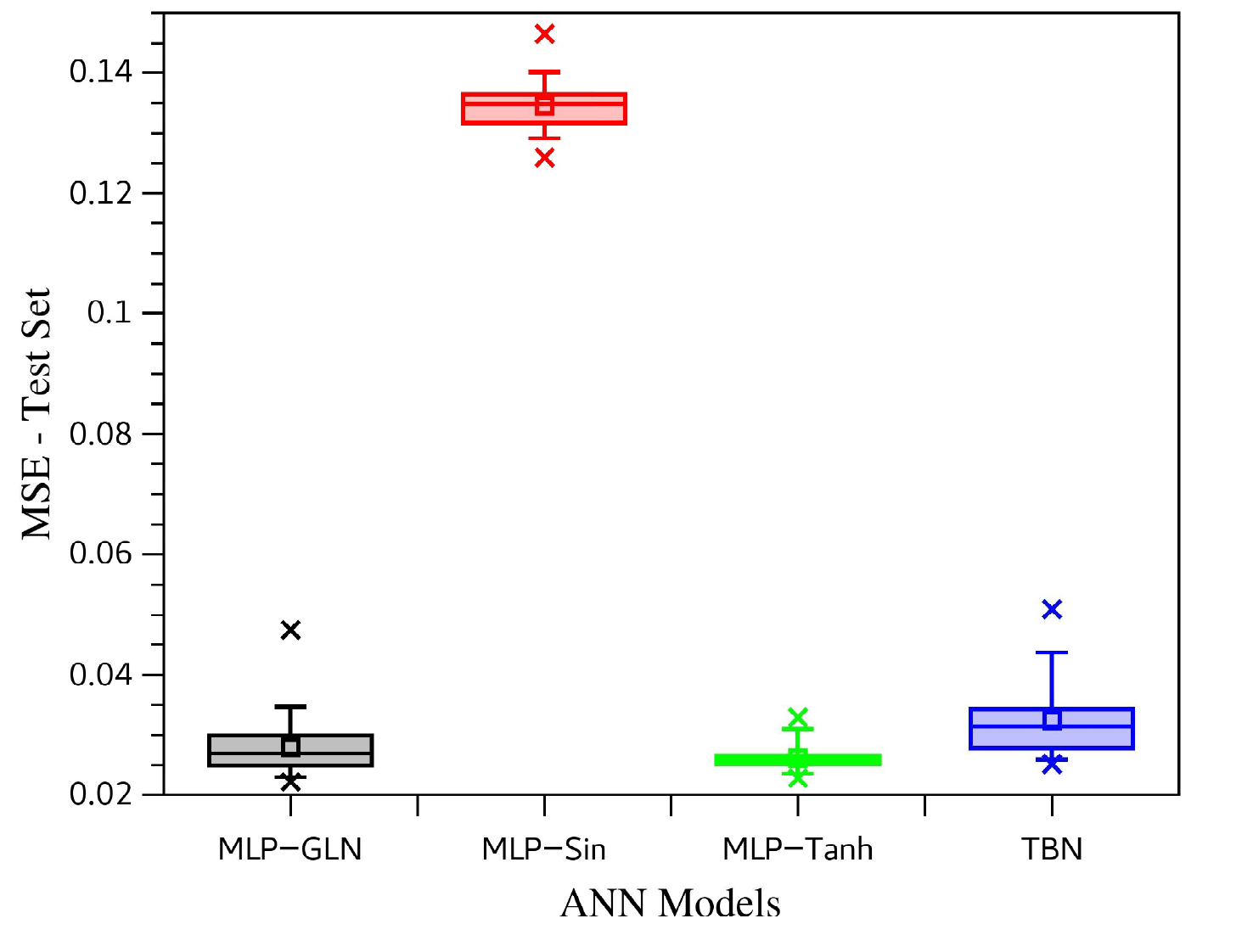}
         \caption{$1-20-1$ Architecture.}
         \label{fig:KuSiTestMSE_a}
     \end{subfigure}
     \hfill
     \begin{subfigure}[b]{0.47\textwidth}
         \centering
         \includegraphics[width=\textwidth]{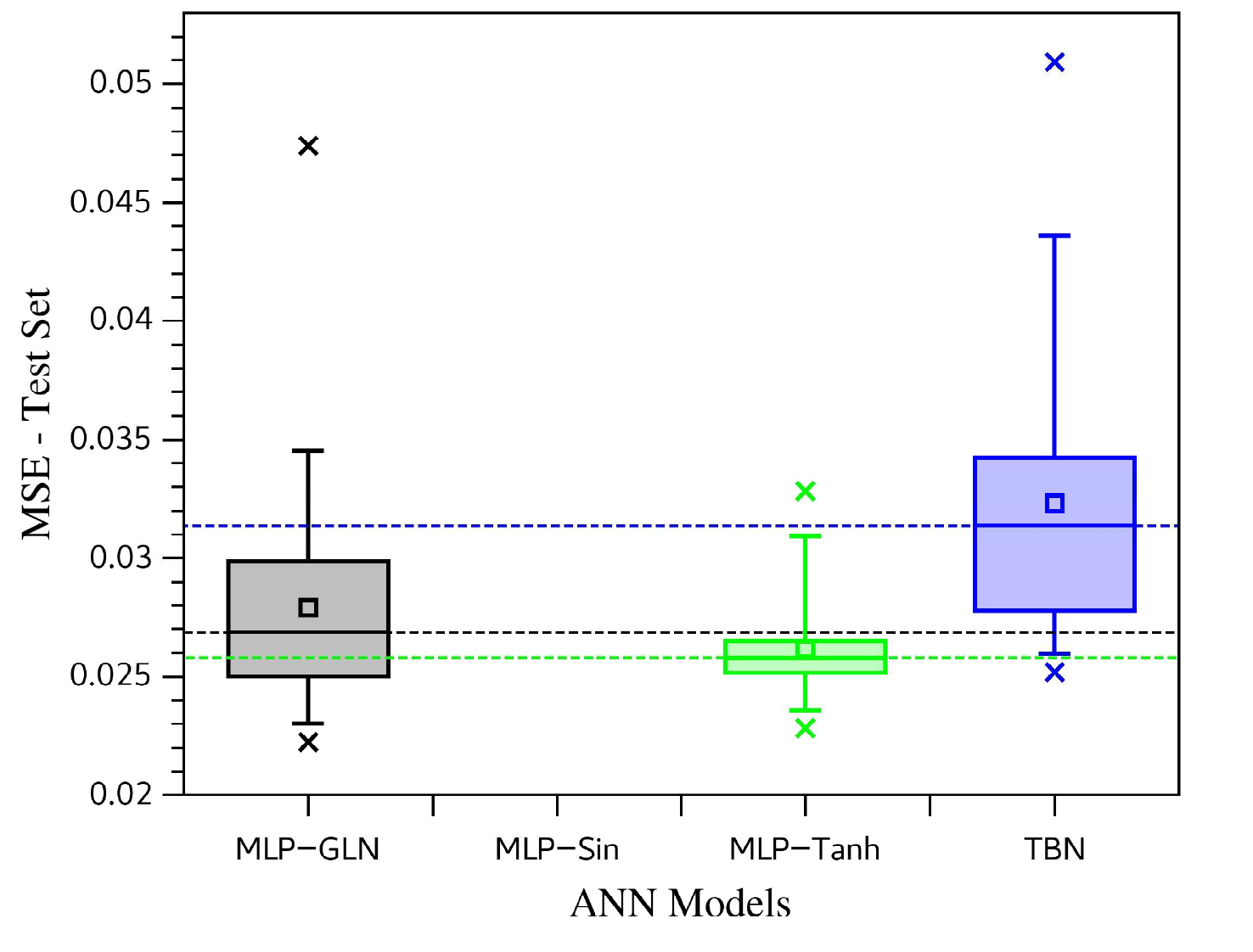}
         \caption{$1-20-20-1$ Architecture - Zoom scale.}
         \label{fig:KuSiTestMSE_b}
     \end{subfigure}
     \\
     \vspace{20pt}    
     \begin{subfigure}[b]{0.47\textwidth}
         \centering
         \includegraphics[width=\textwidth]{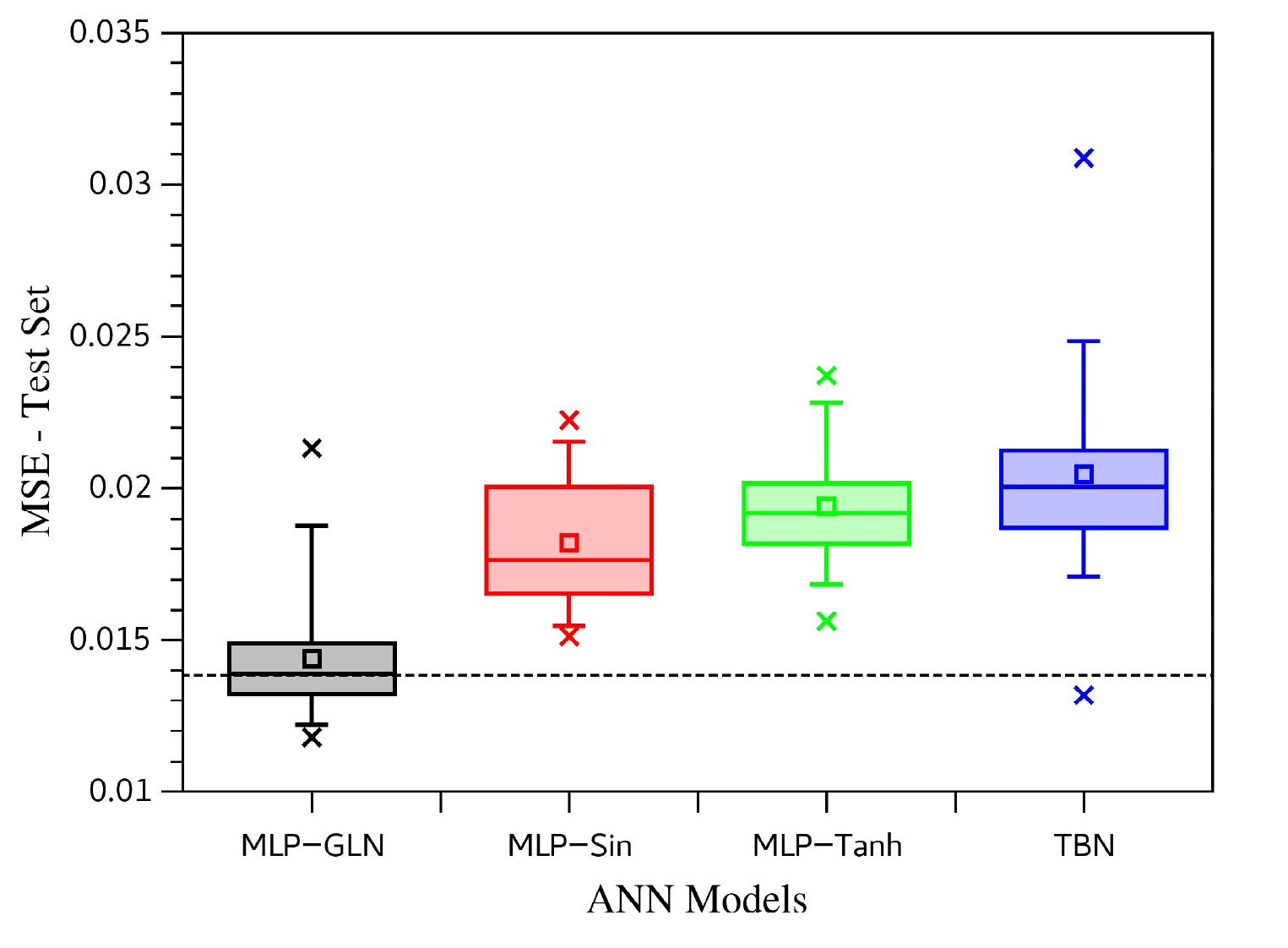}
         \caption{$1-20-20-1$ Architecture.}
         \label{fig:KuSiTestMSE_c}
     \end{subfigure}
    \caption{The test set MSE box plot for all ANN models studied and both architectures for the Kuramoto-Sivashinsky differential equation solving. In (a) and (b) are presented the MSE distributions for the 30 repetitions for each ANN model for architectures $1-20-1$. In (c) is shown the results for architecture $1-20-20-1$. The dashed lines are the median values references.} \label{fig:KuSiTestMSE}
\end{figure}

Observing Figure \ref{fig:KuSiTestMSE}, it is possible to view the test set MSE distributions for all ANN models investigated. For the architecture $1-20-1$, Figures \ref{fig:KuSiTestMSE_a} and \ref{fig:KuSiTestMSE_b}, the two best models were the MLP-GLN and MLP-Tan. The MSE distributions' mean values for these two better ANN models were $0.028$ and $0.026$, respectively, for MLP-GLN and MLP-Tanh. The median values were $0.027$ and $0.026$. All descriptive statistics are shown in Table \ref{tab:KuSiMSE}. Although the central values statistical measures for the MSE distributions are very close to each other, the dispersion of the MLP-GLN MSE distribution is bigger than the dispersion of the MLP-Tanh. The MLP-GLN has a CV of $0.174$, and the MLP-Tanh has a CV of $0.085$. However, it is observed that the minimum MSE value reached by the ANN models, the MLP-GLN achieved the best result.

For the architecture $1-20-20-1$, Figure \ref{fig:KuSiTestMSE_c}, the MLP-GLN presents the best MSE performance, with both mean and median values of $0.014$, as described in Table \ref{tab:KuSiMSE}. In this case, the MLP-GLN also reached the best results for minimum and maximum MSE values. 

\begin{table}[!htb]
    \centering
     \caption{The descriptive statistics for all ANN models analyzed. All test set MSE measures are relative to the Kuramoto-Sivashinsky differential equation solving. The best results are highlighted in bold-face.}
    \label{tab:KuSiMSE}
    \begin{tabular}{cccccc}
    \hline
    \multicolumn{2}{c}{\multirow{2}{*}{\textbf{MSE}}} & \multicolumn{4}{c}{\textbf{ANN Models}}\\ 
    \cline{3-6}
     & & MLP-GLN & MLP-Sin & MLP-Tanh & TBN \\
    \hline
    \multirow{6}{*}{\rotatebox[origin=c]{90}{$1-20-1$}} 
    & Min.   & $\mathbf{0.022}$ & $0.126$ & $0.023$ & $0.025$\\
    & Max.   & $0.047$ & $0.147$ & $\mathbf{0.033}$ & $0.051$ \\
    & Mean   & $0.028$ & $0.135$ & $\mathbf{0.026}$ & $0.032$ \\
    & Median & $0.027$ & $0.135$ & $\mathbf{0.026}$ & $0.031$ \\
    & Std.   & $4.843\cdot10^{-3}$ & $4.119\cdot10^{-3}$ & $\mathbf{2.217\cdot10^{-3}}$ & $5.793\cdot10^{-3}$ \\
    & CV     & $0.174$ & $0.031$ & $\mathbf{0.085}$ & $0.179$ \\
    \hline
    \multirow{6}{*}{\rotatebox[origin=c]{90}{$1-20-20-1$}}  
    & Min.   & $\mathbf{0.012}$ & $0.015$ & $0.016$ & $0.013$ \\
    & Max.   & $\mathbf{0.021}$ & $0.022$ & $0.024$ & $0.031$ \\
    & Mean   & $\mathbf{0.014}$ & $0.018$ & $0.019$ & $0.020$ \\
    & Median & $\mathbf{0.014}$ & $0.018$ & $0.019$ & $0.020$ \\
    & Std.   & $2.119\cdot10^{-3}$ & $2.080\cdot10^{-3}$ & $\mathbf{1.942\cdot10^{-3}}$ & $3.148\cdot10^{-3}$ \\
    & CV     & $0.147$ & $0.114$ & $\mathbf{0.100}$ & $0.154$ \\
    \hline
    \end{tabular}
\end{table}

\begin{table}[!htb]
    \centering
    \caption{Two-sample Kolmogorov-Smirnov Test at the $5\%$ significance level for the MSE distributions between the MLP-GLN and all other models for both architectures studied for solving the Kuramoto-Sivashinsky differential equation.}
    \label{tab:KS_KuSi}
    \setlength{\extrarowheight}{5.5pt}
    \begin{tabular}{ccccc}
    \hline
    \multicolumn{3}{c}{\textbf{Tested Model}} & \multicolumn{2}{c}{\textbf{KS Test Results}}\\ 
    \cline{4-5}
    \multicolumn{3}{c}{\textbf{whit MLP-GLN}} & Statistically Similar & $p$-values\\
    \hline
    \multirow{3}{*}{\scriptsize \rotatebox[origin=c]{90}{$1-20-1$} }  &\multirow{3}{*}{\rotatebox[origin=c]{90}{MSE}}
    & MLP-Sin    & No & $1.797\cdot 10^{-14}$ \\
    & & MLP-Tanh & No & $0.0259$ \\
    & & TBN      & No & $0.002$ \\
    \hline
    \multirow{3}{*}{\scriptsize \rotatebox[origin=c]{90}{$1-20-20-1$} }  &\multirow{3}{*}{\rotatebox[origin=c]{90}{MSE}}
    & MLP-Sin    & No & $2.048\cdot 10^{-9}$ \\
    & & MLP-Tanh & No & $8.384\cdot 10^{-12}$ \\
    & & TBN      & No & $8.384\cdot 10^{-12}$ \\
    \hline
    \end{tabular}
\end{table}

For both architectures, the MLP-GLN presents an MSE distribution that comes from a distinct populational distribution, when compared with other ANN models. Table \ref{tab:KS_KuSi} shows the results of the two samples KS test at $5\%$ of significance.

\begin{figure}[!htb]
    \centering
     \begin{subfigure}[b]{0.49\textwidth}
         \centering
         \includegraphics[width=\textwidth]{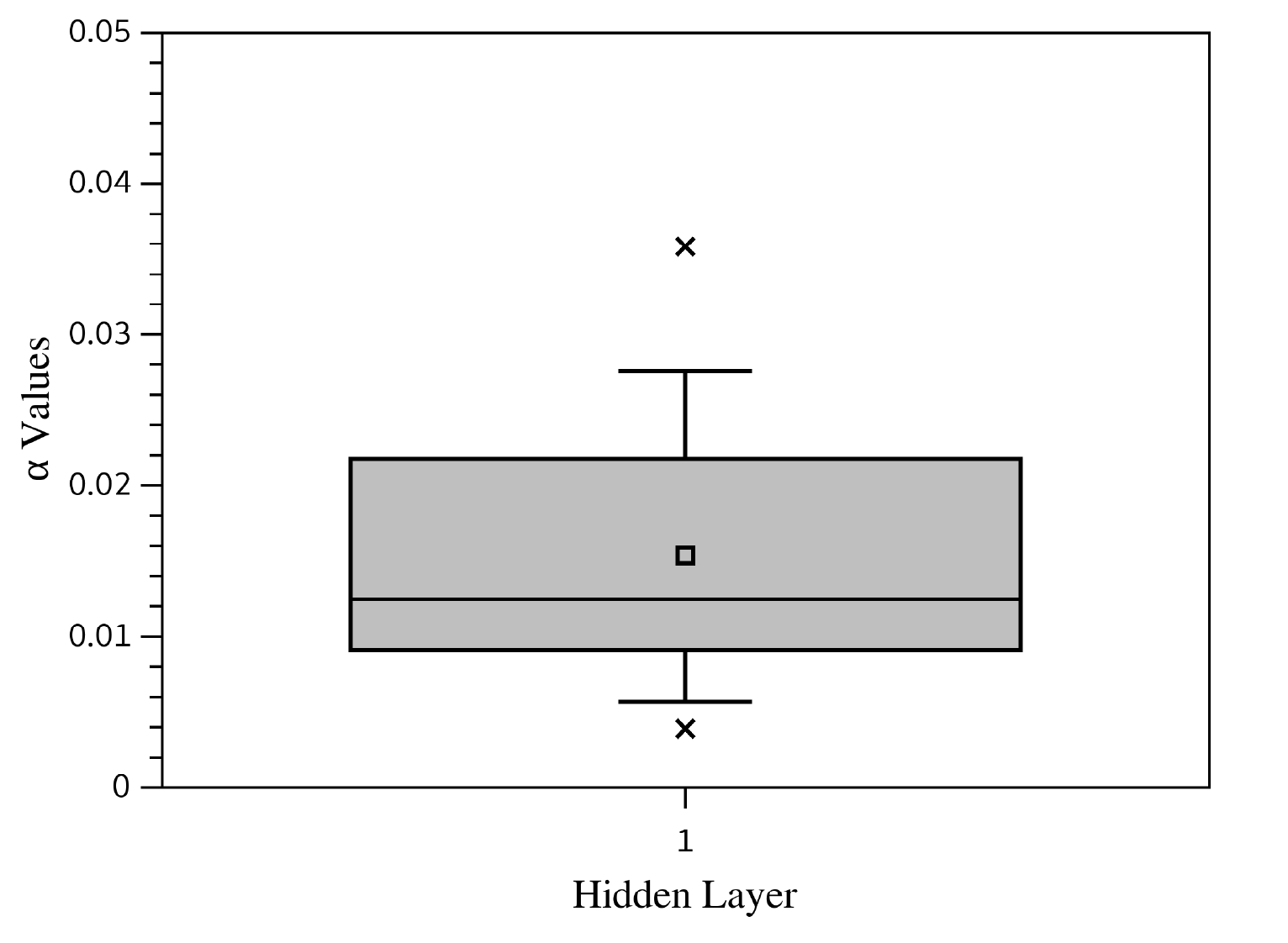}
         \caption{$1-20-1$ Architecture.}
         \label{fig:KuSi_alphas_a}
     \end{subfigure}
     \hfill
     \begin{subfigure}[b]{0.49\textwidth}
         \centering
         \includegraphics[width=\textwidth]{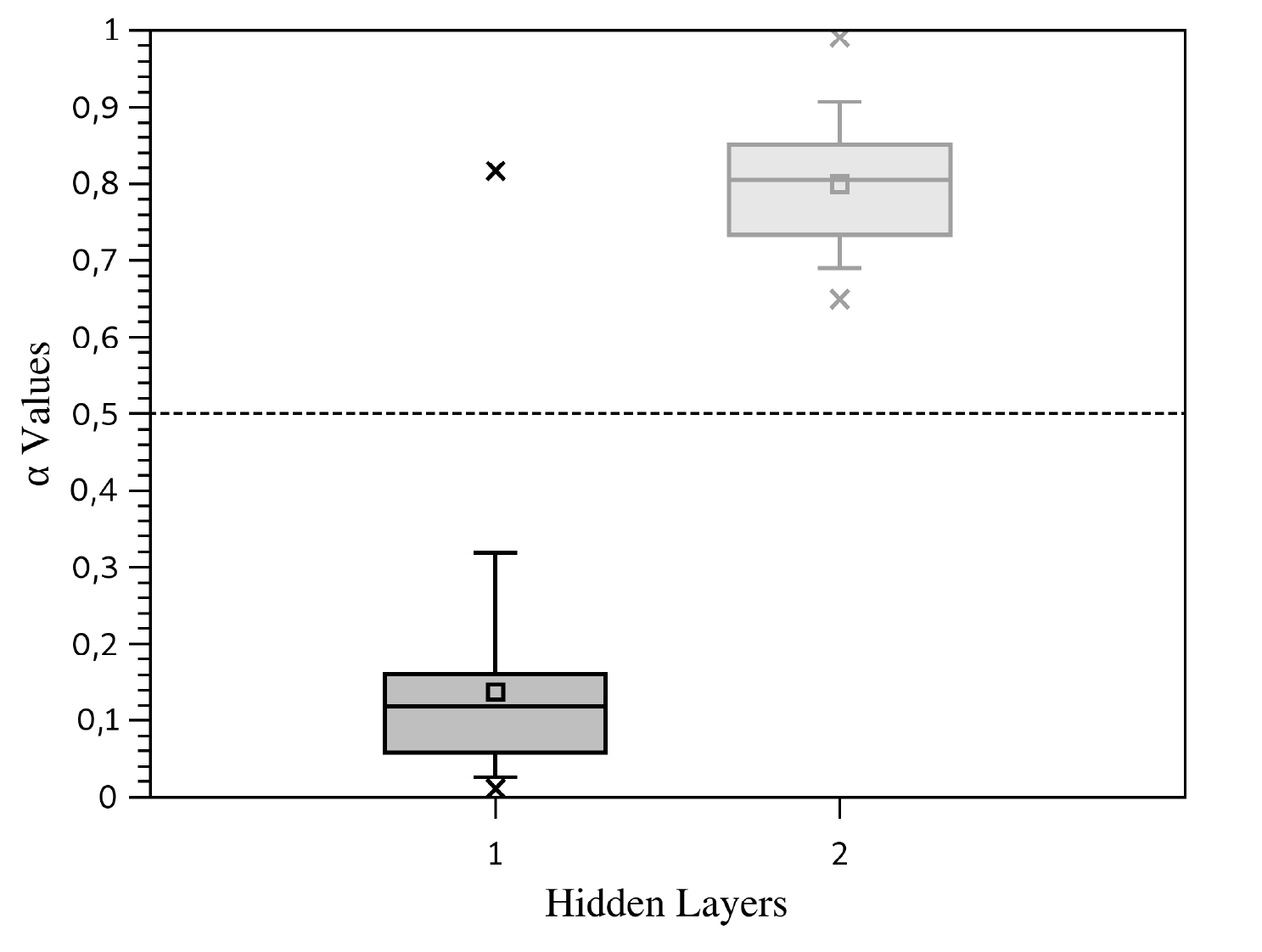}
         \caption{$1-20-20-1$ Architecture.}
         \label{fig:KuSi_alphas_b}
     \end{subfigure}
     \caption{The $\alpha$ distributions for the MLP-GLN models after the training process for the Kuramoto-Sivashinsky differential equation solving.}
     \label{fig:KuSi_alphas}
\end{figure}

Figure \ref{fig:KuSi_alphas} presents the $\alpha$ values distribution for both MLP-GLN architectures. For the architecture with only one hidden layer, Figure \ref{fig:KuSi_alphas_a}, the MLP-GLN reaches a behavior where the activation function local component has more importance than the global component. With a mean value of $0.015$ and a median of $0.012$, the MLP-GLN has a very similar MLP-Tanh structure. For the architecture $1-20-20-1$, Figure \ref{fig:KuSi_alphas_b}, the MLP-GLN presents a first hidden layer with predominate local component, with $\alpha$ mean value of $0.138$ and a media value of $0.119$. The second hidden shows a dominance of the global component, with $\alpha$ mean value of $0.800$ and a median value of $0.805$. Thus, for this architecture with two hidden layers, the MLP-GLN specialized the hidden layers with distinct abilities, demonstrating the MLP-GLN capability to combine the local and global components in different layers.

\section{Conclusions}\label{sec:Conclusions}

This article proposes a new artificial neuron with an activation function composed of two different mathematical functions, a function with local behavior and a function with global features. Here, the local function used was the hyperbolic tangent, and the global was the trigonometrical sine. The term local for the $\tanh(x)$ was employed because the relevant variation of the function is located in a narrow range around $x = 0$, and it tends to $1$ or $-1$ when $x \to \infty$ or $x \to -\infty$. On the other hand, the $\sin(x)$ is called global component because it has the same variation behavior for all domain $x \in [-\infty, \infty]$.

The $\sin(x)$ and $\tanh(x)$ were chosen here because the problems studied in general have global and local features, or some combination of those. Other mathematical functions could also be used, where the proposed neuron would combine the characteristics of the chosen mathematical functions.

It was done a massive experimental test with the proposed neuron approaching two types of problems, the regression problem and differential equation solving, totaling $2400$ simulations. These problem classes were chosen because it is common to find situations where the problem solution combines a local and global feature. More specifically, ten individual problems were studied, three regression problems and seven differential equations.

In general, the proposed MLP-GLN reached a better MSE performance distribution for the tested problems. However, in some problems, the MLP-GLN had statistically similar behavior with a traditional MLP, with only sine function or hyperbolic tangent function, or it had statistically identical behavior to a combination of these two traditional MLP, the TBN model. Thus, the MLP-GLN could adapt its activation function composition to efficiently solve the problems. 

All MLP-GLN experiments used the same activation function per hidden layer. In this way, the $\alpha$ weight, that combines the local and global activation function components, is the same for all neurons in the same hidden layer. Therefore, it was possible to verify each hidden layer tendency by choosing the importance of the local and global components and thus, extract  information about the importance of each global and local component in the analyzed problem. 

Since the output layer of the MLP-GLN is only one neuron with a linear activation function, the architecture with one hidden layer makes a combination of the local and global functions without doing a mathematical composition of these two functions. Thus, the MLP-GLN could define an efficient and automatic combination of these two local and global functions in such a way to reach a better or equivalent performance to the ANN with a single activation function or with the linear combination of the two sine and hyperbolic tangent ANN, the TBN model. 

The MLP-GLN demonstrated the best MSE performance for all experiments done for the architecture with two hidden layers. For this architecture, the MLP-GLN makes a mathematical composition of the first hidden layer activation function with the activation function of the second hidden layer. Combining the two components plus the composition of the activation functions of the hidden layers gave MLP-GLN the best versatility to solve the problem with better MSE precision than using one layer. Thus, a deeper architecture yields better performance.

Therefore, the MLP-GLN presents a possibility to reach better results than the ANN with single activation functions or a linear combination of these ANNs, with automatic adjusting of the relative importance between the activation function components. In this way, many other function types with different features can be also combined in pairs to better describe problems containing different feature components.

\section*{Acknowledges}

To the Science and Technology Support Foundation of Pernambuco (FACE\-PE) Brazil, Brazilian National Council for Scientific and Technological Development (CNPq), Coordena\c{c}\~{a}o de Aperfei\c{c}oamento de Pessoal de N\'{i}vel Superior - Brasil (CAPES) - Finance Code 001 by financial support for the development of this research, and the Institute for Applied Computational Science (IACS) at Harvard University for host the professor Tiago A. E. Ferreira.






\bibliographystyle{elsarticle-num-names}
\bibliography{Biblio.bib}







\end{document}